%% file: main.tex
\documentclass[10pt,journal,compsoc]{IEEEtran}

\ifCLASSOPTIONcompsoc
  \usepackage[nocompress]{cite}
\else
  \usepackage{cite}
\fi

\usepackage[utf8]{inputenc} %
\usepackage[T1]{fontenc}    %
\usepackage{url}            %
\usepackage{booktabs}       %
\usepackage{amsfonts}       %
\usepackage{amsmath}        %
\usepackage{nicefrac}       %
\usepackage{microtype}      %
\usepackage{graphicx}
\usepackage{makecell}
\usepackage{multirow}
\usepackage{subcaption}
\usepackage{paralist} %

\usepackage[pagebackref=false,breaklinks=true,colorlinks,urlcolor=blue,citecolor=blue,linkcolor=blue,bookmarks=false]{hyperref}
\usepackage{tikz}
\usepackage{algorithm}
\usepackage[noend]{algpseudocode}
\usepackage{algorithmicx}

\newcommand{\etal}{\textit{et~al}\mbox{.}}

\newcommand{\figref}[1]{Fig.~\ref{fig:#1}}%
\newcommand{\tabref}[1]{TABLE~\ref{tab:#1}} %

\newcommand{\revision}[1]{#1}
\newcommand{\revisionsecond}[1]{#1}

\newcommand{\heading}[1]{\vspace{2mm}\noindent\textbf{#1}}
\long\def\ignorethis#1{}

\newcommand{\tb}[1]{\textbf{#1}}

\newlength{\threeimg}

\newlength\paramargin
\newlength\figmargin
\newlength\secmargin
\newlength\figcapmargin

\newboolean{revising}
\setboolean{revising}{true}

\ifthenelse{\boolean{revising}} {
\newcommand{\yylin}[1]{\textcolor{blue}{#1}}}
{
\newcommand{\yylin}[1]{}
}

\begin{document}

\title{Learning to See Through Obstructions with Layered Decomposition}

\author{
Yu-Lun~Liu,
Wei-Sheng~Lai,
Ming-Hsuan~Yang,
Yung-Yu~Chuang
and~Jia-Bin~Huang
\IEEEcompsocitemizethanks{
\IEEEcompsocthanksitem Y.-L. Liu and Y.-Y. Chuang are with the Department of Computer Science and Information Engineering, National Taiwan University, Taipei, Taiwan 10617, 
E-mail: yulunliu@cmlab.csie.ntu.edu.tw, cyy@csie.ntu.edu.tw
\IEEEcompsocthanksitem W.-S. Lai and M.-H. Yang are with School of Engineering, University of California, Merced, CA, US, 
E-mail: wlai24@ucmerced.edu, mhyang@ucmerced.edu
\IEEEcompsocthanksitem J.-B. Huang is with Department of Electrical and Computer Engineering, Virginia Tech, VA, US, 
E-mail: jbhuang@vt.edu}%
}

\markboth{}%
{Shell \MakeLowercase{\textit{et al.}}: Bare Demo of IEEEtran.cls for Computer Society Journals}

\IEEEtitleabstractindextext{%

\input{0_abstract}

\begin{IEEEkeywords}
reflection removal, fence removal, optical flow, layer decomposition, computational photography.
\end{IEEEkeywords}}

\maketitle

\IEEEdisplaynontitleabstractindextext

\IEEEpeerreviewmaketitle

\input{1_introduction}

\input{2_related-work}

\input{3_method}

\input{4_experiment}

\input{5_conclusion}

\newpage

\ifCLASSOPTIONcaptionsoff
  \newpage
\fi

\bibliographystyle{IEEEtran}
\bibliography{IEEEabrv,reference.bib}

\vspace{-8mm}

\begin{IEEEbiography}[{\includegraphics[width=1in,height=1.25in,clip,keepaspectratio]{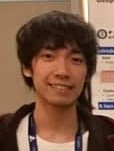}}]{Yu-Lun Liu} is a Ph.D. candidate of Computer Science and Information Engineering at National Taiwan University, Taipei, Taiwan. 
He received the B.S. and M.S. degree in Electronics Engineering from National Chiao-Tung University, Hsinchu, Taiwan, in 2012 and 2014, respectively.
His research interests include computer vision, machine learning, and multimedia.
\end{IEEEbiography}

\vspace{-8mm}

\begin{IEEEbiography}[{\includegraphics[width=1in,height=1.25in,clip,keepaspectratio]{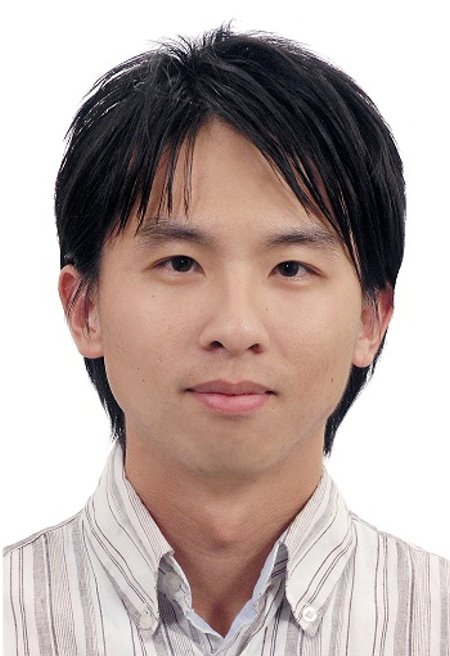}}]{Wei-Sheng Lai} is a software engineer at Google, working on mobile computational photography.
He received the B.S. and M.S. degrees from the EE department at National Taiwan University in 2012 and 2014, respectively, and his Ph.D. degree from the EECS department at University of California, Merced in 2019.
\end{IEEEbiography}

\vspace{-8mm}

\begin{IEEEbiography}[{\includegraphics[width=1in,height=1.25in,clip,keepaspectratio]{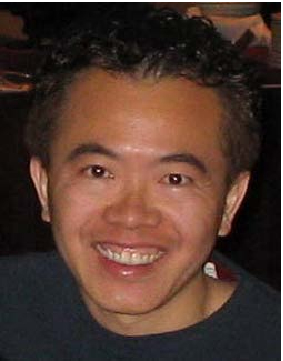}}]{Ming-Hsuan Yang}
is a Professor in Electrical
Engineering and Computer Science at University
of California, Merced. 
He served as an associate editor of the IEEE Transactions on Pattern Analysis and Machine Intelligence from 2007 to 2011, and is an associate editor of the International Journal of Computer Vision, Image and Vision Computing and Journal of Artificial Intelligence Research. Yang received the NSF CAREER award in 2012 and the Google Faculty Award in 2009. 
He is a Fellow of the IEEE.
\end{IEEEbiography}

\vspace{-8mm}

\begin{IEEEbiography}[{\includegraphics[width=1in,height=1.25in,clip,keepaspectratio]{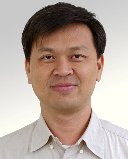}}]{Yung-Yu Chuang} received his B.S. and M.S. from National Taiwan University in 1993 and 1995 respectively, and the Ph.D. from the University
of Washington at Seattle in 2004, all in Computer Science. 
He is currently a professor with the Department of Computer Science and Information Engineering, National Taiwan University. 
His research interests include computational photography, computer vision and rendering.
\end{IEEEbiography}

\vspace{-8mm}

\begin{IEEEbiography}[{\includegraphics[width=1in,height=1.25in,clip,keepaspectratio]{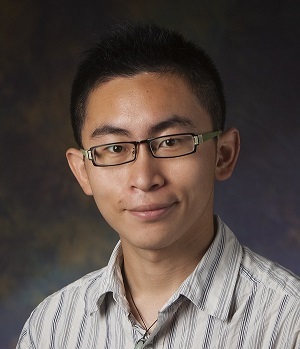}}]{Jia-Bin Huang} is an assistant professor in the Bradley Department of Electrical and Computer Engineering at Virginia Tech.
He received the B.S. degree in Electronics Engineering from National Chiao-Tung University, Hsinchu, Taiwan and his Ph.D. degree in the Department of Electrical and Computer Engineering at University of Illinois, Urbana-Champaign in 2016. 
\end{IEEEbiography}

\end{document}


\title{Learning to See Through Obstructions with Layered Decomposition \\ 
Supplementary Material}

\author{
Yu-Lun~Liu,
Wei-Sheng~Lai,
Ming-Hsuan~Yang,
Yung-Yu~Chuang
and~Jia-Bin~Huang
}

\markboth{}%
{Shell \MakeLowercase{\textit{et al.}}: Bare Demo of IEEEtran.cls for Computer Society Journals}

\maketitle

\IEEEdisplaynontitleabstractindextext

\IEEEpeerreviewmaketitle

\input{supp_content}

\ifCLASSOPTIONcaptionsoff
  \newpage
\fi

\bibliographystyle{IEEEtran}
\bibliography{IEEEabrv,reference.bib}

%% file: 0_abstract.tex
\begin{abstract}
We present a learning-based approach for removing unwanted obstructions, such as window reflections, fence occlusions, or \revision{adherent raindrops}, from a short sequence of images captured by a moving camera.
Our method leverages motion differences between the background and obstructing elements to recover both layers.
Specifically, we alternate between estimating dense optical flow fields of the two layers and reconstructing each layer from the flow-warped images via a deep convolutional neural network. 
This learning-based layer reconstruction module facilitates accommodating potential errors in the flow estimation and brittle assumptions, such as brightness consistency.
We show that the proposed approach learned from synthetically generated data performs well to real images.
Experimental results on numerous challenging scenarios of reflection and fence removal demonstrate the effectiveness of the proposed method.
\end{abstract}

%% file: 1_introduction.tex
\setlength{\threeimg}{0.328\textwidth}
\setlength{\fboxsep}{0pt}
\begin{figure*}[t]
  \centering
  \begin{subfigure}[b]{\threeimg}
    \centering\fbox{\includegraphics[width=0.97\linewidth]{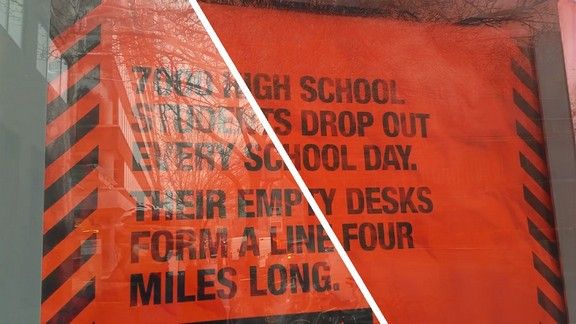}}
    \centering\caption*{(a) Reflection removal}
  \end{subfigure}
  \begin{subfigure}[b]{\threeimg}
    \centering\fbox{\includegraphics[width=0.97\linewidth]{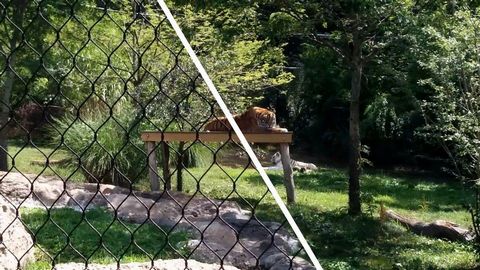}}
    \centering\caption*{(b) Fence removal}
  \end{subfigure}
  \begin{subfigure}[b]{\threeimg}
    \centering\fbox{\includegraphics[width=0.97\linewidth]{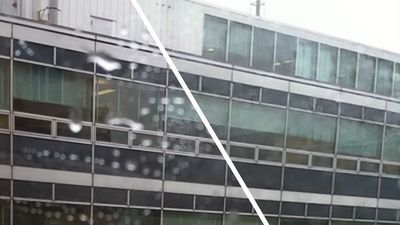}}
    \centering\caption*{(c) \revision{Adherent raindrop} removal}
  \end{subfigure}
  \caption{
  \textbf{Seeing through obstructions.} 
  We present a learning-based method for recovering clean images from a given short sequence of images taken by a moving camera through obstructing elements such as (a) windows, (b) fence, or (c) \revision{adherent raindrop}.
  }
  \label{fig:teaser}
\end{figure*}

\IEEEraisesectionheading{\section{Introduction}\label{sec:introduction}}
\IEEEPARstart{C}{apturing} clean photographs through reflective surfaces (such as windows) or occluding elements (such as fences) is challenging as the captured images inevitably contain both the scenes of interests and the obstructions caused by reflections or occlusions. 
An effective solution to recover the underlying clean image is thus of great interest for improving the quality of the images captured under such conditions or allowing computers to form a correct physical interpretation of the scene, e.g., enabling a robot to navigate in a scene with windows safely.

Recent efforts have been focused on removing unwanted reflections or occlusions from \emph{a single image}~\cite{arvanitopoulos2017single, fan2017generic, jin2018learning, jonna2016deep, park2010image, wei2019single, yang2018seeing, zhang2018single}.
These methods either leverage the ghosting cues~\cite{shih2015reflection} or adopt learning-based approaches to capture the prior of natural images~\cite{fan2017generic, jin2018learning, wei2019single, yang2018seeing, zhang2018single}. 
While significant advances have been shown, separating the clean background from reflection/occlusions is fundamentally ill-posed and often requires a high-level semantic understanding of the scene. %
In particular, the performance of learning-based methods often degrades significantly for out-of-distribution images. 

To tackle these challenges, \emph{multi-frame approaches} exploit the fact that the background scene and the occluding elements are located at different depths with respect to the camera (e.g., virtual depth of window reflections). 
Consequently, taking multiple images from a slightly moving camera reveals the motion differences between the two layers~\cite{szeliski2000layer, be2008blind, liu2008sift, gai2009blind, li2013exploiting, guo2014robust}. 
A number of approaches exploit such visual cues for reflection or fence removal from a video~\cite{szeliski2000layer, liu2008sift, be2008blind, gai2009blind, sinha2012image, li2013exploiting, guo2014robust, nandoriya2017video, du2018accurate, alayrac2019visual}.
Xue et al.~\cite{xue2015computational} propose a unified computational framework for obstruction removal and show impressive results on several real input sequences.
The formulation, however, requires a computationally expensive optimization process and relies on strict assumptions of brightness constancy and accurate dense motion estimation.
To alleviate these issues, recent work~\cite{alayrac2019visual} explores model-free methods by learning a generic 3D convolutional neural network (CNN).
Nevertheless, the CNN-based methods do not produce results with comparable quality as optimization-based algorithms on real input sequences.

In this work, we propose a multi-frame obstruction removal algorithm that exploits the strength of both optimization-based and learning-based methods. 
Inspired by the optimization-based approaches~\cite{nandoriya2017video, xue2015computational}, our algorithm alternates between the dense motion estimation and  background/obstruction layer reconstruction steps in a coarse-to-fine manner.
Our framework builds upon the optimization-based formulation of \cite{nandoriya2017video, xue2015computational} but differs in that our model is purely data-driven and does not rely on classical assumptions such as brightness constancy~\cite{nandoriya2017video, xue2015computational}, accurate flow fields~\cite{li2013exploiting}, or planar surface~\cite{guo2014robust} in the scene.
When these assumptions do not hold (e.g., occlusion/dis-occlusion, motion blur, inaccurate flow), classical approaches may fail to reconstruct clear foreground and background layers.

On the other hand, data-driven approaches learn from diverse training data and can tolerate errors when these assumptions are violated.
The explicit modeling of dense motion within each layer facilitates us to progressively recover detailed content in the respective layers.
Instead of relying on hand-crafted objectives for recovering these layers, we use the learning-based method for fusing flow-warped images to accommodate potential violations of brightness constancy and errors in flow estimation.
We train our fusion network using a synthetically generated dataset and demonstrate that it performs well to unseen real-world sequences.
In addition, we present an online optimization process to further improve the visual quality of particular testing sequences.
We show that the proposed method performs favorably against existing learning-based and optimization-based algorithms on a wide variety of challenging sequences and applications.

\revision{
The preliminary version of this work has been published in CVPR 2020~\cite{liu2020learning}.
In this paper, we further improve our method in three key aspects. 
\begin{enumerate}
\item We present an improved layer reconstruction model that allows us to take an arbitrary number of input frames.
\item We apply meta-learning to facilitate the efficient adaptation of our pre-trained model to a particular testing sequence. Our results show improvement for both the runtime speed and visual quality.
\item We incorporate a realistic reflection image synthesis model \cite{zhang2018single} and extend it with a variety of data augmentation to generate more realistic and diverse training sequences. 
\end{enumerate}
We show extensive experimental results to validate our design choices. Experiments show that our improved method significantly outperforms our CVPR work on both quantitative and qualitative evaluations.
}

The main contributions of this work are:
\begin{enumerate}
\item We integrate the optimization-based formulation into a learning-based method for robustly separating background/obstruction layers. \revision{Meta-learning is employed to reduce the runtime.}
\item We present a transfer learning strategy that first pre-trains the model using synthetic data and then fine-tunes on real sequence with an unsupervised optimization objective function to achieve state-of-the-art performance \revision{in the context of obstruction removal}. 
\item \revision{We show our model can be easily extended to handle other types of obstruction removal problems, e.g., fence and \revision{adherent raindrop} removal}.

\end{enumerate}

%% file: 2_related-work.tex
\section{Related Work}
\label{sec:RelatedWork}

\heading{Multi-frame reflection removal.}
Existing methods often exploit the differences of motion patterns between the background and reflection layers~\cite{guo2014robust,xue2015computational}, and impose natural image priors~\cite{gai2011blind,guo2014robust,xue2015computational}.
These methods differ in modeling the motion fields, e.g., SIFT flow~\cite{li2013exploiting}, homography~\cite{guo2014robust}, and dense optical flow~\cite{xue2015computational}.
Recent advances include optimizing temporal coherence~\cite{nandoriya2017video} and learning-based layer decomposition~\cite{alayrac2019visual} for reflection removal.
In contrast to the scheme based on a generic spatio-temporal CNN~\cite{alayrac2019visual}, our method explicitly models the dense flow fields of the background and obstruction layers to obtain cleaner results on real sequences.

\vspace{\paramargin}
{\flushleft {\bf Single-image reflection removal.}}
A number of approaches have been proposed to remove unwanted reflections with only \emph{one single image} as input.
Existing methods exploit various cues, including ghosting effect~\cite{shih2015reflection}, blurriness caused by depth-of-field~\cite{li2014single,wan2016depth}, image priors (either hand-designed~\cite{arvanitopoulos2017single} or learned from data~\cite{yang2018seeing, zhang2018single}), and the defocus-disparity cues from dual pixel sensors~\cite{punnappurath2019reflection}.
Despite the demonstrated success, reflection removal from a single image remains challenging due to the nature of this highly ill-posed problem and the lack of motion cues.
Our work instead utilizes the motion cues from image sequences captured with a slightly moving camera for separating the background and reflection layers.

\vspace{\paramargin}
{\flushleft {\bf Occlusion and fence removal.}}
Occlusion removal aims to eliminate the captured obstructions, e.g., fence or \revision{adherent raindrops} on an image or sequences, and provide a clear view of the scene.
Existing methods detect fence patterns by exploiting visual parallax~\cite{mu2013video}, dense flow field~\cite{xue2015computational}, disparity maps~\cite{jonna2017stereo}, or using graph-cut~\cite{yi2016automatic}.
One recent work leverages a CNN for fence segmentation~\cite{du2018accurate} and recovers the occluded pixels using optical flow.
Our method also learns deep CNNs for optical flow estimation and background image reconstruction.
Instead of focusing on fence removal, our formulation is more general and applicable to different obstruction removal tasks.

\vspace{\paramargin}
{\flushleft {\bf Layer decomposition.}}
Image layer decomposition is a long-standing problem in computer vision, e.g., intrinsic image~\cite{bell2014intrinsic,zhou2015learning}, depth, normal estimation~\cite{eigen2014depth,jeon2014intrinsic},  relighting~\cite{eisemann2004flash,nestmeyer2020learning}, and inverse rendering~\cite{li2019inverse, sengupta2019neural}.
Our method is inspired by the development of these layer decomposition approaches, particularly in the ways of leveraging both the physical image formation constraints and data-driven priors. 

\heading{Video completion}
Video completion aims to fill in plausible content in missing regions of a video~\cite{ilan2015survey}, with applications ranging from object removal, full-frame video stabilization, and watermark/transcript removal.
State-of-the-art methods estimate the flow fields in both known and missing regions to constrain the content synthesis~\cite{huang2016temporally,xu2019deep,gao2020deep}, and generate temporally coherent completion.
The obstruction removal problem resembles a video completion task.
However, the crucial difference is that no manual mask selection is required for removing the fences/obstructions from videos.

\begin{figure*}[t]
  \begin{center}
  \includegraphics[width=1.0\linewidth]{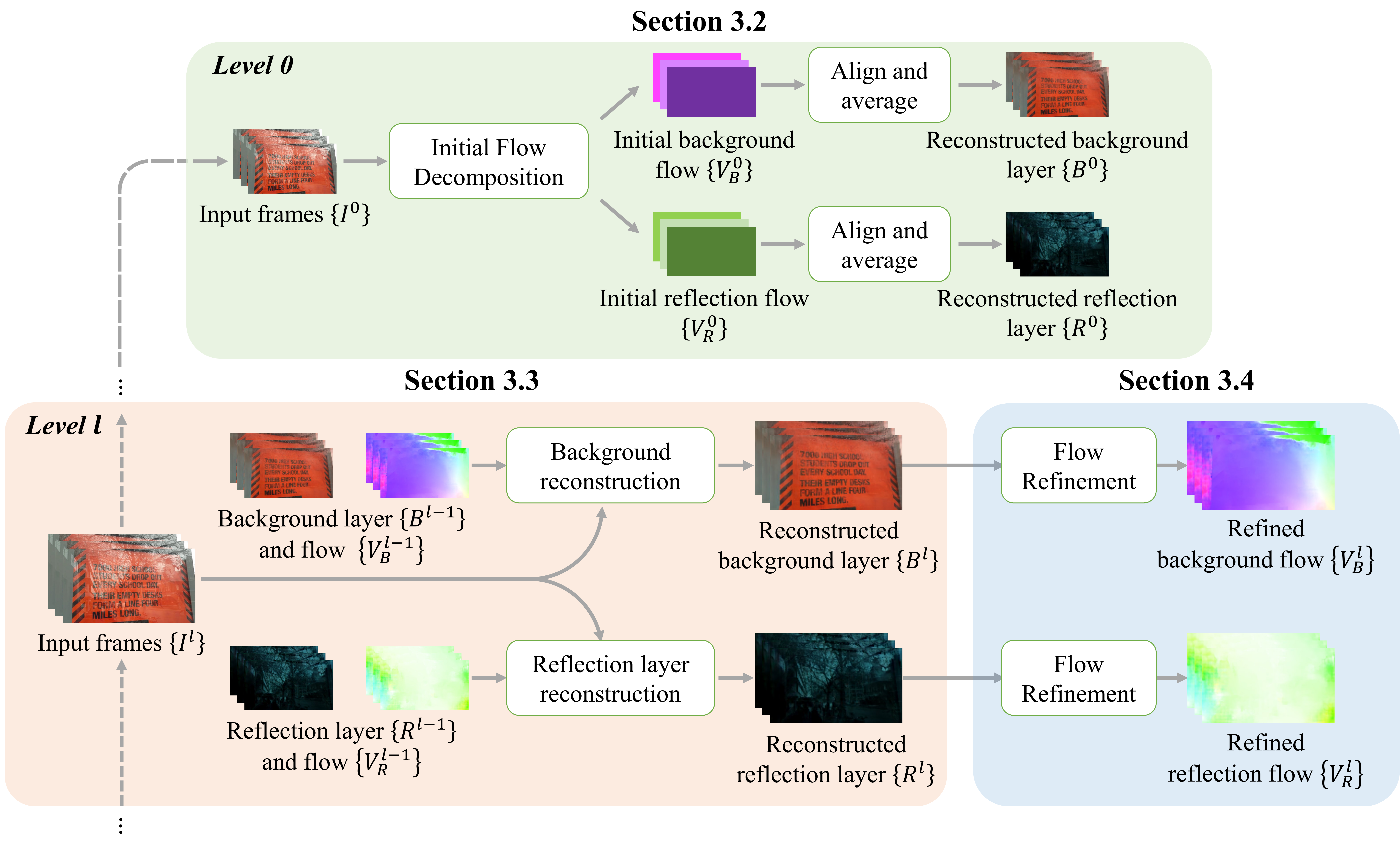}
  \caption{\textbf{Algorithmic overview.} 
    We reconstruct the background/reflection layers in a coarse-to-fine manner.
    At the coarsest level, we estimate uniform flow fields for both the background and reflection layers, and reconstruct coarse background/reflection layers by averaging the aligned frames.
    At level $l$, we apply (1) the background/reflection reconstruction modules to decompose an image, and (2) the PWC-Net to predict the refined flow fields for both layers.
    Our framework progressively reconstructs the background/reflection layers and flow fields until the finest level.
  }
  \label{Fig:archi}
  \end{center}
\end{figure*}

\heading{Online optimization (training on testing data)}
Learning from the test data has been an effective way to reduce the domain discrepancy between the training/testing distributions.
Examples abound, including using geometric constraints~\cite{chen2019self,Luo-VideoDepth-2020}, self-supervised losses~\cite{sun2019test}, deep image priors~\cite{ulyanov2018deep,gandelsman2019double,Chen2020nasdip}, and online template update~\cite{kalal2011tracking}.
Similar to these methods, we fine-tune our background/obstruction reconstruction network on a particular test sequence to further improve the separation. 
Our unsupervised loss directly measures how well the recovered background/obstruction and the dense flow fields explain all the input frames.

\heading{Meta-learning.}
Meta-learning refers to a class of algorithms that aim to learn a ``learner" which can quickly adapt to a new task with few training examples.
Existing meta-learning algorithms can be categorized as black-box adaptation~\cite{hochreiter2001learning,ravi2016optimization}, metric-based~\cite{vinyals2016matching,snell2017prototypical,garcia2018few,tseng2020cross}, and optimization-based methods~\cite{finn2017model,nichol2018reptile}.
Our work applies an optimization-based meta-learning algorithm~\cite{finn2017model,nichol2018reptile,zintgraf2019fast} for learning a weight initialization that can adapt to a new task with few gradient updates from a small number of training examples. 
Specifically, we apply Reptile~\cite{nichol2018reptile} to improve the model adaptation to the testing sequence.

%% file: 3_method.tex
\section{Proposed Algorithm}

Given a sequence  \revision{$\{I_k\}^T_{k=1}$} of $T$ frames, our goal is to decompose each frame $I_k$ into two layers, one for the target (clean) background and the other for the obstruction caused by reflection/fence/raindrops.
Decomposing an image sequence into background and obstruction is difficult as it involves solving two tightly coupled problems: motion decomposition and layer reconstruction.
Without an accurate motion decomposition, the layers cannot be reconstructed faithfully due to the misalignment from inaccurate motion estimation (e.g., optical flow).
On the other hand, without well-reconstructed background and obstruction layers, the motion cannot be accurately estimated because of the mixed contents.
Due to the nature of this chicken-and-egg problem, there is no ground to start with because we do not have information for both motion and layer content.

\subsection{Algorithmic overview}

In this work, we propose to learn deep CNNs to tackle the above-mentioned challenges.
Our proposed method mainly consists of three modules: 
1) initial flow decomposition, 
2) background and obstruction layer reconstruction, and 
3) optical flow refinement.
Our method takes $T$ frames as input and decomposes the keyframe frame $I_k$ into a background layer $B_k$ and reflection layer $R_k$ at a time.
We reconstruct the output images in a coarse-to-fine manner within an $L$-level hierarchy.
First, we estimate the flows at the coarsest level from the initial flow decomposition module (Section~\ref{initial_flow_decomposition}).
Next, we progressively reconstruct the background/obstruction layers (Section~\ref{background_reflection_layer_reconstruction}) and refine optical flows (Section~\ref{optical_flow_refinement}) until the finest level.
Figure~\ref{Fig:archi} shows an overview of our method.
Our framework can be applied to several layer decomposition problems, such as reflection/obstruction/fence/rain removal.
Without loss of generality, we use the reflection removal task as an example to introduce our algorithm. 
We describe the details of the three modules in the following sections.

\subsection{Initial flow decomposition}
\label{initial_flow_decomposition}

We first predict optical flows for both the background and reflection layers at the coarsest level ($l = 0$), which is the essential starting point of our algorithm.
Instead of estimating dense flow fields, we propose to learn a \emph{uniform} motion vector for each layer.
Our initial flow decomposition network consists of two sub-modules: 1) a feature extractor, and 2) a layer flow estimator.
The feature extractor first generates feature maps for all the input frames at a $1/{2^L}\times$ spatial resolution.
We then construct a cost volume between frame $j$ and frame $k$ via a correlation layer~\cite{sun2018pwc}:
\begin{equation} \label{eq:cost_volume}
    CV_{jk}(\mathbf{x_1}, \mathbf{x_2}) = c_j(\mathbf{x_1})^\top c_k(\mathbf{x_2}),
\end{equation}
where $c_j$ and $c_k$ are the extracted features of frame $j$ and $k$, respectively, and $\mathbf{x}$ indicates the pixel index.  
Since the spatial resolution is quite small at this level, we set the correlation layer's search range to only 4 pixels.
The cost volume $CV$ is then concatenated with the feature $c_j$ and fed into the layer flow estimator.

The layer flow estimator uses the global average pooling and fully-connected layers to generate two global motion vectors.
Next, we tile the global motion vectors into two uniform flow fields (at a $1/{2^L}\times$ spatial resolution):
$\{V^0_{B, j\rightarrow k}\}$ for the background layer
and 
$\{V^0_{R, j\rightarrow k}\}$ for the reflection layer.
We provide the detailed architecture of our initial flow decomposition module in the supplementary material.

\begin{figure*}
    \centering
    \includegraphics[width=1.0\linewidth]{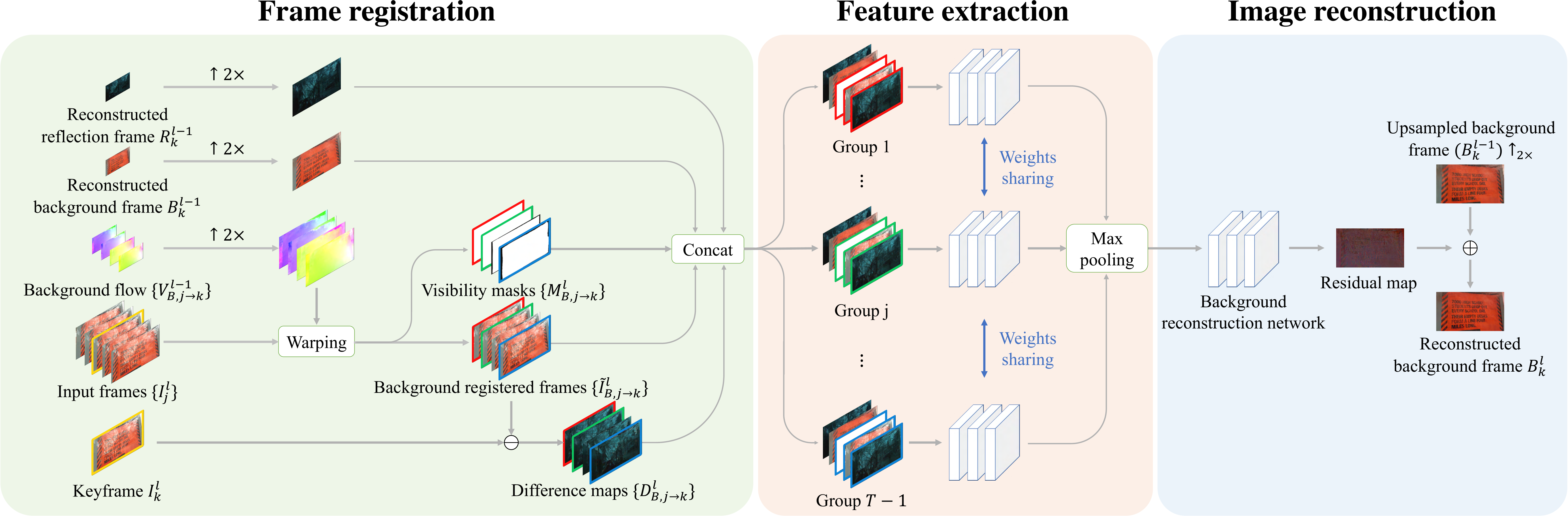}
    \caption{
        \textbf{Overview of layer reconstruction module.}
        At level $l$, we first upsample the background flows $\{V^{l-1}_{B, j\rightarrow k}\}$ from level $l - 1$ to warp and align the input frames \revision{$\{I^l_j\}$} with the keyframe $I^l_k$.
        We then compute the difference maps between the background-registered frames and the keyframe.
        For each non-keyframe image, we group the following five frames as a group: 
        (1) background-registered frames $\{\tilde{I}_{B, j \rightarrow k}^l\}$, 
        (2) difference maps $\{D^l_{B,j \rightarrow k}\}$, 
        (3) visibility masks $\{M^l_{B,j \rightarrow k}\}$, 
        (4) upsampled background $(B^{l-1}_{k})\uparrow_2$, and 
        (5) reflection layers $(R^{l-1}_{k})\uparrow_2$.
        After collecting these $T-1$ groups, where $T$ is the number of input frames, we apply convolutional layers (with weights sharing) to extract the features from each group.
        We then apply a max operation to collapse these groups into one feature map.
        The background reconstruction network takes the collapsed feature as input, and learns to predict the residual map of the background keyframe.
        We add the predicted residual map to the upsampled background frame $(B^{l-1}_{k})\uparrow_2$ and generate the reconstructed background frame $B^{l}_{k}$ at level $l$.
        For the reflection layer reconstruction, we use the same architecture but learn a different set of network parameters.
        Thanks to the use of max operation, our layer reconstruction module is able to take an arbitrary number of input frames.
    }
    \label{fig:imageRecons}
\end{figure*}

\subsection{Background/Reflection layer reconstruction}
\label{background_reflection_layer_reconstruction}

The layer reconstruction module aims to reconstruct a clean background image $B_k$ and a reflection image $R_k$. 
Although the goals of these two tasks are similar in spirit, the characteristics of the background and reflection layers are essentially different.
For example, the background layers are often more dominant in appearance but could be occluded in some frames.
On the other hand, the reflection layers are often blurry and darker.
Consequently, we train two independent networks for reconstructing the background and reflection layers.
These two models have the same architecture but do not share the network parameters.
In the following, we only describe the details for background layer reconstruction; the reflection layer is reconstructed in a similar fashion.

We reconstruct the background layer in a coarse-to-fine fashion.
At the coarsest level ($l = 0$), we first use the flow fields estimated from the initial flow decomposition module to align the neighboring frames.
Then, we compute the average of all the background-registered frames as the predicted background image:
\begin{equation} \label{eq:init_B}
     B^0_k = \frac{1}{T} \sum_{j=1}^{T} \boldsymbol{W}(I^0_j, V^0_{B, j\rightarrow k}),
\end{equation}
where $I^0_j$ is the $1/2^L \times$ downsampled frame $j$, and $\boldsymbol{W}()$ is the warping operation with bilinear sampling.

At the $l$-th level, the network takes as input the reconstructed background image $B_k^{l-1}$, reflection image $R_k^{l-1}$, background optical flows $\{V^{l-1}_{B, k\rightarrow j}\}$ from the previous level as well as the input frames $\{I^l_t\}$ at the current level.
To reconstruct the background image of the keyframe $B_k^{l}$ at the current level, the background reconstruction module consists of three steps: 1) frame registration, 2) feature extraction, and 3) image reconstruction. 

\heading{Frame registration.}
We first upsample the background flow fields $\{V^{l-1}_{B, k\rightarrow j}\}$ by $2\times$ and align all the input frames $\{I_j^l\}$ to the keyframe $\{I_k^l\}$:
\begin{equation} \label{eq:warp}
     \tilde{I}^l_{B, j\rightarrow k} = \boldsymbol{W}( I^l_j, (V^{l-1}_{B, j\rightarrow k})\!\uparrow_2 ),
\end{equation}
where $()\!\uparrow_2$ denotes the $2\times$ bilinear upsampling operator.

\heading{Feature extraction.}
For each non-keyframe image $B^l_{j, j \neq k}$, we first concatenate the following five features into a group: 
\begin{compactenum}
\item background-registered frame $\tilde{I}_{B, j \rightarrow k}^l$, 
\item difference map $D^l_{B,j\rightarrow k} = | I^l_{B, j\rightarrow k} - I^l_k |$, 
\item visibility mask $M^l_{B,j \rightarrow k}$, 
\item upsampled background $(B^{l-1}_{k})\!\uparrow_2$, and 
\item upsampled reflection layers $(R^{l-1}_{k})\!\uparrow_2$.
\end{compactenum}
The visibility mask $M^l_{B,j \rightarrow k}$ is computed by warping the grid coordinates with the flow $V^{l-1}_{B, j\rightarrow k}$ and checking whether each pixel stays within the image boundary.

For $T$ input frames, we can construct $T - 1$ groups.
We then apply 5 convolutional layers to extract features from each group:
\begin{align} \label{eq:background_image_feature_extraction}
     {\theta}^l_{j\rightarrow k} = g_{\theta}  \Big( & \{\tilde{I}_{B, j \rightarrow k}^l\}, \{D^l_{B,j \rightarrow k}\}, \{M^l_{B,j \rightarrow k}\}, (B^{l-1}_{k})\!\uparrow_2, \nonumber \\
     & (R^{l-1}_{k})\!\uparrow_2  \Big),
\end{align}
where $g_{\theta}$ is the feature extraction network.
We use a max pooling layer to collapse these $T - 1$ groups of features into one representative feature.
Note that the weights in $g_{\theta}$ are shared among all the groups.
By using the weight sharing and max pooling, our layer reconstruction module is capable of processing arbitrary numbers of input frames.

\heading{Image reconstruction.}
Our background reconstruction network takes the collapsed feature as input and learns to predict the residual map of the background keyframe.
The background frame $B^l_k$ is reconstructed by:
\begin{align} \label{eq:background_image_reconstruction}
     B^l_k = g_B \Big( \max_{j=1, j\neq k}^{T}({\theta}^l_{j\rightarrow k})  \Big) + (B^{l-1}_{k})\uparrow_2,
\end{align}
where $g_B$ is the background reconstruction network.

Note that the reflection layer is also involved in the reconstruction of the background layer, which couples the background and reflection reconstruction networks together for joint training.
Figure~\ref{fig:imageRecons} illustrates an overview of the background reconstruction network at the $l$-th level.

\subsection{Optical flow refinement}
\label{optical_flow_refinement}

After reconstructing all the background images $\{B^l\}$ at level $l$, we then refine the background optical flows.
We use the pre-trained PWC-Net~\cite{sun2018pwc} to estimate the flow fields between a paired of background images:
\begin{equation} \label{eq:pwcnet1}
    V^l_{B, j\rightarrow k} = \text{PWC}(B^l_j, B^l_k),
\end{equation}
where $\text{PWC}$ denotes the pre-trained PWC-Net.
Note that the PWC-Net is fixed and not updated with the other sub-modules of our model during the training stage.

\subsection{Network training}

To improve training stability, we employ a two-stage training procedure.
At the first stage, we train the initial flow decomposition network with the following loss:
\begin{equation} \label{eq:decomposition_loss}
\begin{split}
    \mathcal{L}_{\text{dec}} = \sum_{k=1}^{T} \sum_{j=1, j\neq k}^{T} &\|V^0_{B, j\rightarrow k} - \text{PWC}(\hat{B}_j, \hat{B}_k){\downarrow}^{2^L} \|_1 + \\
    &\|V^0_{R, j\rightarrow k} - \text{PWC}(\hat{R}_j, \hat{R}_k){\downarrow}^{2^L} \|_1\,,
\end{split}
\end{equation}
where ${\downarrow}$ is the bilinear downsampling operator, $\hat{B}$ and $\hat{R}$ denote the ground-truth background and reflection layers, respectively. 
We use the pre-trained PWC-Net to compute optical flows and downsample the flows by $2^L\times$ as the pseudo ground-truth flows to train the initial flow decomposition network.

Next, we freeze the initial flow decomposition network and train the layer reconstruction networks with an image reconstruction loss:
\begin{equation} \label{eq:l1_loss}
    \mathcal{L}_{\text{img}} =  \frac{1}{T\!\times\!L} \sum_{t=1}^{T} \sum_{l=0}^{L}  ( \|\hat{B}^l_t - B^l_t\|_1 + \|\hat{R}^l_t - R^l_t\|_1),
\end{equation}
and a gradient loss:
\begin{equation} \label{eq:grdient_loss}
    \mathcal{L}_{\text{grad}} = \frac{1}{T\!\times\!L }\sum_{t=1}^{T} \sum_{l=0}^{L}  ( \|\nabla\hat{B}^l_t - \nabla B^l_t\|_1 + \|\nabla\hat{R}^l_t - \nabla R^l_t\|_1),
\end{equation}
where $\nabla$ is the spatial gradient operator.
The gradient loss encourages the network to reconstruct faithful edges to further improve visual quality.
The overall loss for training the layer reconstruction networks is:
\begin{equation}
    \mathcal{L}_{\text{supervised}} = \mathcal{L}_{\text{img}} + \lambda_{\text{grad}} \mathcal{L}_{\text{grad}},
    \label{eq:validation}
\end{equation}
where the weight $\lambda_{\text{grad}}$ is empirically set to 1 in all our experiments.
We train both the initial flow decomposition and layer reconstruction networks with the Adam optimizer~\cite{kingma2014adam} with a batch size of 2.
We set the learning rate to $10^{-4}$ for the first 100K iterations and then decrease to $10^{-5}$ for another 100K iterations.
The number of pyramid levels $L$ is set to 5.
We describe the training steps of our two-stage training strategy Algorithm~\ref{two_stage}.

\renewcommand{\algorithmicrequire}{\textbf{Input:}}  %
\renewcommand{\algorithmicensure}{\textbf{Output:}} %
\begin{algorithm}[t]
\caption{Pre-training}\label{two_stage}
\begin{algorithmic}[1]
\Ensure Initial flow decomposition network $\Theta_F$, background reconstruction network $\Theta_B$, and reflection reconstruction network $\Theta_R$
\State \% Stage 1.
\State Randomly initialize $\Theta_F$, $\Theta_B$, and $\Theta_R$.
\While {iterations $k <$ 100K}
\State Update $\Theta_F$ with loss function $\mathcal{L}_{dec}$ in~\eqref{eq:decomposition_loss}.
\EndWhile
\State \% Stage 2.
\State Fix the weights of $\Theta_F$.
\While {iterations $k <$ 200K}
\State Update $\Theta_B$ and $\Theta_R$ for all pyramid levels with loss function in~\eqref{eq:validation}.
\EndWhile
\end{algorithmic}
\end{algorithm}

\subsection{Meta-learning for fast adaptation}
To ensure that our model can be adapted to handle real data more effectively and efficiently, we apply a meta-learning technique to finetune our pre-trained model with both the synthetic and real sequences.
Specifically, we use the first-order meta-learning algorithm~\cite{nichol2018first} and describe our meta training step in Algorithm~\ref{meta_algorithm}.
When a training batch is sampled from synthetic data, we minimize the supervised loss~\eqref{eq:validation}.
On the other hand, when a training batch is sampled from real data, we optimize a warping consistency loss: 
\begin{equation} \label{eq:warp_loss}
\begin{split}
    \mathcal{L}_{\text{warp}} = \sum_{k=1}^{T} \sum_{\substack{j=1\\j \ne k}}^{T} \sum_{l=0}^{L} \|I^l_j - \tilde{I}^l_j\|_1,
\end{split}
\end{equation}
where $\tilde{I}^l_j = \boldsymbol{W}(B^l_k, V^l_{B, j\rightarrow k}) + \boldsymbol{W}(R^l_k, V^l_{R, j\rightarrow k})$ is the reconstructed input frame from the warped background and reflection layers.
The warping consistency loss enhances fidelity by enforcing that the predicted background and reflection layers should be warped back and composited into the original input frames.
In addition, we also incorporate the total variation loss:
\begin{equation} \label{eq:tv_loss}
\begin{split}
    \mathcal{L}_{\text{tv}} =  \sum_{t=1}^{T} \sum_{l=0}^{L}  (\|\nabla B^l_t\|_1 + \|\nabla R^l_t\|_1),
\end{split}
\end{equation}
which encourages the network to generate natural images by following the sparse gradient image prior.
The overall unsupervised loss for training on real data is defined as:
\begin{equation} \label{eq:unsupervised_loss}
\begin{split}
    \mathcal{L}_{\text{unsupervised}} = \mathcal{L}_{\text{warp}} + \lambda_{\text{tv}} \mathcal{L}_{\text{tv}},
\end{split}
\end{equation}
where the weight $\lambda_{\text{tv}}$ is empirically set to 0.1 in all our experiments.
The update parameter $\epsilon$ is set to $0.1$ in our experiment.
We show in Section~\ref{sec:analysis} that the meta-learning is able to speed up the online optimization as well as improve the reconstruction performance.

\begin{algorithm}[t]
\caption{Meta-learning training with Reptile~\cite{nichol2018reptile}}
\label{meta_algorithm}
\begin{algorithmic}[1]
\Require Pre-trained initial flow decomposition network $\Theta_F$, background reconstruction network $\Theta_B$, and reflection reconstruction network $\Theta_R$.
\Ensure Updated background reconstruction network $\Theta_B^\prime$ and reflection reconstruction network $\Theta_R^\prime$.
\State Fix the weights of $\Theta_F$.
\State Initialize $\Theta_B^\prime \leftarrow \Theta_B$.
\State Initialize $\Theta_R^\prime \leftarrow \Theta_R$.
\While {iterations $k < $ 100K}
\State Randomly sample a training mini-batch, denoted as task $\tau$.
\If {$\tau$ is sampled from synthetic data}
\State Get the updated weights $\Theta_B^{\tau}$ and $\Theta_R^{\tau}$ by minimizing the supervised loss function~\eqref{eq:validation}.
\Else
\State Get the updated weights $\Theta_B^{\tau}$ and $\Theta_R^{\tau}$ by minimizing the unsupervised loss function~\eqref{eq:unsupervised_loss}.
\EndIf
\State Update $\Theta_B^\prime \leftarrow \Theta_B^\prime + \epsilon(\Theta_B^{\tau} - \Theta_B^\prime)$
\State Update $\Theta_R^\prime \leftarrow \Theta_R^\prime + \epsilon(\Theta_R^{\tau} - \Theta_R^\prime)$
\EndWhile
\end{algorithmic}
\end{algorithm}

\subsection{Online optimization}

We adopt an online refinement method to fine-tune our pre-trained model with a real test sequence by optimizing the unsupervised loss in~\eqref{eq:unsupervised_loss}.
Note that we freeze the weights of the PWC-Net and only update the background/reflection layer reconstruction modules in both the meta-learning and online optimization stages.
We fine-tune our model on every single input sequence for 200 iterations. 
The fine-tuning step takes about 3 minutes for a sequence with a 1296 $\times$ 864 spatial resolution. 
We describe the training steps of our unsupervised online optimization in Algorithm~\ref{online_optimization}.

\begin{algorithm}[t]
\caption{Online optimization}\label{online_optimization}
\begin{algorithmic}[1]
\Require Pre-trained initial flow decomposition network $\Theta_F$, updated background reconstruction network $\Theta_B^\prime$, and reflection reconstruction network $\Theta_R^\prime$.
\Ensure Fine-tuned background reconstruction network $\Theta_B^{\prime\prime}$ and reflection reconstruction network $\Theta_R^{\prime\prime}$.
\State Fix the weights of $\Theta_F$.
\State Initialize $\Theta_B^\prime \leftarrow \Theta_B$.
\State Initialize $\Theta_R^\prime \leftarrow \Theta_R$.
\While {iterations $k < 200$}
\State Update $\Theta_B$ and $\Theta_R$ with unsupervised loss function $\mathcal{L}_{\text{online}}$ in~\eqref{eq:unsupervised_loss}. %
\EndWhile
\end{algorithmic}
\end{algorithm}

\subsection{Extension to other obstruction removal tasks}

\begin{figure*}
    \centering
    \includegraphics[width=0.8\textwidth]{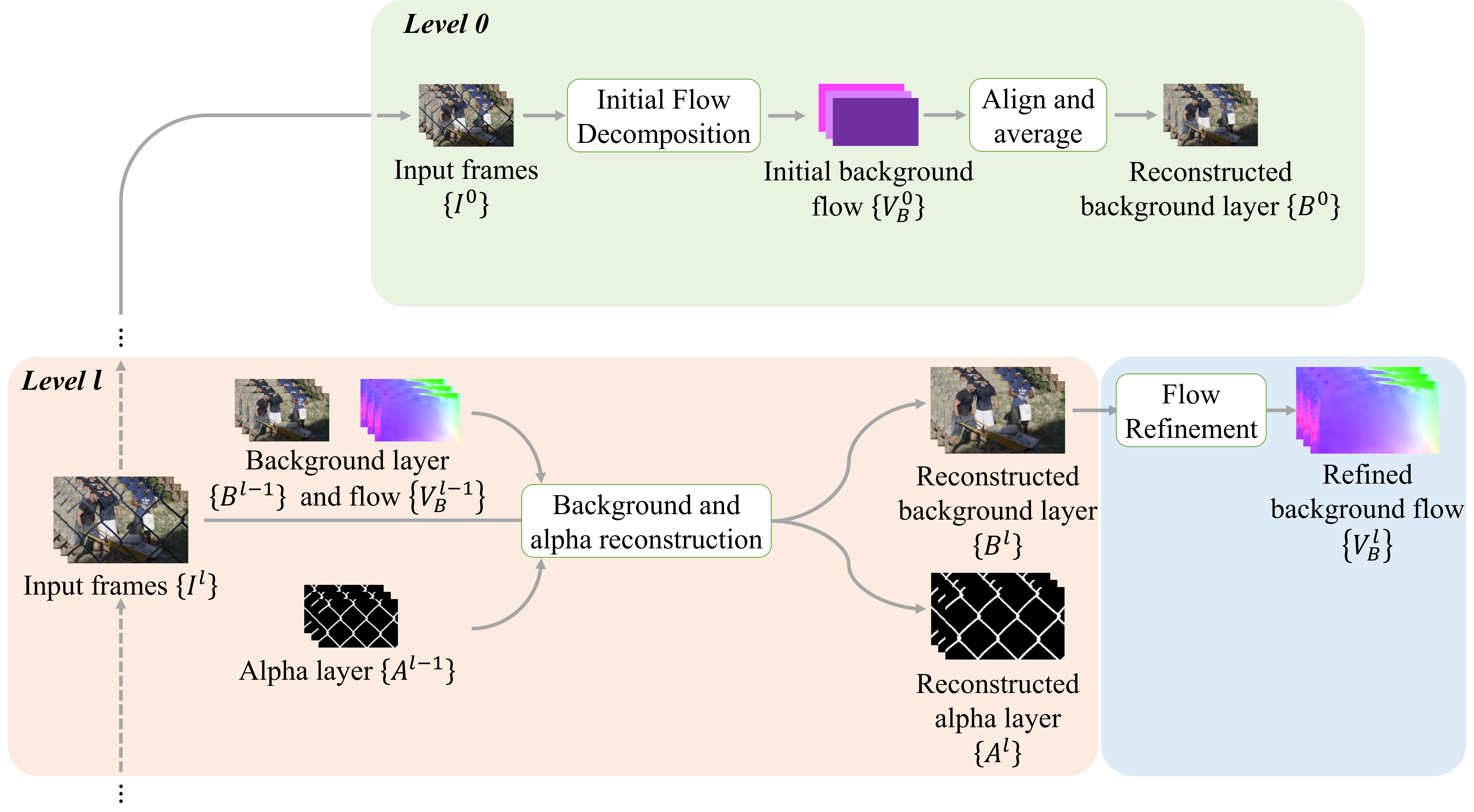}
    \caption{
    \revision{\textbf{Overview of the fence removal task.}}
    }
  \label{fig:minimum_design}
\end{figure*}

Our proposed framework can be easily extended to handle other obstruction removal tasks, such as fence and \revision{adherent raindrop} removal.
First, we remove the reflection layer reconstruction module and only predict the background layers.
Second, the background image reconstruction network outputs an additional channel as the alpha map for segmenting the obstruction layer.
We do not estimate flow fields for the obstruction layer as the flow estimation network cannot handle the repetitive structures (e.g., fence) or tiny objects (e.g., raindrops) well and often predicts noisy results.
With such a design change, our model is able to perform well on the fence and \revision{adherent raindrop} removal tasks.
We use the fence segmentation dataset~\cite{du2018accurate} and alpha matting dataset~\cite{xu2017deep} to generate training data for both tasks.
\revision{\figref{minimum_design} gives an overview of adapting our framework to the fence removal task.}

To apply our online optimization for obstruction removal, we first extract the foreground layer by $F_k^l = I^l_k \cdot A^l_k$, where $A$ denotes the predicted alpha map.
Then, we compute the foreground flow $V^l_{F, j\rightarrow k}$ with the pre-trained PWC-Net.
The reconstructed frame $\tilde{I}_j^l$ can be approximated by:
\begin{equation}
\begin{split}
    \tilde{I}_j^l &= \boldsymbol{W}(F^l_k, V^l_{F, j\rightarrow k}) \\
    &+ \boldsymbol{W}(1-A^l_k, V^l_{F, j\rightarrow k}) \cdot \boldsymbol{W}(B^l_k, V^l_{B, j\rightarrow k}).
\end{split}
\end{equation}
We replace $\tilde{I}_j^l$ in~\eqref{eq:warp_loss} as the warping consistency loss used in the meta-learning and online optimization stages for fence removal.

\subsection{Synthetic sequence generation}
\label{sec:data_generation}

Since it is difficult to collect real sequences with ground-truth reflection and background layers, we use the Vimeo-90k dataset~\cite{xue2019video} to generate synthetic sequences for training.
Out of the 91,701 sequences in the Vimeo-90k training set, we randomly select two sequences as the background and reflection layers.
In our preliminary work~\cite{liu2020learning}, we adopt the following three steps to generate a synthetic sequence.
First, we warp the sequences using random homography transformations.
We then randomly crop the sequences to a spatial resolution of $320\times192$ pixels.
The composition is applied frame by frame using the realistic reflection image synthesis model proposed by previous work~\cite{fan2017generic, zhang2018single}.

In this work, we improve the data synthesis pipeline in~\cite{liu2020learning} to generate more diverse training data.
During the training stage, we apply on-the-fly random color augmentation, including hue, saturation, brightness, and contrast, on both background and reflection layers.
\revision{As suggested by previous work~\cite{fan2017generic, zhang2018single}}, to simulate the effect that the reflection layer is usually out-of-focus, we apply a Gaussian filter on the reflection layer with kernel size randomly selected from [3, 17] and standard deviation randomly sampled from [0.8, 2.9].

After blending the background and reflection layers, we apply Gaussian noise with standard deviation randomly selected from $[0, 0.02]$ and JPEG compression artifacts with compression quality randomly selected from $[50, 100]$.
\revision{In addition, for simulating vignetting, we add a Gaussian falloff with a randomly selected kernel size to the synthetic image.}
As our model is able to tackle arbitrary input frames, we randomly sample 2 to 7 input frames at each training iteration.
We also provide examples of the training pairs generated from our pipeline in the supplementary materials. %

%% file: 4_experiment.tex
\section{Experimental Results}
\label{sec:experiments}

\begin{figure}
\centering
\footnotesize
\renewcommand{\tabcolsep}{1pt} %
\renewcommand{\arraystretch}{1} %
\begin{tabular}{cccc}
    \raisebox{2.7\normalbaselineskip}[0pt][0pt]{\rotatebox[origin=c]{90}{Stone}} &  
    \includegraphics[width=0.31\linewidth]{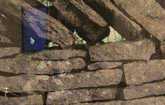} & \includegraphics[width=0.31\linewidth]{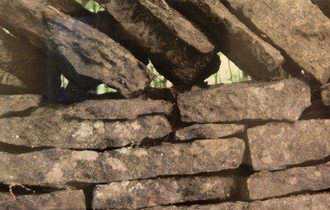} & \includegraphics[width=0.31\linewidth]{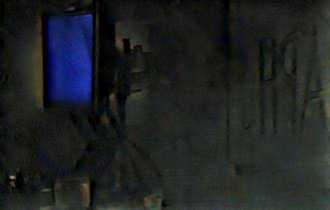} \\
    \raisebox{2.9\normalbaselineskip}[0pt][0pt]{\rotatebox[origin=c]{90}{Toy}} &  
    \includegraphics[width=0.31\linewidth]{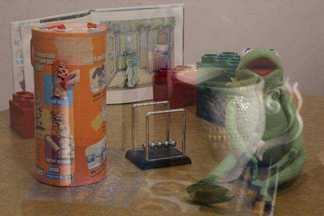} & \includegraphics[width=0.31\linewidth]{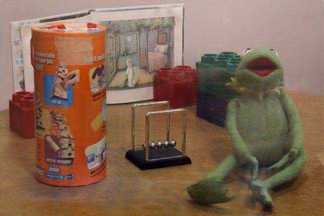} & \includegraphics[width=0.31\linewidth]{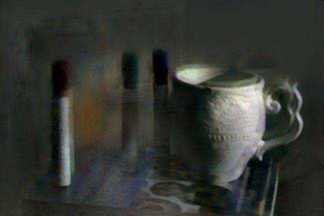} \\
    \raisebox{2.4\normalbaselineskip}[0pt][0pt]{\rotatebox[origin=c]{90}{Hanoi}} &  
    \includegraphics[width=0.31\linewidth]{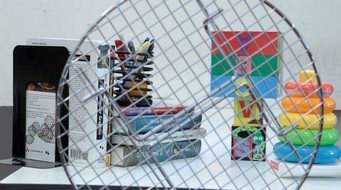} & \includegraphics[width=0.31\linewidth]{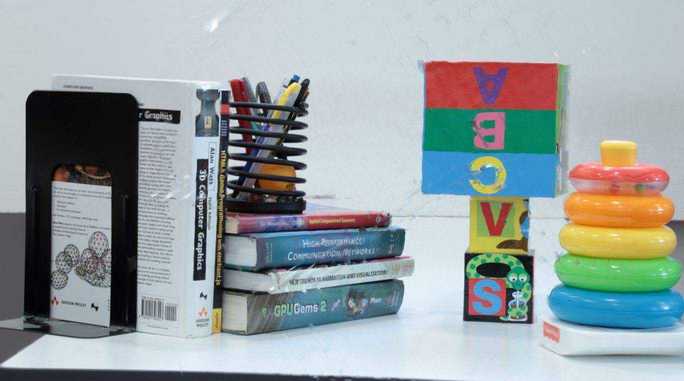} & \includegraphics[width=0.31\linewidth]{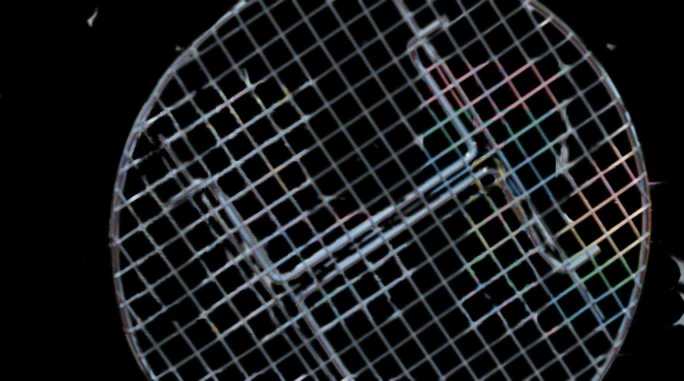} \\
    & Input (keyframe) & Recovered background & Recovered obstruction \\
\end{tabular}
\caption{\textbf{Visual results on controlled sequences~\cite{xue2015computational}.} For each sequence, we show the keyframe (\emph{left}), recovered background layer (\emph{middle}), and reflection/occlusion layer (\emph{right}).
}
\label{fig:controlled_visual}
\end{figure}

\begin{table*}
\caption{\textbf{Quantitative evaluation on controlled sequences~\cite{xue2015computational}.} We report the SSIM, NCC, and LMSE of the recovered background and reflection layers on each sequence. 
}
\label{tab:controlled}
\vspace{-2mm}
\centering
\footnotesize
\renewcommand{\tabcolsep}{3pt} %
\renewcommand{\arraystretch}{1} %
\resizebox{\linewidth}{!}{
\begin{tabular}{l|cccccc|cccccc|cccccc}
\toprule
\multirow{3}{*}{Method} & \multicolumn{6}{c|}{Stone} & \multicolumn{6}{c|}{Toy} & \multicolumn{6}{c}{Hanoi} \\
& \multicolumn{3}{c}{Background} & \multicolumn{3}{c|}{Reflection} & \multicolumn{3}{c}{Background} & \multicolumn{3}{c|}{Reflection} & \multicolumn{3}{c}{Background} & \multicolumn{3}{c}{Obstruction} \\
& SSIM $\uparrow$ & NCC $\uparrow$ & LMSE $\downarrow$ & SSIM $\uparrow$ & NCC $\uparrow$ & LMSE $\downarrow$ &
SSIM $\uparrow$ & NCC $\uparrow$ & LMSE $\downarrow$ & SSIM $\uparrow$ & NCC $\uparrow$ & LMSE $\downarrow$ &
SSIM $\uparrow$ & NCC $\uparrow$ & LMSE $\downarrow$ & SSIM $\uparrow$ & NCC $\uparrow$ & LMSE $\downarrow$ \\
\midrule
Li and Brown~\cite{li2013exploiting} & 
0.7993 & 0.9334 & 0.0114 & 
0.2038 & 0.3668 & \textcolor{red}{\pmb{0.0134}} &
0.6877 & 0.7068 & 0.0196 & 
0.1092 & 0.6607 & 0.0105 &
- & - & - & - & - & - \\
Guo~\etal~\cite{guo2014robust} &
0.5292 & 0.7251 & 0.0571 &
0.4749 & 0.1006 & 0.2664 &
0.7081 & 0.7215 & 0.0231 & \textcolor{blue}{\underline{0.4892}} & 0.6625 & 0.0983 &
- & - & - & - & - & - \\
Xue~\etal~\cite{xue2015computational} &
- & \textcolor{red}{\pmb{0.9738}} & - &
- & \textcolor{red}{\pmb{0.8433}} & - &
- & 0.8985 & - &
- & 0.7536 & - &
- & 0.9921 & - &
- & \textcolor{blue}{\underline{0.7079}} & - \\
Alayrac~\etal~\cite{alayrac2019visual} &
0.7942 & 0.9351 & 0.0092 &
\textcolor{red}{\pmb{0.7633}} & 0.1641 & 0.0407 &
0.7569 & 0.7972 & 0.0088 &
0.3652 & 0.5260 & 0.0152 &
- & - & - & - & - & - \\
Liu~\etal~\cite{liu2020learning} &
\textcolor{blue}{\underline{0.8598}} & \textcolor{blue}{\underline{0.9632}} & \textcolor{red}{\pmb{0.0052}} &
0.2041 & \textcolor{blue}{\underline{0.7002}} & 0.0277 &
\textcolor{blue}{\underline{0.7696}} & \textcolor{blue}{\underline{0.9477}} & \textcolor{blue}{\underline{0.0053}} &
\textcolor{red}{\pmb{0.5342}} & \textcolor{blue}{\underline{0.8696}} & \textcolor{blue}{\underline{0.0100}} &
\textcolor{blue}{\underline{0.9238}} & \textcolor{blue}{\underline{0.9929}} & \textcolor{red}{\pmb{0.0018}} &
\textcolor{blue}{\underline{0.2991}} & 0.5621 & \textcolor{blue}{\underline{0.2034}} \\
Ours & 
\textcolor{red}{\pmb{0.8635}} & 0.9315 & \textcolor{blue}{\underline{0.0062}} &
\textcolor{blue}{\underline{0.5146}} & 0.3018 & \textcolor{blue}{\underline{0.0268}} &
\textcolor{red}{\pmb{0.8494}} & \textcolor{red}{\pmb{0.9542}} & \textcolor{red}{\pmb{0.0048}} &
0.3371 & \textcolor{red}{\pmb{0.9046}} & \textcolor{red}{\pmb{0.0056}} &
\textcolor{red}{\pmb{0.9457}} & \textcolor{red}{\pmb{0.9938}} & \textcolor{red}{\pmb{0.0018}} &
\textcolor{red}{\pmb{0.3115}} & \textcolor{red}{\pmb{0.8549}} & \textcolor{red}{\pmb{0.1583}} \\
\bottomrule
\end{tabular}
}
\vspace{-2mm}
\end{table*}

In this section, we present visual and quantitative comparisons with the state-of-the-art obstruction removal algorithms as well as provide detailed ablation study justifying the design choices our method.  
Complete visual results can be found at \url{https://alex04072000.github.io/SOLD/}. 
We also provide the source code and pre-trained models at \url{https://github.com/alex04072000/SOLD}.

\subsection{Comparisons with State-of-the-arts}

\vspace{\paramargin}
{\flushleft {\bf Controlled sequences.}}
We first evaluate on the controlled sequences provided by Xue et al.~\cite{xue2015computational}, which contain three videos with ground-truth background and reflection layers.
We evaluate the proposed method with the approaches by Li and Brown~\cite{li2013exploiting}, Guo et al.~\cite{guo2014robust}, Xue et al.~\cite{xue2015computational}, and Alayrac et al.~\cite{alayrac2019visual} and report the SSIM, normalized cross-correlation (NCC) scores~\cite{wan2017benchmarking, xue2015computational}, and LMSE~\cite{grosse2009ground} in \tabref{controlled}.
\revision{
SSIM measures the structural similarity between the recovered and ground-truth images.
NCC measures the overall quality while ignoring the global scaling of the intensity since the ground-truth images are only defined up to a scaling factor.
LMSE is also a scale-invariant metric but often used to measure errors locally.
They are not always consistent since they are designed to compare images from different aspects and characteristics.
}
\figref{controlled_visual} shows 
our method performs favorably against other approaches in 
reconstructing background and reflection/obstruction layers.

\revision{
Other than the three sequences provided by Xue et al.~\cite{xue2015computational}, to the best of our knowledge, there are no other publicly available sequences with obstruction having ground truths.
In light of this, we collect six sequences with ground truth.
Specifically, we put a camera on a tripod to capture the background scenes behind the obstructions (e.g., glasses or fences).
The captured images, therefore, contain the background objects \emph{and} the obstructions (reflections or obstacles).
We then lace a black flannel behind the obstruction to occlude the background for capturing the ground-truth obstruction images.
Finally, we remove the obstruction to capture the ground-truth images of the background scenes.
We repeat the process for five different camera positions.
Our dataset contains six scenes: two with reflection, two with a fence, and two with semi-transparency.
For the scenes with semi-transparency, we cannot obtain the ground-truth obstruction images.
The reason is, after we occlude the background with a black flannel, the captured image becomes extremely dark as there is no light from behind.
Thus, we only provide ground-truth background images for the scenes with semi-transparency.
We conduct quantitative comparisons with other methods with these six sequences, and TABLE~\ref{tab:controlled_gt} shows the results.
Note that Li and Brown~\cite{li2013exploiting} and Guo~\etal~\cite{guo2014robust} are not methods designed for removing fences and semi-transparent objects.
We still include the results as references.
Our method significantly outperforms the compared methods, including the preliminary version of this work \cite{liu2020learning}. 
\figref{controlled_visual_gt} displays the dataset and visual results of our method.
}

\begin{table*}
\caption{\revision{\textbf{Quantitative comparisons on collected controlled scenes.}}
}
\label{tab:controlled_gt}
\vspace{-2mm}
\centering
\footnotesize
\renewcommand{\tabcolsep}{3pt} %
\renewcommand{\arraystretch}{1} %
\resizebox{\linewidth}{!}{
\begin{tabular}{l|cccccc|cccccc|cccccc}
\toprule
\multirow{3}{*}{Method} & \multicolumn{6}{c|}{Reflection 1} & \multicolumn{6}{c|}{Reflection 2} & \multicolumn{6}{c}{Fence 1} \\
& \multicolumn{3}{c}{Background} & \multicolumn{3}{c|}{Reflection} & \multicolumn{3}{c}{Background} & \multicolumn{3}{c|}{Reflection} & \multicolumn{3}{c}{Background} & \multicolumn{3}{c}{Obstruction} \\
& SSIM $\uparrow$ & NCC $\uparrow$ & LMSE $\downarrow$ & SSIM $\uparrow$ & NCC $\uparrow$ & LMSE $\downarrow$ &
SSIM $\uparrow$ & NCC $\uparrow$ & LMSE $\downarrow$ & SSIM $\uparrow$ & NCC $\uparrow$ & LMSE $\downarrow$ &
SSIM $\uparrow$ & NCC $\uparrow$ & LMSE $\downarrow$ & SSIM $\uparrow$ & NCC $\uparrow$ & LMSE $\downarrow$ \\
\midrule
Li and Brown [14] & 0.7829 & 0.9259 & 0.0177 & 0.3546 & 0.3971 & \textcolor{red}{\pmb{0.0148}} & 0.5746 & 0.6991 & 0.0846 & 0.3064 & 0.0822 & \textcolor{red}{\pmb{0.0187}} & 0.6576 & 0.7571 & 0.0406 & 0.0964 & 0.0134 & 0.2921 \\
Guo~\etal [15] & 0.6034 & 0.6690 & 0.0702 & 0.3279 & 0.1013 & 0.2005 & 0.6213 & 0.6784 & 0.0694 & 0.3288 & 0.0073 & 0.2953 & 0.6227 & 0.8791 & 0.0275 & 0.2088 & -0.1744 & 0.5705 \\
Alayrac~\etal [19] & 0.8304 & 0.9641 & 0.0085 & \textcolor{red}{\pmb{0.6884}} & 0.0630 & 0.0540 & 0.7947 & 0.9287 & 0.0161 & \textcolor{red}{\pmb{0.6060}} & -0.1024 & 0.0876 & 0.7478 & 0.9106 & 0.0311 & \textcolor{blue}{\underline{0.3593}} & 0.0822 & 0.3448 \\
Liu~\etal [61] & \textcolor{blue}{\underline{0.8852}} & \textcolor{blue}{\underline{0.9788}} & \textcolor{blue}{\underline{0.0054}} & 0.3475 & \textcolor{blue}{\underline{0.4181}} & 0.2371 & \textcolor{blue}{\underline{0.8543}} & \textcolor{blue}{\underline{0.9674}} & \textcolor{red}{\pmb{0.0065}} & 0.3425 & \textcolor{blue}{\underline{0.2245}} & 0.1486 & \textcolor{blue}{\underline{0.9519}} & \textcolor{blue}{\underline{0.9956}} & \textcolor{blue}{\underline{0.0017}} & \textcolor{red}{\pmb{0.3698}} & \textcolor{red}{\pmb{0.9170}} & \textcolor{red}{\pmb{0.0797}} \\
Ours & \textcolor{red}{\pmb{0.9167}} & \textcolor{red}{\pmb{0.9867}} & \textcolor{red}{\pmb{0.0036}} & \textcolor{blue}{\underline{0.4839}} & \textcolor{red}{\pmb{0.4968}} & \textcolor{blue}{\underline{0.0412}} & \textcolor{red}{\pmb{0.8939}} & \textcolor{red}{\pmb{0.9826}} & \textcolor{blue}{\underline{0.0069}} & \textcolor{blue}{\underline{0.3639}} & \textcolor{red}{\pmb{0.5265}} & \textcolor{blue}{\underline{0.0396}} & \textcolor{red}{\pmb{0.9711}} & \textcolor{red}{\pmb{0.9975}} & \textcolor{red}{\pmb{0.0010}} & 0.3530 & \textcolor{blue}{\underline{0.8932}} & \textcolor{blue}{\underline{0.0994}} \\
\bottomrule
\end{tabular}
}
\vspace{2mm}
\vspace{2mm}
\resizebox{\linewidth}{!}{
\begin{tabular}{l|cccccc|cccccc|cccccc}
\toprule
\multirow{3}{*}{Method} & \multicolumn{6}{c|}{Fence 2} & \multicolumn{6}{c|}{Semi-transparency} & \multicolumn{6}{c}{\revision{Adherent raindrop}} \\
& \multicolumn{3}{c}{Background} & \multicolumn{3}{c|}{Reflection} & \multicolumn{3}{c}{Background} & \multicolumn{3}{c|}{Reflection} & \multicolumn{3}{c}{Background} & \multicolumn{3}{c}{Obstruction} \\
& SSIM $\uparrow$ & NCC $\uparrow$ & LMSE $\downarrow$ & SSIM $\uparrow$ & NCC $\uparrow$ & LMSE $\downarrow$ &
SSIM $\uparrow$ & NCC $\uparrow$ & LMSE $\downarrow$ & SSIM $\uparrow$ & NCC $\uparrow$ & LMSE $\downarrow$ &
SSIM $\uparrow$ & NCC $\uparrow$ & LMSE $\downarrow$ & SSIM $\uparrow$ & NCC $\uparrow$ & LMSE $\downarrow$ \\
\midrule
Li and Brown [14] & 0.7458 & 0.9162 & 0.0116 & 0.2102 & 0.4125 & \textcolor{blue}{\underline{0.1109}} & 0.7694 & 0.8618 & 0.0272 & - & - & - & 0.8102 & 0.8584 & 0.0095 & - & - & - \\
Guo~\etal [15] & 0.7536 & 0.9339 & 0.0228 & \textcolor{blue}{\underline{0.4345}} & 0.4920 & \textcolor{red}{\pmb{0.1059}} & 0.6630 & 0.6953 & 0.0649 & - & - & - & 0.7375 & 0.7958 & 0.0255 & - & - & - \\
Alayrac~\etal [19] & 0.8003 & 0.9754 & 0.0069 & \textcolor{red}{\pmb{0.4816}} & 0.3996 & 0.1946 & 0.8073 & 0.8880 & 0.0213 & - & - & - & 0.8043 & 0.8514 & 0.0119 & - & - & - \\
Liu~\etal [61] & \textcolor{blue}{\underline{0.9118}} & \textcolor{blue}{\underline{0.9904}} & \textcolor{blue}{\underline{0.0038}} & 0.3658 & \textcolor{blue}{\underline{0.6800}} & 0.3509 & \textcolor{blue}{\underline{0.9009}} & \textcolor{blue}{\underline{0.9832}} & \textcolor{blue}{\underline{0.0078}} & - & - & - & \textcolor{blue}{\underline{0.9031}} & \textcolor{blue}{\underline{0.9763}} & \textcolor{blue}{\underline{0.0034}} & - & - & - \\
Ours & \textcolor{red}{\pmb{0.9416}} & \textcolor{red}{\pmb{0.9941}} & \textcolor{red}{\pmb{0.0025}} & 0.3392 & \textcolor{red}{\pmb{0.6865}} & 0.3619 & \textcolor{red}{\pmb{0.9378}} & \textcolor{red}{\pmb{0.9903}} & \textcolor{red}{\pmb{0.0047}} & - & - & - & \textcolor{red}{\pmb{0.9312}} & \textcolor{red}{\pmb{0.9869}} & \textcolor{red}{\pmb{0.0019}} & - & - & - \\
\bottomrule
\end{tabular}
}
\vspace{-2mm}
\end{table*}

\begin{figure*}
\centering
\footnotesize
\renewcommand{\tabcolsep}{1pt} %
\renewcommand{\arraystretch}{1} %
\begin{tabular}{cccccc}
    \raisebox{3.7\normalbaselineskip}[0pt][0pt]{\rotatebox[origin=c]{90}{Reflection 1}} &  
    \includegraphics[width=0.19\linewidth]{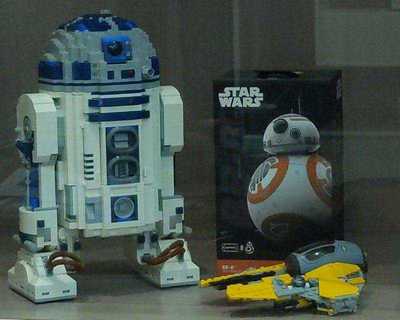} & \includegraphics[width=0.19\linewidth]{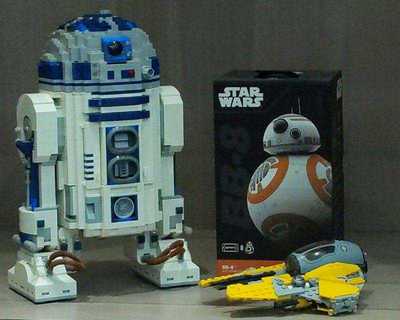} & \includegraphics[width=0.19\linewidth]{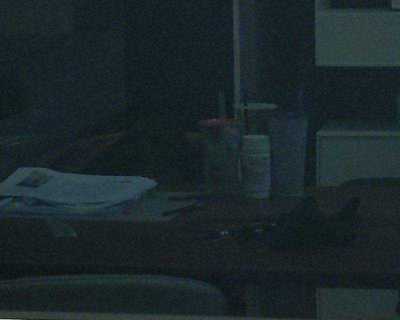} & \includegraphics[width=0.19\linewidth]{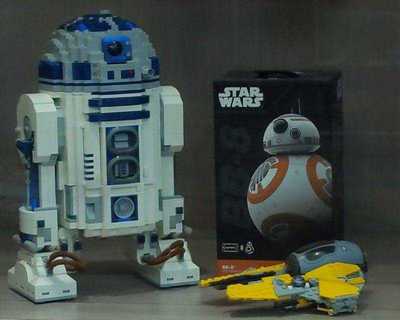} & \includegraphics[width=0.19\linewidth]{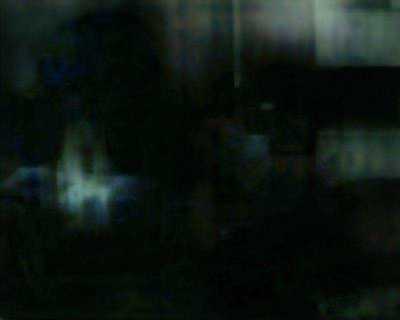} \\
    \raisebox{2.3\normalbaselineskip}[0pt][0pt]{\rotatebox[origin=c]{90}{Reflection 2}} &  
    \includegraphics[width=0.19\linewidth]{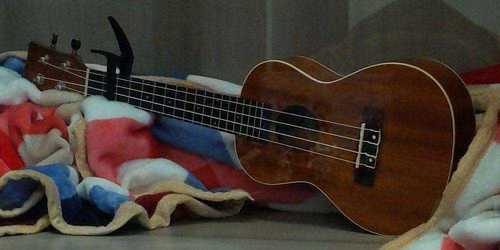} & \includegraphics[width=0.19\linewidth]{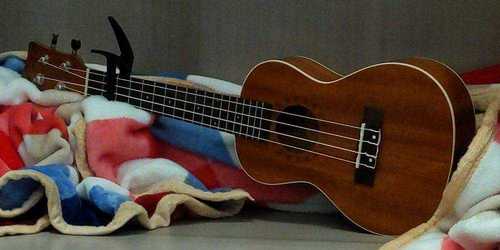} & \includegraphics[width=0.19\linewidth]{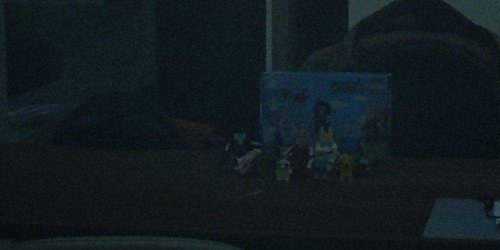} & \includegraphics[width=0.19\linewidth]{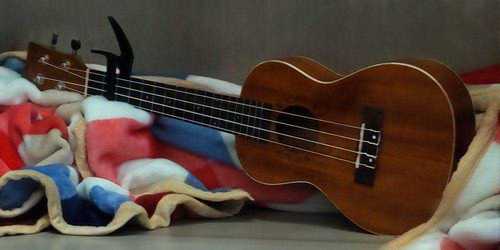} & \includegraphics[width=0.19\linewidth]{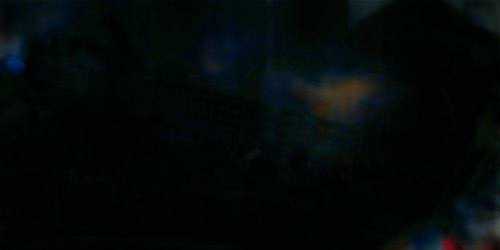} \\
    \raisebox{4.0\normalbaselineskip}[0pt][0pt]{\rotatebox[origin=c]{90}{Fence 1}} &  
    \includegraphics[width=0.19\linewidth]{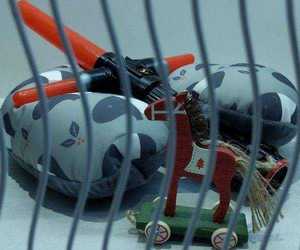} & \includegraphics[width=0.19\linewidth]{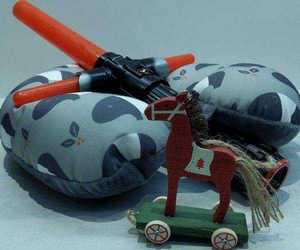} & \includegraphics[width=0.19\linewidth]{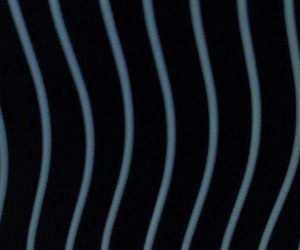} & \includegraphics[width=0.19\linewidth]{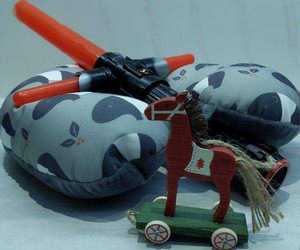} & \includegraphics[width=0.19\linewidth]{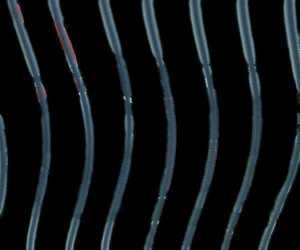} \\
    \raisebox{4.0\normalbaselineskip}[0pt][0pt]{\rotatebox[origin=c]{90}{Fence 2}} &  
    \includegraphics[width=0.19\linewidth]{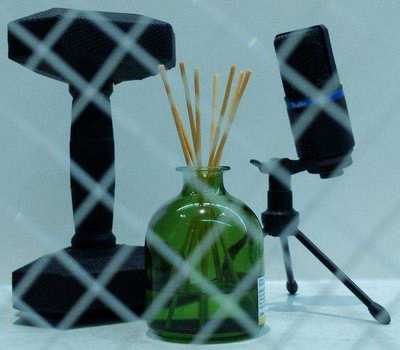} & \includegraphics[width=0.19\linewidth]{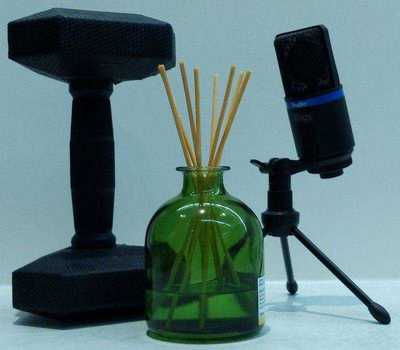} & \includegraphics[width=0.19\linewidth]{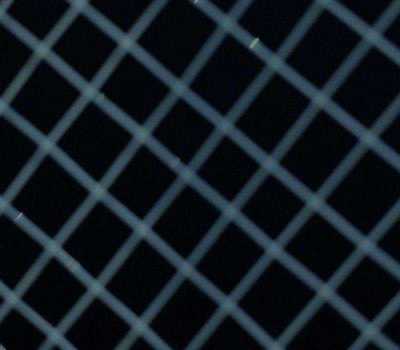} & \includegraphics[width=0.19\linewidth]{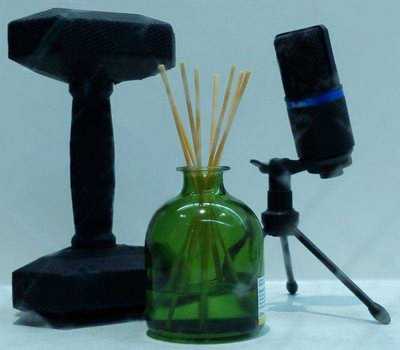} & \includegraphics[width=0.19\linewidth]{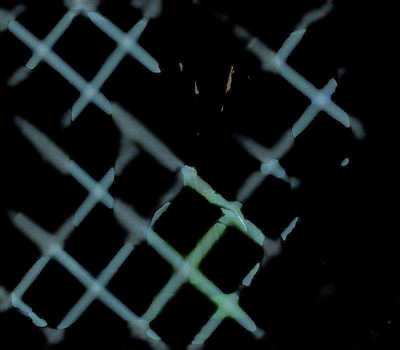} \\
    \raisebox{4.5\normalbaselineskip}[0pt][0pt]{\rotatebox[origin=c]{90}{Semi-transparency}} &  
    \includegraphics[width=0.19\linewidth]{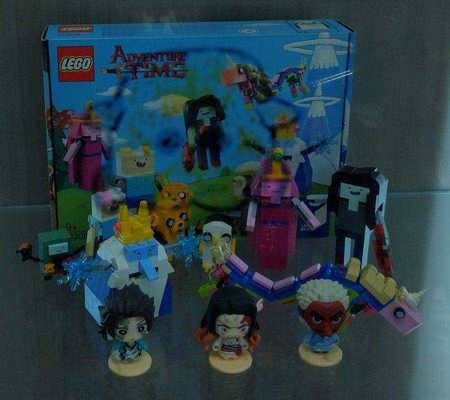} & \includegraphics[width=0.19\linewidth]{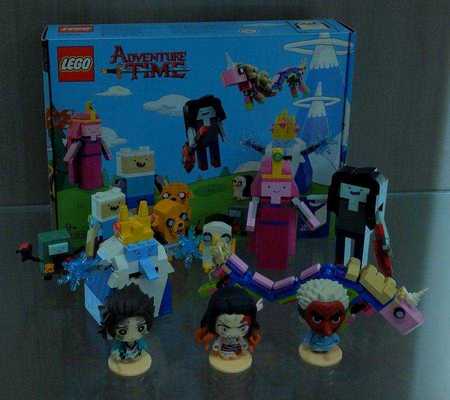} & & \includegraphics[width=0.19\linewidth]{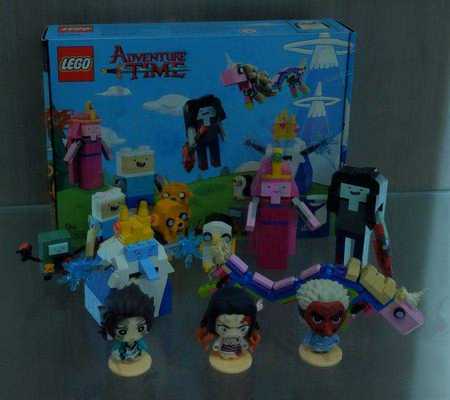} & \includegraphics[width=0.19\linewidth]{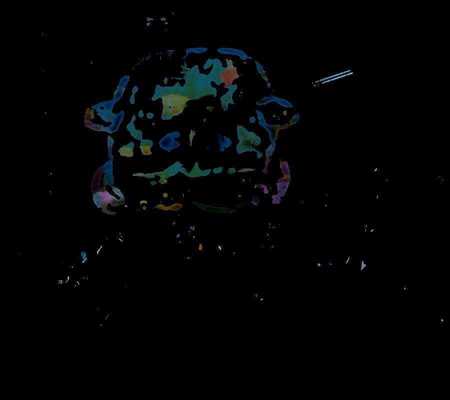} \\
    \raisebox{3.8\normalbaselineskip}[0pt][0pt]{\rotatebox[origin=c]{90}{\revision{Adherent raindrop}}} &  
    \includegraphics[width=0.19\linewidth]{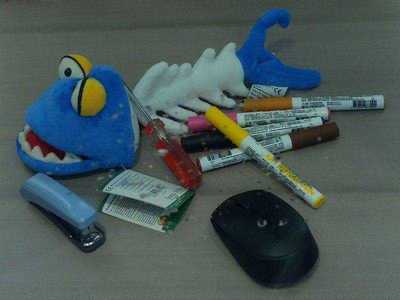} & \includegraphics[width=0.19\linewidth]{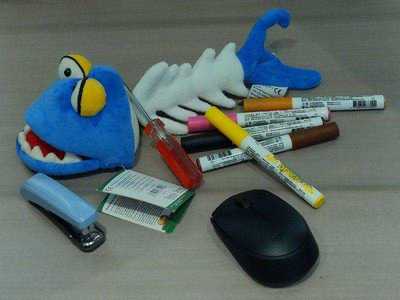} & & \includegraphics[width=0.19\linewidth]{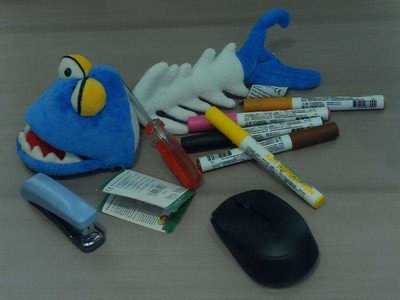} & \includegraphics[width=0.19\linewidth]{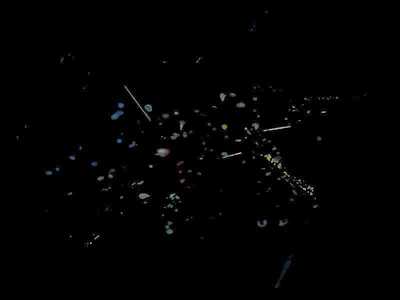} \\
    & Input (keyframe) & Background GT & Obstruction GT & Recovered background & Recovered obstruction \\
\end{tabular}
\caption{\revision{\textbf{Visual results on the collected controlled sequences.} For each sequence, from left to right, we show the keyframe, the ground-truth background, the ground-truth obstruction (if available), the background layer and reflection/occlusion layer recovered by our method.}
}
\label{fig:controlled_visual_gt}
\end{figure*}

\vspace{\paramargin}
{\flushleft {\bf Synthetic sequences.}}
We synthesize 100 sequences from the Vimeo-90k test set using the method described in Section~\ref{sec:data_generation}.
We evaluate our approach with five single-image reflection removal methods~\cite{fan2017generic,jin2018learning,wei2019single,yang2018seeing,zhang2018single}, and three multi-frame approaches~\cite{alayrac2019visual, guo2014robust,li2013exploiting}.
We use the default parameters of each method to generate the results.
Since the source code or pre-trained models of Alayrac et al.~\cite{alayrac2019visual} are not available, we re-implement their model and train on our training dataset (with the help from the authors).
\revision{Note that our reimplementation results for Alayrac et al.~\cite{alayrac2019visual} are not as good as those presented in their paper. The reason is that we train their method using our data generation scheme for fair comparison while their model was trained on a much large dataset containing 400k 250-frame videos. It shows, thanks to having more inductive bias, our method does not require a large training dataset of long sequences compared to their method.}
\tabref{compare_stoa_reflection} shows the average PSNR, SSIM~\cite{wang2004image}, NCC, and LMSE~\cite{grosse2009ground} metrics.
Note that the proposed method without the online optimization already performs favorably against existing approaches.
By incorporating the online optimization, we can further improve the average SSIM and NCC on both the background and reflection layers.

\begin{table*}
\caption{
        \textbf{Quantitative comparison of reflection removal on synthetic sequences.}
        We evaluate on our synthetic dataset with 100 sequences, where each sequence contains five consecutive frames.
        For the single-image based methods~\cite{fan2017generic, jin2018learning, wei2019single, yang2018seeing, zhang2018single}, we generate the results frame-by-frame.
        For multi-frame algorithms~\cite{alayrac2019visual, guo2014robust, li2013exploiting, liu2020learning} and our method, we use five input frames to generate the results.
    }
\label{tab:compare_stoa_reflection}
\vspace{-2mm}
\centering
\resizebox{\linewidth}{!}{
\begin{tabular}{l|l|l|cccc|cccc}
\toprule
\multicolumn{3}{l|}{\multirow{2}{*}{Method}} & \multicolumn{4}{c|}{Background} & \multicolumn{4}{c}{Reflection} \\
\multicolumn{3}{l|}{}                        & PSNR $\uparrow$ & SSIM $\uparrow$ & NCC $\uparrow$ & LMSE $\downarrow$ & PSNR $\uparrow$ & SSIM $\uparrow$ & NCC $\uparrow$ & LMSE $\downarrow$ \\
\midrule
\multirow{5}{*}{Single image} & CEILNet~\cite{fan2017generic} & CNN-based & 18.64 & 0.6808 & 0.8102 & 0.0408 & - & - & - & - \\
& Zhang~\etal~\cite{zhang2018single} & CNN-based & 17.27 & 0.6861 & 0.8142 & 0.0272 & 15.61 & 0.4271 & 0.5368 & 0.1173  \\
& BDN~\cite{yang2018seeing} & CNN-based & 15.49 & 0.6654 & 0.7076 & 0.0426 & - & - & - & - \\
& ERRNet~\cite{wei2019single} & CNN-based & 20.19 & 0.7530 & 0.8157 & 0.0198 & - & - & - & - \\
& Jin~\etal~\cite{jin2018learning} & CNN-based & 16.78 & 0.6993 & 0.7321 & 0.0242 & 9.13 & 0.3069 & 0.3779 & 0.1276 \\
\midrule
\multirow{6}{*}{Multiple images} & Li and Brown~\cite{li2013exploiting} & Optimization-based & 15.36 & 0.5950 & 0.6155 & 0.0802 & 7.00 & 0.2047 & 0.2809 & 0.1335 \\
& Guo~\etal~\cite{guo2014robust} & Optimization-based & 13.51 & 0.4835 & 0.5460 & 0.0909 & 11.85 & 0.2408 & 0.2517 & 0.1975 \\
& Alayrac~\etal~\cite{alayrac2019visual} & CNN-based & 21.12 & 0.7277 & 0.8520 & 0.0248 & 16.84 & 0.5283 & 0.6225 & 0.1706 \\
& Liu~\etal~\cite{liu2020learning} & CNN-based & 23.82 & 0.8082 & 0.8936 & \textcolor{blue}{\underline{0.0150}} & 17.65 & 0.5338 & 0.6299 & 0.1103 \\
& Ours w/o online optimization & CNN-based & \textcolor{red}{\pmb{26.75}} & \textcolor{blue}{\underline{0.8742}} & \textcolor{blue}{\underline{0.9247}} & \textcolor{red}{\pmb{0.0114}} & \textcolor{red}{\pmb{20.40}} & \textcolor{blue}{\underline{0.6329}} & \textcolor{blue}{\underline{0.7731}} & \textcolor{blue}{\underline{0.0974}} \\
& Ours & CNN-based & \textcolor{blue}{\underline{25.98}} & \textcolor{red}{\pmb{0.8916}} & \textcolor{red}{\pmb{0.9516}} & 0.0169 & \textcolor{blue}{\underline{19.81}} & \textcolor{red}{\pmb{0.7141}} & \textcolor{red}{\pmb{0.7894}} & \textcolor{red}{\pmb{0.0921}} \\
\bottomrule
\end{tabular}
}
\vspace{-2mm}
\end{table*}

\begin{figure*}
\centering
\footnotesize
\renewcommand{\tabcolsep}{1pt} %
\renewcommand{\arraystretch}{1} %
\newcommand{\imagewidth}{0.31\columnwidth}
\begin{tabular}{ccccccc}
    \includegraphics[width=\imagewidth]{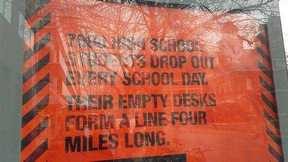} & 
    \raisebox{2.0\normalbaselineskip}[0pt][0pt]{\rotatebox[origin=c]{90}{Background}} &  
    \includegraphics[width=\imagewidth]{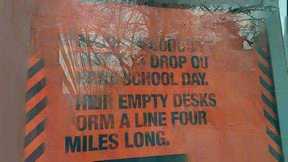} &  
    \includegraphics[width=\imagewidth]{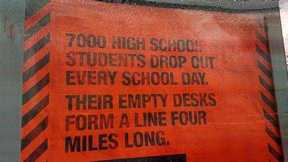} &  
    \includegraphics[width=\imagewidth]{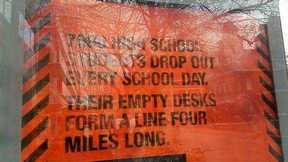} & 
    \includegraphics[width=\imagewidth]{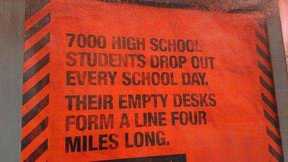} &
    \includegraphics[width=\imagewidth]{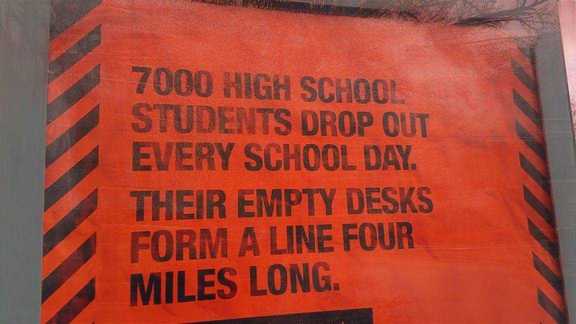}
    \\
    & 
    \raisebox{2.0\normalbaselineskip}[0pt][0pt]{\rotatebox[origin=c]{90}{Reflection}} &  
    \includegraphics[width=\imagewidth]{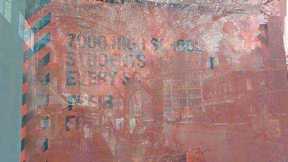} &  
    \includegraphics[width=\imagewidth]{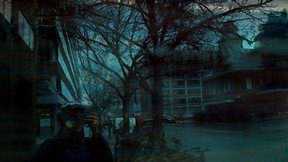} &  
    \includegraphics[width=\imagewidth]{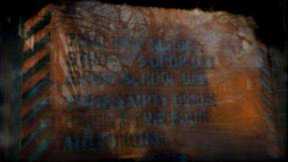} & 
    \includegraphics[width=\imagewidth]{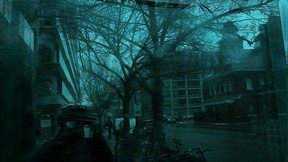} &
    \includegraphics[width=\imagewidth]{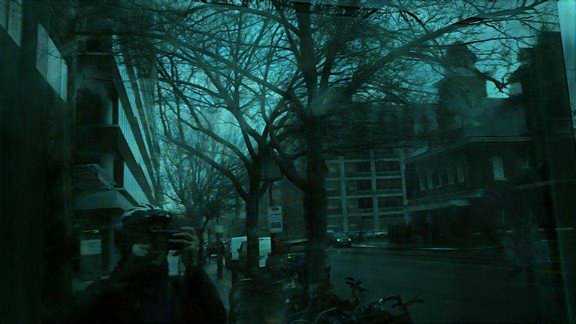} 
    \\
    \includegraphics[width=\imagewidth]{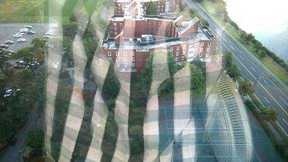} & 
    \raisebox{2.0\normalbaselineskip}[0pt][0pt]{\rotatebox[origin=c]{90}{Background}} &  
    \includegraphics[width=\imagewidth]{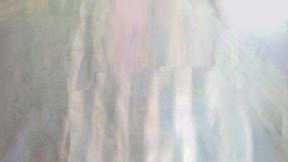} &  
    \includegraphics[width=\imagewidth]{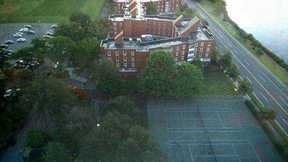} &  
    \includegraphics[width=\imagewidth]{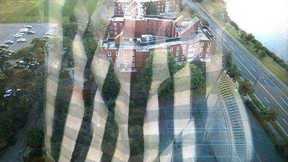} & 
    \includegraphics[width=\imagewidth]{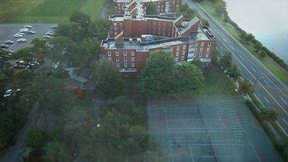} &
    \includegraphics[width=\imagewidth]{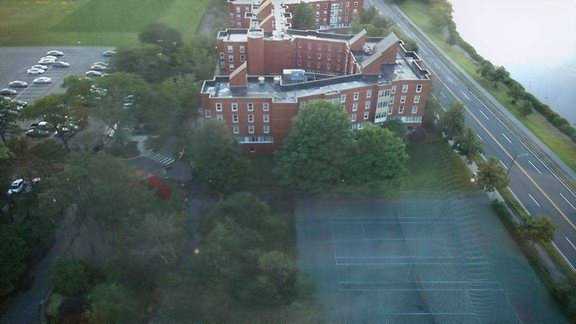}
    \\
    & 
    \raisebox{2.0\normalbaselineskip}[0pt][0pt]{\rotatebox[origin=c]{90}{Reflection}} &  
    \includegraphics[width=\imagewidth]{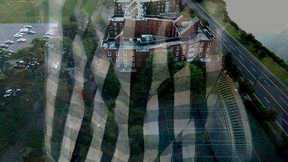} &  
    \includegraphics[width=\imagewidth]{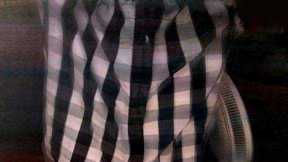} &  
    \includegraphics[width=\imagewidth]{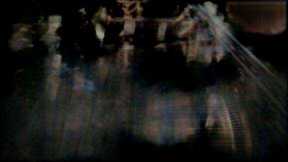} & 
    \includegraphics[width=\imagewidth]{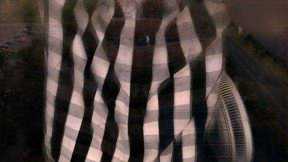} &
    \includegraphics[width=\imagewidth]{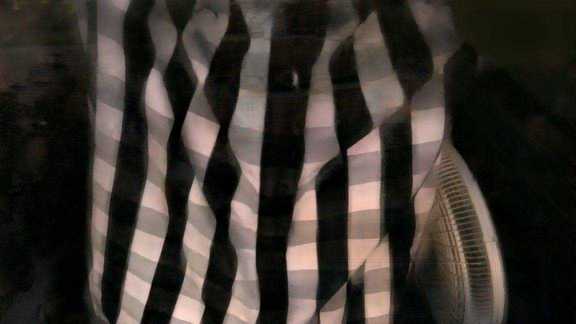} 
    \\
    \includegraphics[width=\imagewidth]{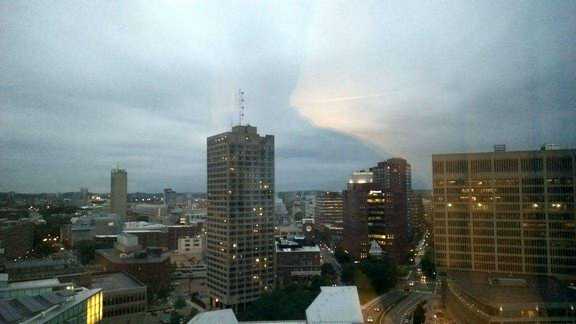} & 
    \raisebox{2.0\normalbaselineskip}[0pt][0pt]{\rotatebox[origin=c]{90}{Background}} &  
    \includegraphics[width=\imagewidth]{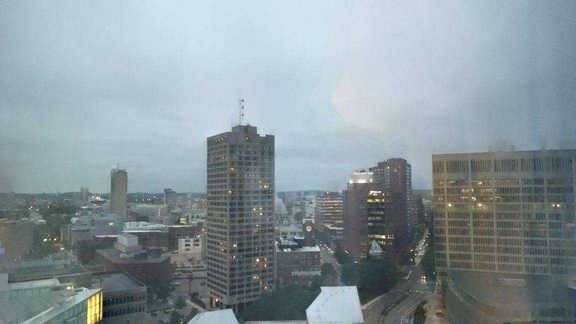} &  
    \includegraphics[width=\imagewidth]{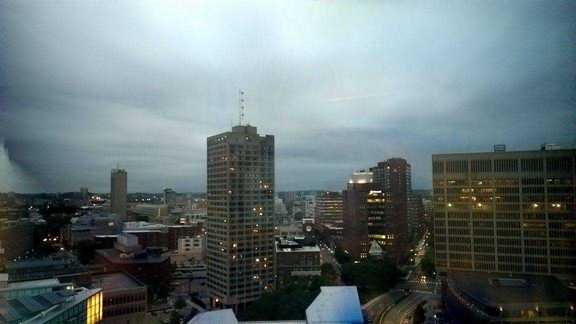} &  
    \includegraphics[width=\imagewidth]{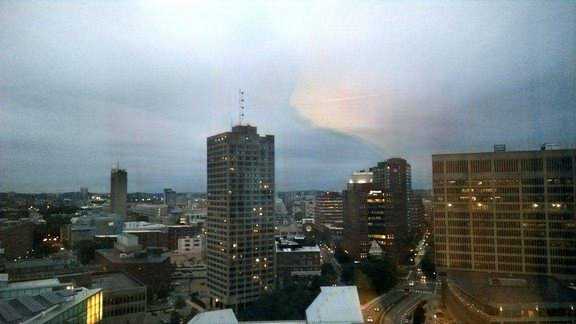} & 
    \includegraphics[width=\imagewidth]{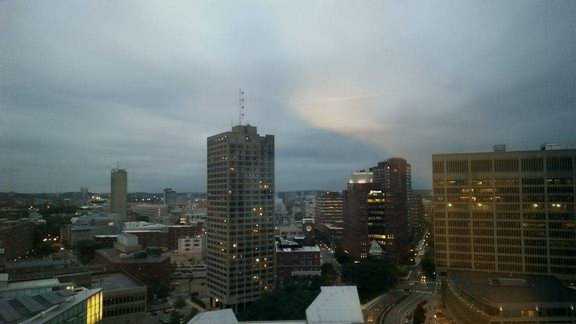} &
    \includegraphics[width=\imagewidth]{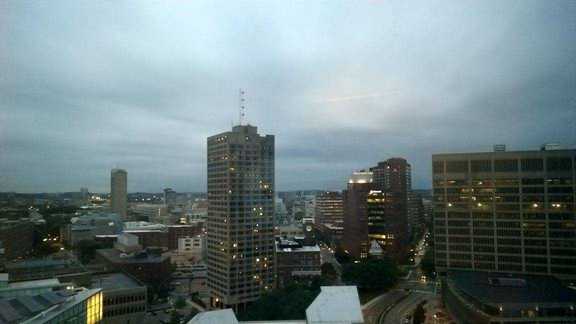}
    \\
    & 
    \raisebox{2.0\normalbaselineskip}[0pt][0pt]{\rotatebox[origin=c]{90}{Reflection}} &  
    \includegraphics[width=\imagewidth]{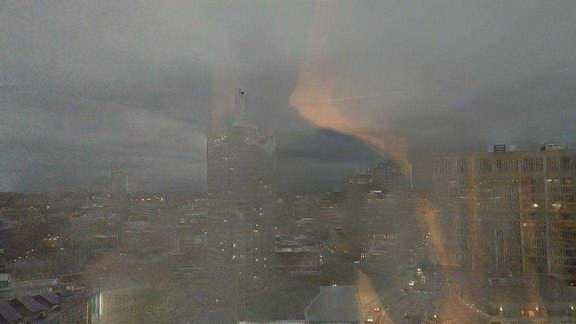} &  
    \includegraphics[width=\imagewidth]{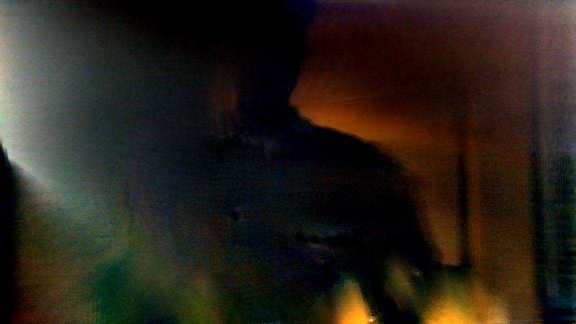} &  
    \includegraphics[width=\imagewidth]{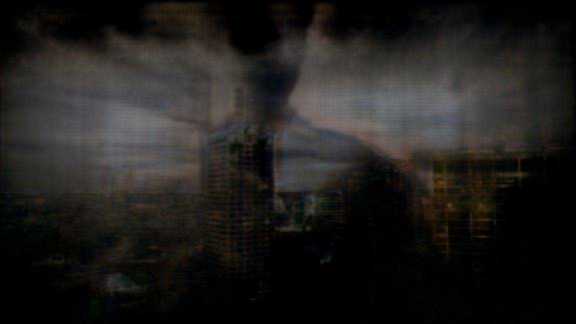} & 
    \includegraphics[width=\imagewidth]{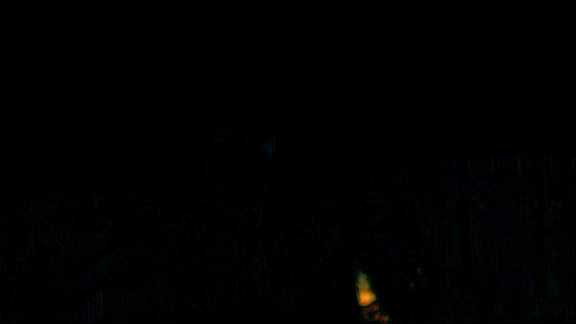} &
    \includegraphics[width=\imagewidth]{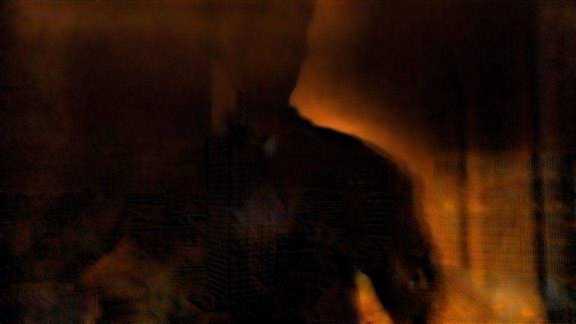} 
    \\
    \includegraphics[width=\imagewidth]{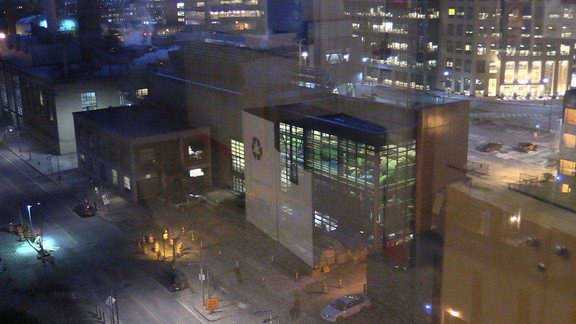} & 
    \raisebox{2.0\normalbaselineskip}[0pt][0pt]{\rotatebox[origin=c]{90}{Background}} &  
    \includegraphics[width=\imagewidth]{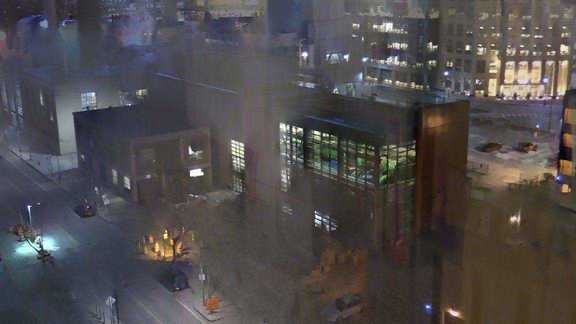} &  
    \includegraphics[width=\imagewidth]{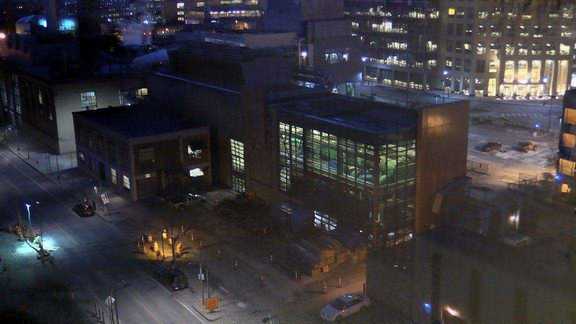} &  
    \includegraphics[width=\imagewidth]{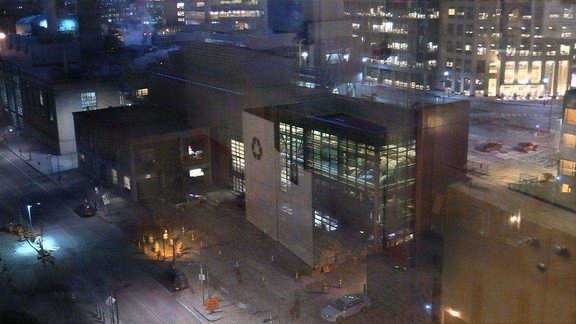} & 
    \includegraphics[width=\imagewidth]{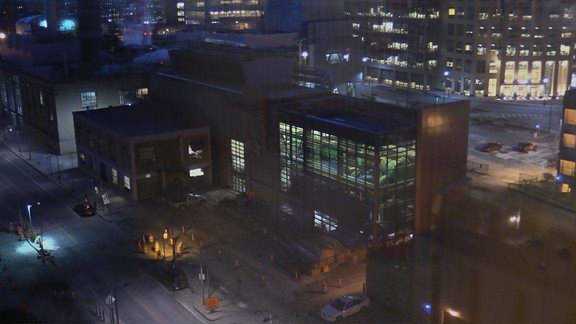} &
    \includegraphics[width=\imagewidth]{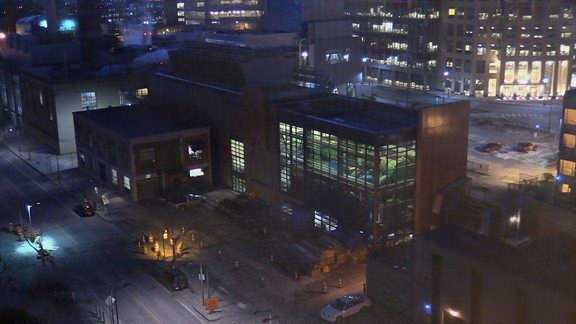}
    \\
    & 
    \raisebox{2.0\normalbaselineskip}[0pt][0pt]{\rotatebox[origin=c]{90}{Reflection}} &  
    \includegraphics[width=\imagewidth]{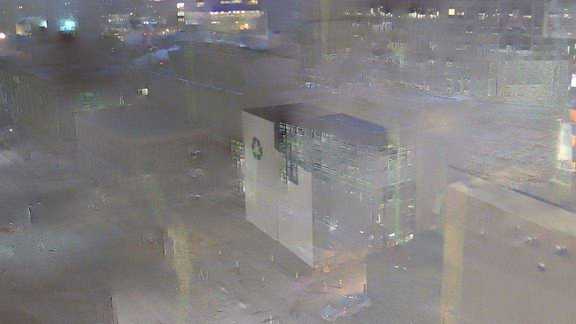} &  
    \includegraphics[width=\imagewidth]{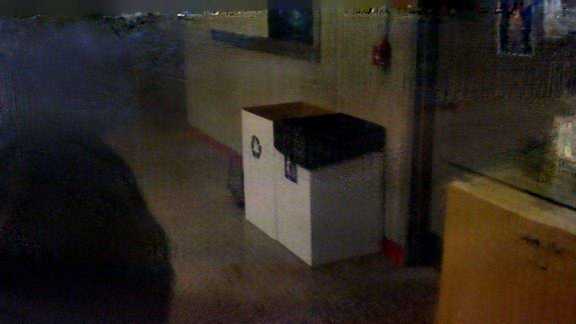} &  
    \includegraphics[width=\imagewidth]{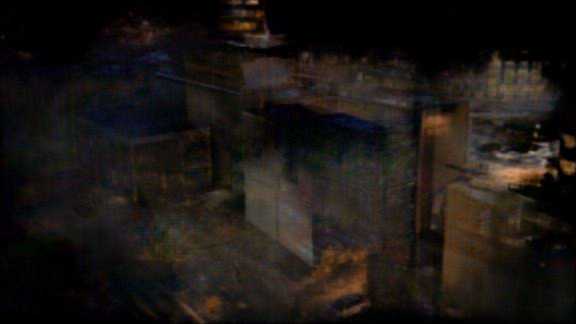} & 
    \includegraphics[width=\imagewidth]{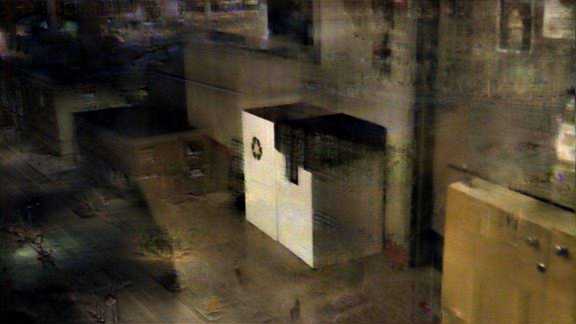} &
    \includegraphics[width=\imagewidth]{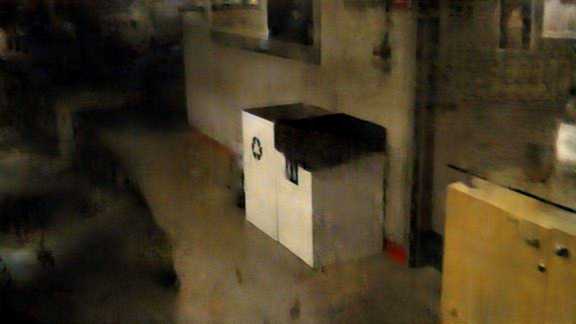} 
    \\
    Representative input frame & & 
    Li and Brown~\cite{li2013exploiting} &
    Xue~\etal~\cite{xue2015computational} &
    Alayrac~\etal~\cite{alayrac2019visual} & 
    Liu~\etal~\cite{liu2020learning} &
    Ours
\end{tabular}
\caption{\textbf{Visual comparisons of background-reflection separation on natural sequences provided by~\cite{xue2015computational}.}
} 
\label{fig:reflection_visual_1}
\end{figure*}

\begin{figure*}
\centering
\footnotesize
\renewcommand{\tabcolsep}{1pt} %
\renewcommand{\arraystretch}{1} %
\newcommand{\imagewidth}{0.31\columnwidth}
\begin{tabular}{ccccccc}
    \includegraphics[width=\imagewidth]{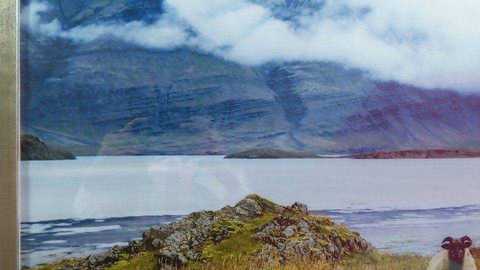} & 
    \raisebox{2.0\normalbaselineskip}[0pt][0pt]{\rotatebox[origin=c]{90}{Background}} &  
    \includegraphics[width=\imagewidth]{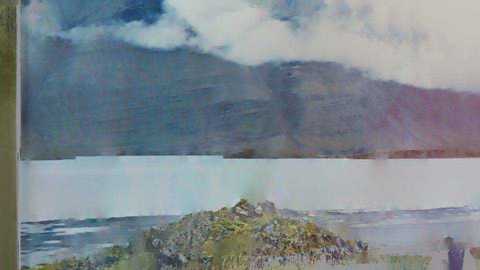} &  
    \includegraphics[width=\imagewidth]{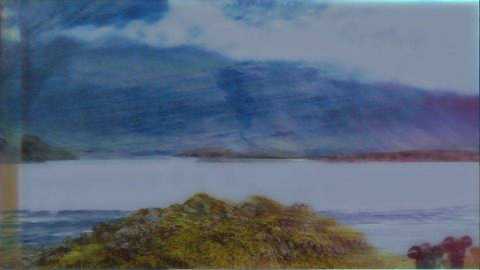} &  
    \includegraphics[width=\imagewidth]{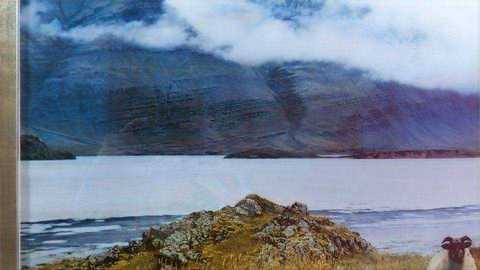} &  
    \includegraphics[width=\imagewidth]{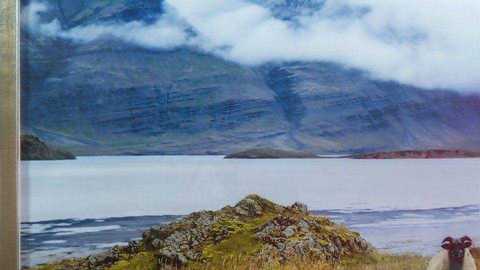}  & 
    \includegraphics[width=\imagewidth]{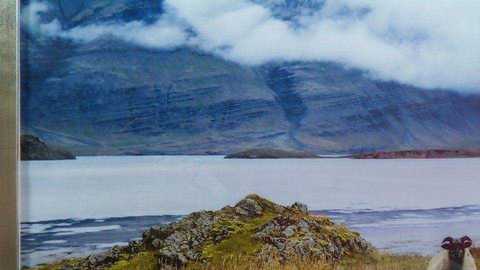} 
    \\
    & 
    \raisebox{2.0\normalbaselineskip}[0pt][0pt]{\rotatebox[origin=c]{90}{Reflection}} &  
    \includegraphics[width=\imagewidth]{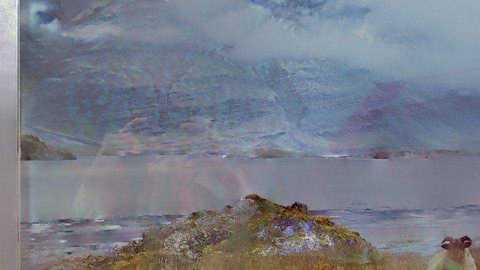} &  
    \includegraphics[width=\imagewidth]{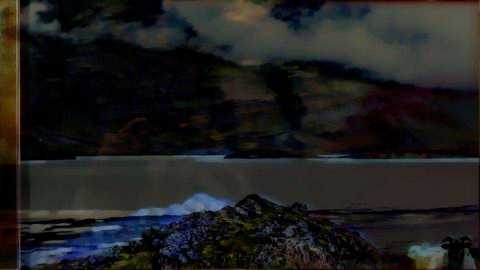} &  
    \includegraphics[width=\imagewidth]{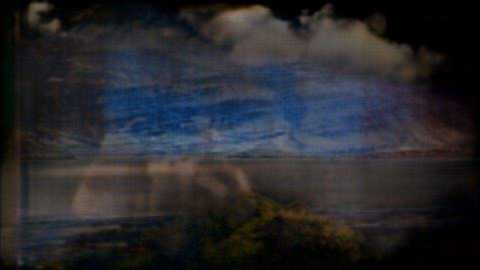} &  
    \includegraphics[width=\imagewidth]{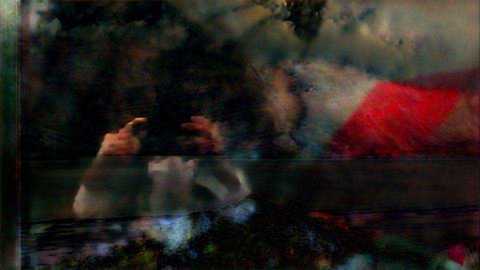}  & 
    \includegraphics[width=\imagewidth]{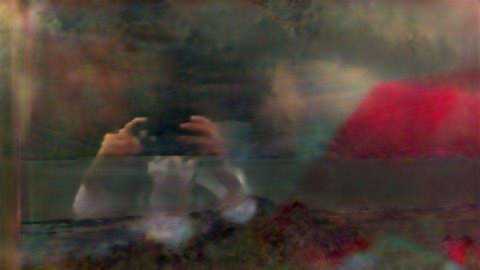} 
    \\
    
    \includegraphics[width=\imagewidth]{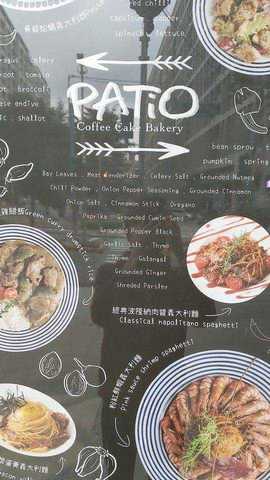} & 
    \raisebox{2.0\normalbaselineskip}[0pt][0pt]{\rotatebox[origin=c]{90}{Background}} &  
    \includegraphics[width=\imagewidth]{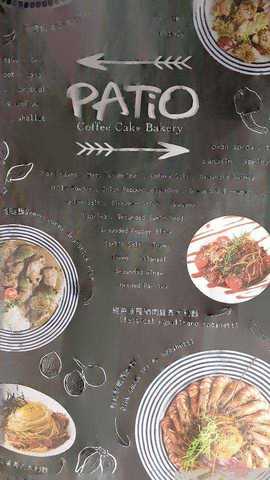} &  
    \includegraphics[width=\imagewidth]{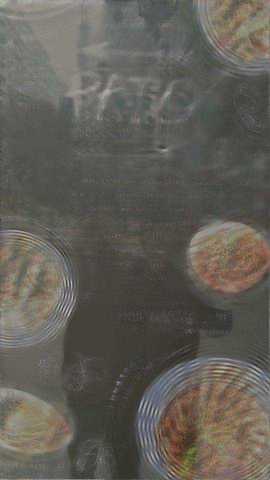} &  
    \includegraphics[width=\imagewidth]{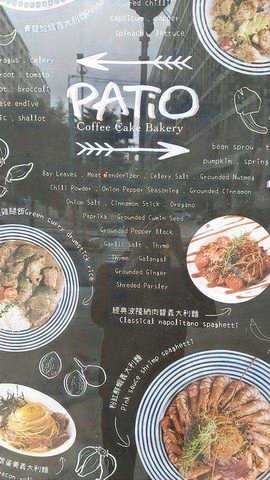} &  
    \includegraphics[width=\imagewidth]{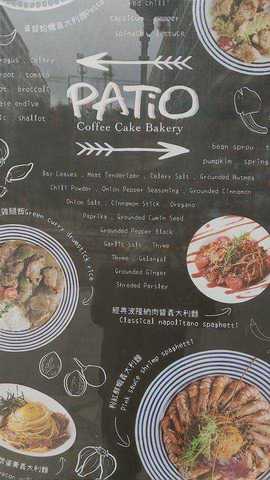}  & 
    \includegraphics[width=\imagewidth]{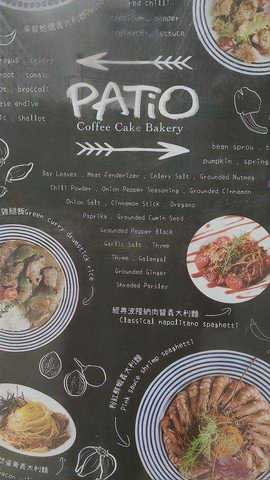} 
    \\
    & 
    \raisebox{2.0\normalbaselineskip}[0pt][0pt]{\rotatebox[origin=c]{90}{Reflection}} &  
    \includegraphics[width=\imagewidth]{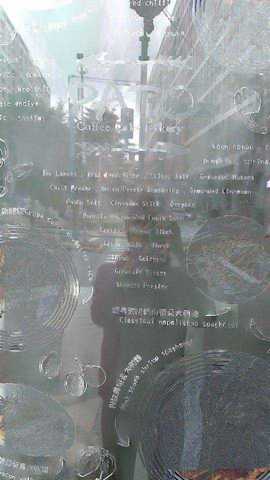} &  
    \includegraphics[width=\imagewidth]{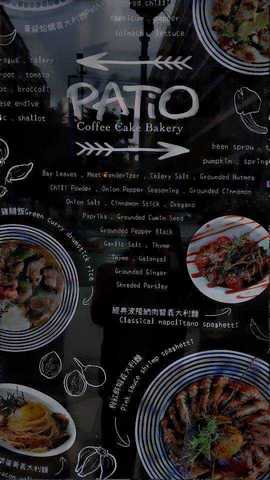} &  
    \includegraphics[width=\imagewidth]{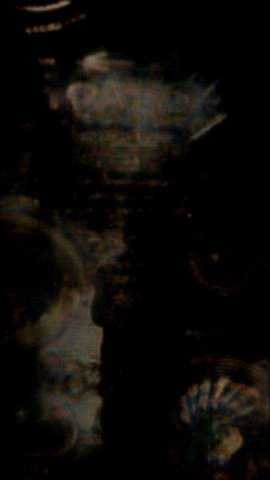} &  
    \includegraphics[width=\imagewidth]{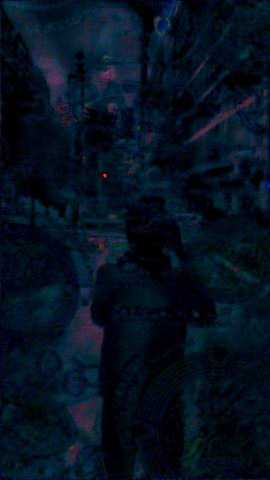}  & 
    \includegraphics[width=\imagewidth]{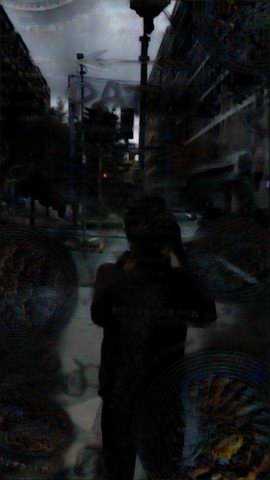} 
    \\
    
    Representative input frame & & 
    Li and Brown~\cite{li2013exploiting} & 
    Guo~\etal~\cite{guo2014robust} & 
    Alayrac~\etal~\cite{alayrac2019visual} & 
    Liu~\etal~\cite{liu2020learning} &
    Ours
\end{tabular}
\caption{\textbf{Visual comparisons of background-reflection separation on natural sequences. 
}
} 
\label{fig:reflection_visual_2}
\end{figure*}

\begin{figure*}
\centering
\footnotesize
\renewcommand{\tabcolsep}{1pt} %
\renewcommand{\arraystretch}{1} %
\newcommand{\imagewidth}{0.18\linewidth}
\begin{tabular}{lccccc}
    \includegraphics[width=\imagewidth]{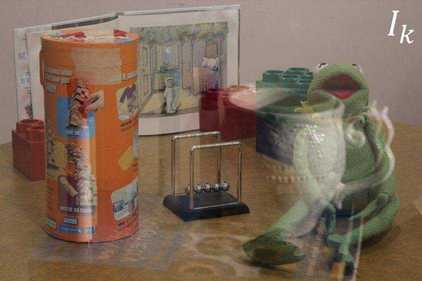} & 
    \raisebox{3\normalbaselineskip}[0pt][0pt]{\rotatebox[origin=c]{90}{Background}} &
    \includegraphics[width=\imagewidth]{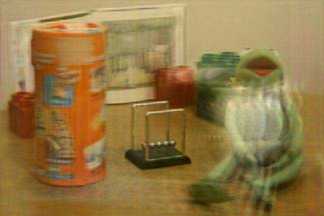} & 
    \includegraphics[width=\imagewidth]{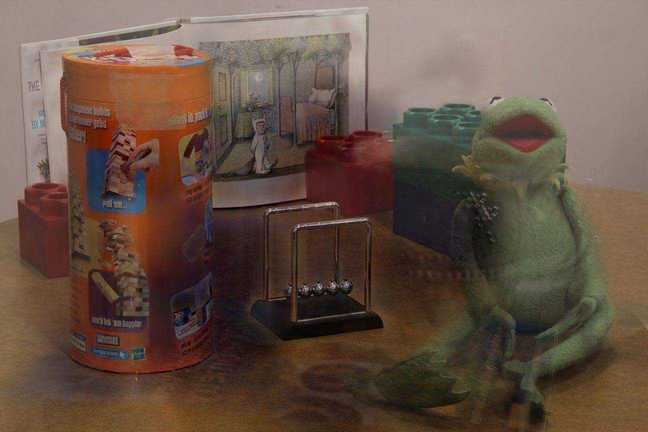} &  
    \includegraphics[width=\imagewidth]{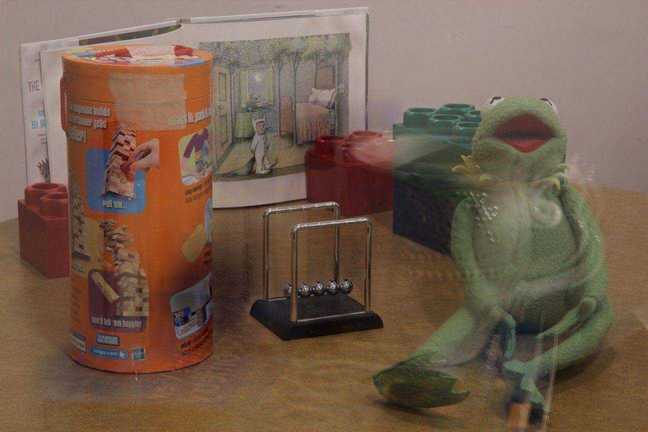} &
    \includegraphics[width=\imagewidth]{Figures/results/reflection/PAMI_online/00001B2_norm.jpg} 
    \\
    \raisebox{20mm}{Representative input frame} & 
    \raisebox{3\normalbaselineskip}[0pt][0pt]{\rotatebox[origin=c]{90}{Reflection}} &  
    \includegraphics[width=\imagewidth]{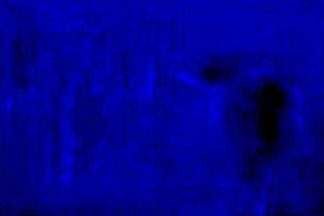} & 
    \includegraphics[width=\imagewidth]{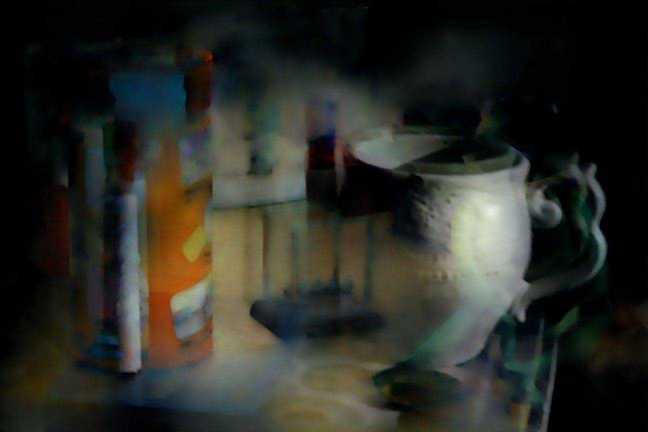} &  
    \includegraphics[width=\imagewidth]{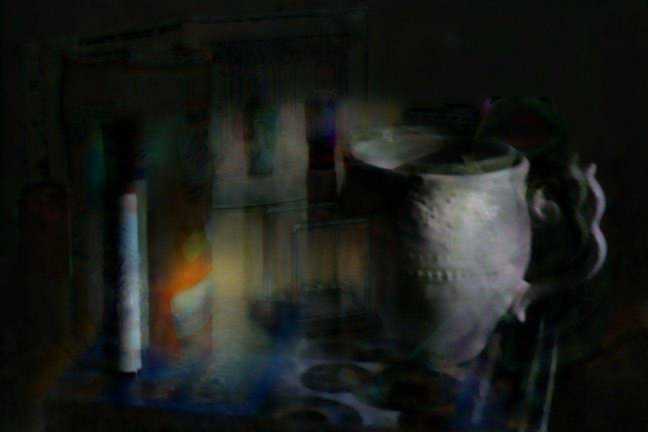} &
    \includegraphics[width=\imagewidth]{Figures/results/reflection/PAMI_online/00001F2_norm.jpg} 
    \\
    \includegraphics[width=\imagewidth]{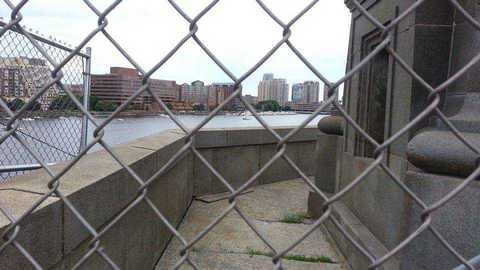} & 
    \raisebox{2.6\normalbaselineskip}[0pt][0pt]{\rotatebox[origin=c]{90}{Background}} &
    \includegraphics[width=\imagewidth]{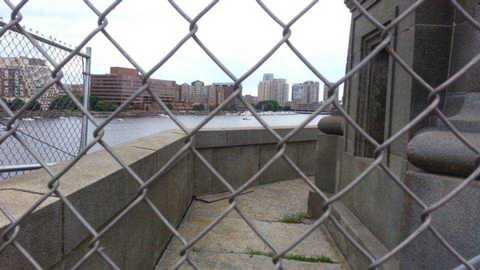}  &
    \includegraphics[width=\imagewidth]{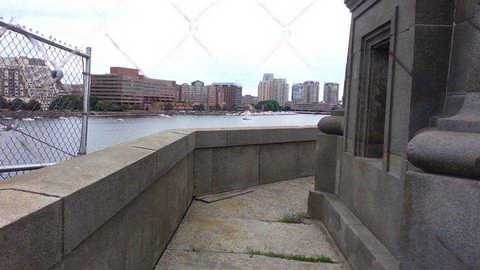} &   \includegraphics[width=\imagewidth]{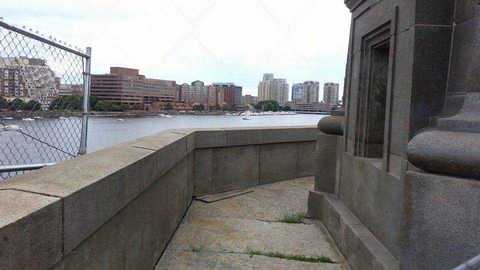}  &
    \includegraphics[width=\imagewidth]{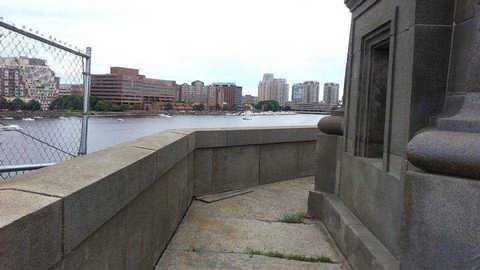} 
    \\
    \raisebox{17mm}{Representative input frame} & 
    \raisebox{2.6\normalbaselineskip}[0pt][0pt]{\rotatebox[origin=c]{90}{Alpha}} &  
    \includegraphics[width=\imagewidth]{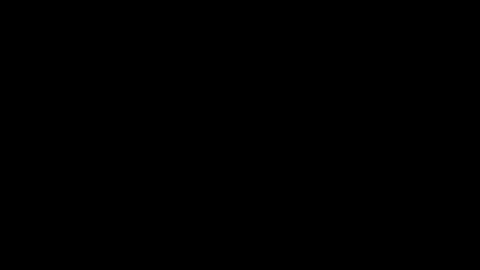} &
    \includegraphics[width=\imagewidth]{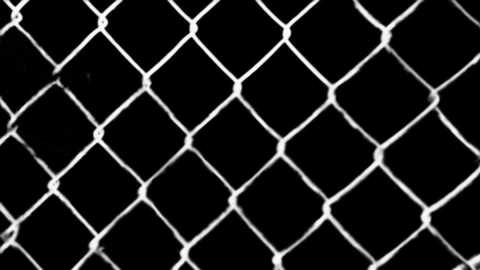} &   \includegraphics[width=\imagewidth]{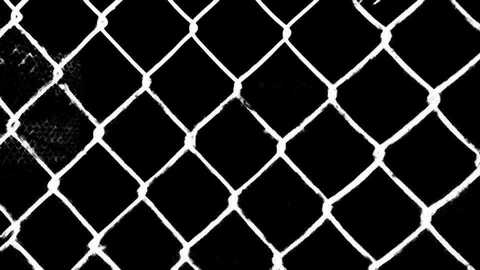}  &
    \includegraphics[width=\imagewidth]{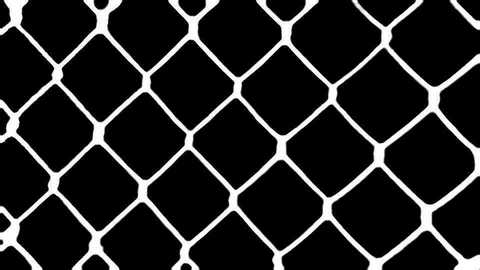} 
    \\
    \toprule
    \tb{Pre-training}        & & $\times$ & $\checkmark$ & $\checkmark$ & $\checkmark$ \\
    \tb{Online optimization} & & $\checkmark$ & $\times$ & $\checkmark$ & $\checkmark$ \\
    \tb{Meta-learning}       & & $\times$ & $\times$ & $\times$ & $\checkmark$ \\
    \bottomrule
\end{tabular}
\caption{\textbf{Effect of pre-training, online optimization, and meta-learning.} All three steps are crucial to achieving high-quality results.} 
\label{fig:online}
\end{figure*}

\begin{figure}
\centering
\footnotesize
\renewcommand{\tabcolsep}{1pt} %
\renewcommand{\arraystretch}{1} %
\newcommand{\imagewidth}{0.45\columnwidth}
\newcommand{\patchwidth}{0.23\columnwidth}
\begin{tabular}{cccc}
    \multicolumn{2}{c}{\includegraphics[width=\imagewidth]{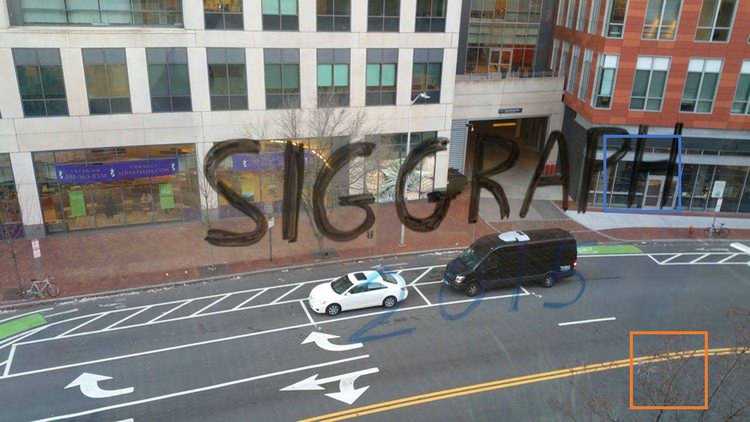}} &
    \multicolumn{2}{c}{\includegraphics[width=\imagewidth]{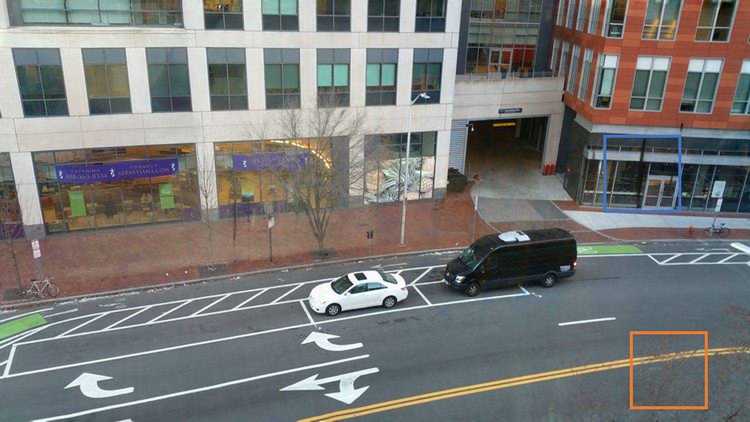}} 
    \\ 
    \includegraphics[width=\patchwidth]{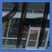} &
    \includegraphics[width=\patchwidth]{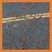} &
    \includegraphics[width=\patchwidth]{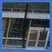} &
    \includegraphics[width=\patchwidth]{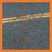} 
    \\
    \multicolumn{2}{c}{\makecell{Input}} & 
    \multicolumn{2}{c}{Xue~\etal~\cite{xue2015computational}} 
    \\
    \multicolumn{2}{c}{\includegraphics[width=\imagewidth]{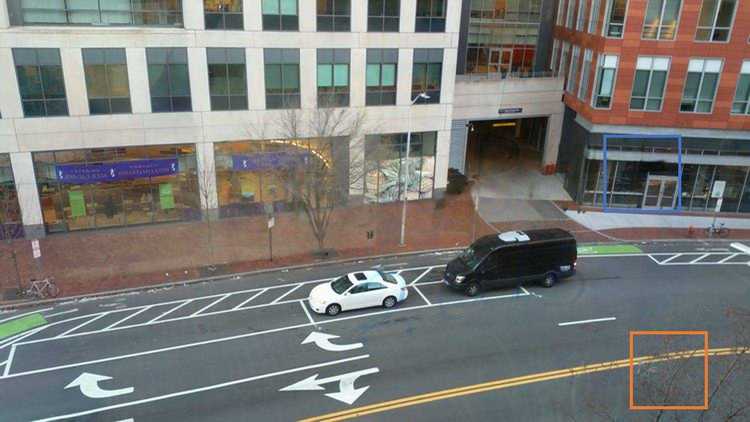}} & 
    \multicolumn{2}{c}{\includegraphics[width=\imagewidth]{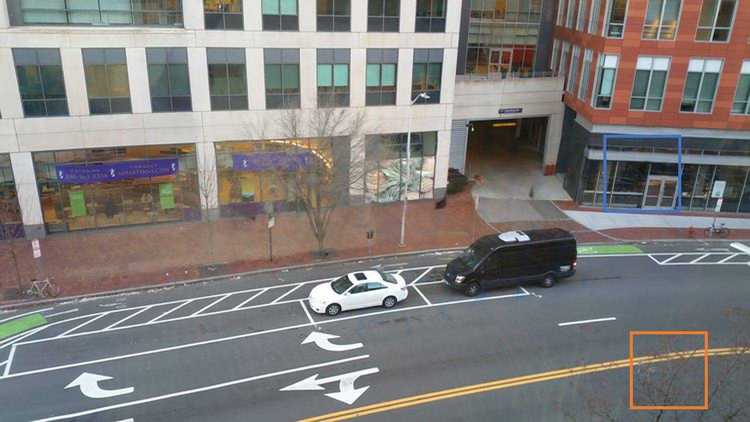}}
    \\
    \includegraphics[width=\patchwidth]{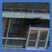} &
    \includegraphics[width=\patchwidth]{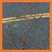} &
    \includegraphics[width=\patchwidth]{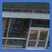} &
    \includegraphics[width=\patchwidth]{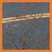} 
    \\
    \multicolumn{2}{c}{Liu~\etal~\cite{liu2020learning}} &
    \multicolumn{2}{c}{Ours}
\end{tabular}
\caption{\textbf{Occlusion removal.} The proposed method can also be applied to other obstruction removal tasks, e.g., \revision{adherent raindrop}, fence, and occlusion. 
} 
\label{fig:obstruction_visual_2}
\end{figure}

\begin{figure}
\centering
\scriptsize
\renewcommand{\tabcolsep}{1pt} %
\renewcommand{\arraystretch}{1} %
\resizebox{\columnwidth}{!}{%
\begin{tabular}{cccccccc}
\multicolumn{2}{c}{\includegraphics[width=0.13\textwidth]{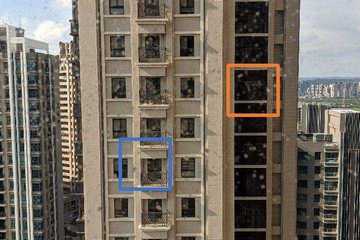}} & \multicolumn{2}{c}{\includegraphics[width=0.13\textwidth]{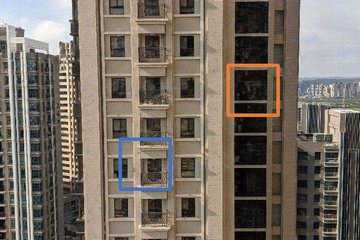}} & \multicolumn{2}{c}{\includegraphics[width=0.13\textwidth]{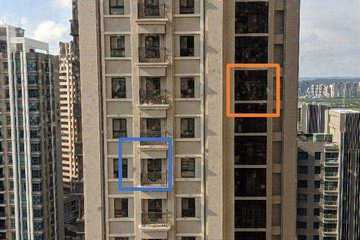}} & \multicolumn{2}{c}{\includegraphics[width=0.13\textwidth]{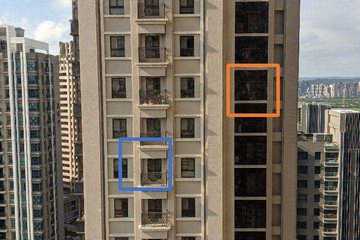}} \\
\includegraphics[width=0.063\textwidth]{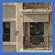} & \includegraphics[width=0.063\textwidth]{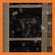} & \includegraphics[width=0.063\textwidth]{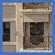} & \includegraphics[width=0.063\textwidth]{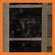} & \includegraphics[width=0.063\textwidth]{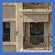} & \includegraphics[width=0.063\textwidth]{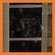} & \includegraphics[width=0.063\textwidth]{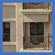} & \includegraphics[width=0.063\textwidth]{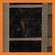} \\
\multicolumn{2}{c}{\includegraphics[width=0.13\textwidth]{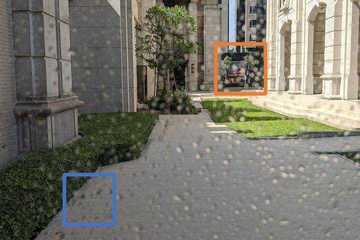}} & \multicolumn{2}{c}{\includegraphics[width=0.13\textwidth]{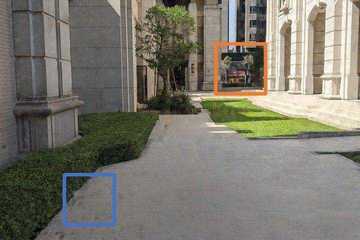}} & \multicolumn{2}{c}{\includegraphics[width=0.13\textwidth]{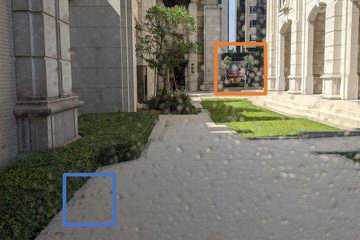}} & \multicolumn{2}{c}{\includegraphics[width=0.13\textwidth]{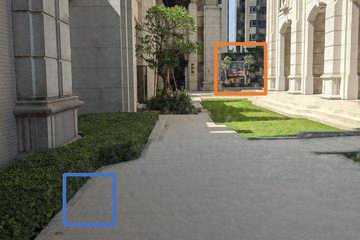}} \\
\includegraphics[width=0.063\textwidth]{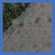} & \includegraphics[width=0.063\textwidth]{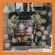} & \includegraphics[width=0.063\textwidth]{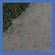} & \includegraphics[width=0.063\textwidth]{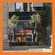} & \includegraphics[width=0.063\textwidth]{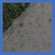} & \includegraphics[width=0.063\textwidth]{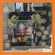} & \includegraphics[width=0.063\textwidth]{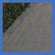} & \includegraphics[width=0.063\textwidth]{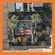} \\
\multicolumn{2}{c}{\includegraphics[width=0.13\textwidth]{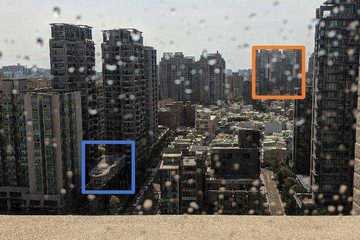}} & \multicolumn{2}{c}{\includegraphics[width=0.13\textwidth]{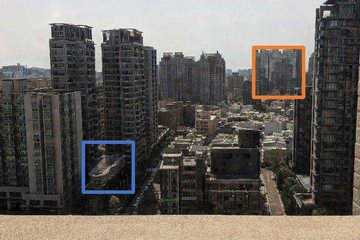}} & \multicolumn{2}{c}{\includegraphics[width=0.13\textwidth]{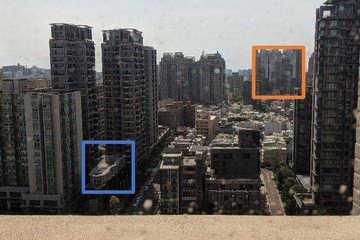}} & \multicolumn{2}{c}{\includegraphics[width=0.13\textwidth]{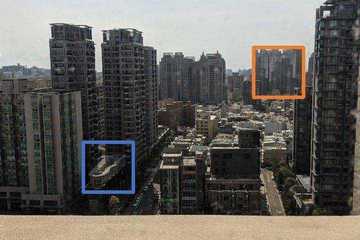}} \\
\includegraphics[width=0.063\textwidth]{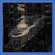} & \includegraphics[width=0.063\textwidth]{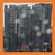} & \includegraphics[width=0.063\textwidth]{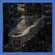} & \includegraphics[width=0.063\textwidth]{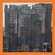} & \includegraphics[width=0.063\textwidth]{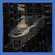} & \includegraphics[width=0.063\textwidth]{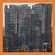} & \includegraphics[width=0.063\textwidth]{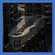} & \includegraphics[width=0.063\textwidth]{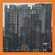} \\
\multicolumn{2}{c}{\includegraphics[width=0.13\textwidth]{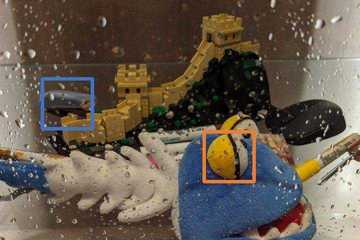}} & \multicolumn{2}{c}{\includegraphics[width=0.13\textwidth]{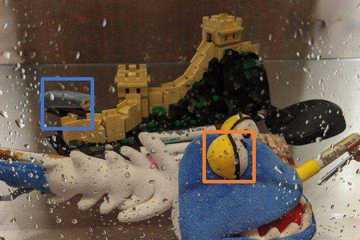}} & \multicolumn{2}{c}{\includegraphics[width=0.13\textwidth]{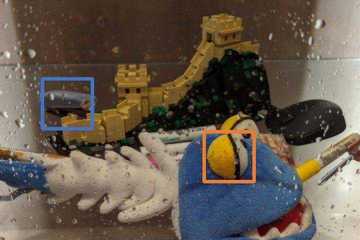}} & \multicolumn{2}{c}{\includegraphics[width=0.13\textwidth]{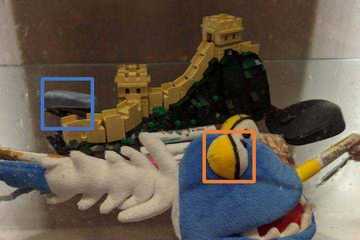}} \\
\includegraphics[width=0.063\textwidth]{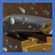} & \includegraphics[width=0.063\textwidth]{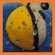} & \includegraphics[width=0.063\textwidth]{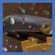} & \includegraphics[width=0.063\textwidth]{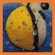} & \includegraphics[width=0.063\textwidth]{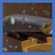} & \includegraphics[width=0.063\textwidth]{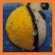} & \includegraphics[width=0.063\textwidth]{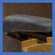} & \includegraphics[width=0.063\textwidth]{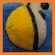} \\
\multicolumn{2}{c}{Input} & \multicolumn{2}{c}{DeRaindrop~\cite{qian2018attentive}} & \multicolumn{2}{c}{Liu et al.~\cite{liu2020learning}} & \multicolumn{2}{c}{Ours} \\
\end{tabular}%
}
\caption{\textbf{\revisionsecond{Visual comparisons on scenes with dense adherent raindrops.}}}
\label{fig:raindrop_visual_compare}
\end{figure}

\vspace{\paramargin}
{\flushleft {\bf Real sequences.}}
In \figref{reflection_visual_1}, we present visual comparisons of real input sequences from~\cite{xue2015computational}.
Our method is able to separate the reflection layers better and reconstruct clearer and sharper background images than existing approaches~\cite{alayrac2019visual, li2013exploiting, xue2015computational, liu2020learning}.
In addition, we capture another 35 real sequences using iPhone 11 and Google Pixel 3.
Some of the sequences contain non-planar background or moving objects in the scenes, which make these sequences particularly challenging.
In \figref{reflection_visual_2}, we present visual comparisons with~\cite{li2013exploiting, guo2014robust, alayrac2019visual, liu2020learning} on self-captured real input sequences. 
\figref{obstruction_visual_2} shows one example where the inputs contain \emph{semi-transparant} obstruction such as texts on the glass.
Our method can remove the obstruction layer and reconstruct clear background images.
Our method can also be applied to remove \revisionsecond{dense} static water drops that attach to the glass. 
\figref{raindrop_visual_compare} shows the visual comparisons between our method and a state-of-the-art \revisionsecond{adherent} raindrop removal method DeRaindrop~\cite{qian2018attentive}.
Our method can better remove the raindrops and maintain details in the recovered background in the scenarios that the method targets for.

\begin{table*}
\caption{
        \textbf{Ablations.} We analyze the design choices of the proposed method and report the validation loss of~\eqref{eq:validation} on the synthetic reflection-background Vimeo-90k test set. %
    }
\label{tab:ablation}
\centering
\footnotesize
\begin{tabular}{p{5cm}p{6.2cm}p{5.8cm}}
(a) \textbf{Initial flow decomposition}: Predicting uniform flow fields as initialization achieves better results. & 
(b) \textbf{Fusion method}: Our image reconstruction network recovers better background/reflection than temporal mean/median filtering. & 
(c) \textbf{Model training}: Both the network pre-training and online optimization are important to the performance of our method.
\end{tabular}

\begin{tabular}{p{5cm}p{6.2cm}p{5.8cm}}[t]
\makecell{
    \begin{tabular}{l|c}
    \toprule
    Flow initialization & Loss \\
    \midrule
    \emph{Zero} initialization & 0.478 \\
    \emph{Dense} flow field & 0.236 \\
    \emph{Uniform} flow field (Ours) & \pmb{0.197} \\
    \bottomrule
    \end{tabular}
}
& 
\makecell{
    \begin{tabular}{l|c}
    \toprule
    Image fusion method & Loss \\
    \midrule
    Temporal mean filtering & 0.652 \\
    Temporal median filtering & 0.555 \\
    Image reconstruction network (Ours) & \pmb{0.197} \\
    \bottomrule
    \end{tabular}
} 
& 
\makecell{
    \begin{tabular}{cc|c}
    \toprule
    \makecell{Online optimization} & Pre-training & \makecell{Loss} \\
    \midrule
    \checkmark & - & 0.468 \\
    - & \checkmark & 0.283 \\
    \checkmark & \checkmark & \pmb{0.197} \\
    \bottomrule
    \end{tabular}
}
\end{tabular}
\end{table*}

\subsection{Analysis and Discussions}
\label{sec:analysis}

\vspace{\paramargin}
{\flushleft {\bf Initial flow decomposition.}}
We demonstrate that the uniform flow field initialization plays an important role in our algorithm.
We train our model with the following settings: 
1) removing the initial flow decomposition network, where the flows at the coarsest level are set to zero, and 
2) predicting spatially-varying dense flow fields as the initial flows.
\tabref{ablation}(a) reports the validation loss of~\eqref{eq:validation} on our Vimeo-90k validation set, where the model with uniform flow prediction achieves a lower validation loss compared to other alternatives.
Initializing the flow fields to zero makes it difficult for the following levels to decompose the background and reflection layers.

\begin{table}
    \centering
    \footnotesize
    \caption{
        \textbf{\revision{Ablation study on the number of pyramid levels.}}
    }
    \label{tab:ablation_coarse_to_fine}
    \begin{tabular}{c|c}
        \toprule
        Pyramid level & Validation loss \\
        \midrule
        1 (without coarse-to-fine) & 0.382 \\
        2 & 0.321\\
        3 & 0.308 \\
        4 & 0.289 \\
        5 & 0.283 \\
        6 & 0.281 \\
        \bottomrule
    \end{tabular} \\
\end{table}

\begin{figure}
\centering
\footnotesize
\renewcommand{\tabcolsep}{1pt} %
\renewcommand{\arraystretch}{1} %
\begin{tabular}{ccccc}
     \multicolumn{1}{c|}{\includegraphics[width=0.23\columnwidth]{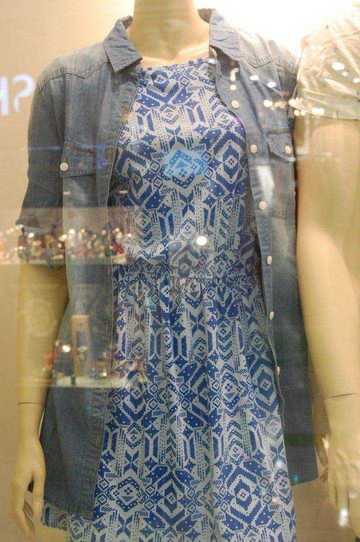}} &  \raisebox{5.0\normalbaselineskip}[0pt][0pt]{\rotatebox[origin=c]{90}{Background}} &
     \includegraphics[width=0.23\columnwidth]{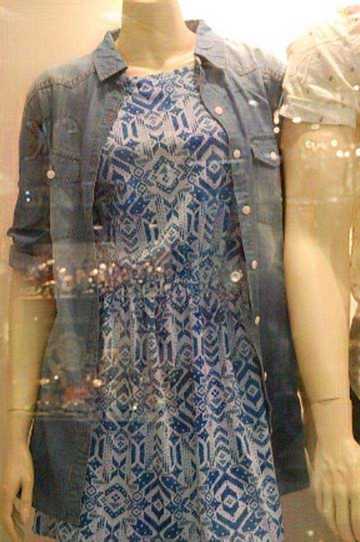} &
     \includegraphics[width=0.23\columnwidth]{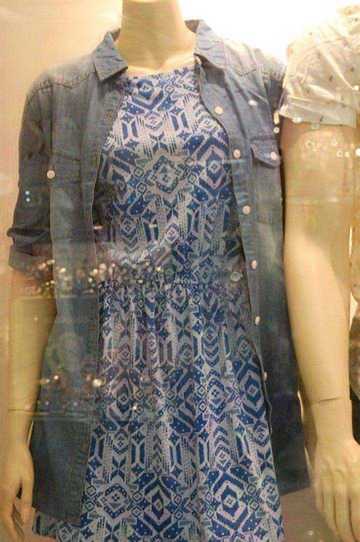} & \includegraphics[width=0.23\columnwidth]{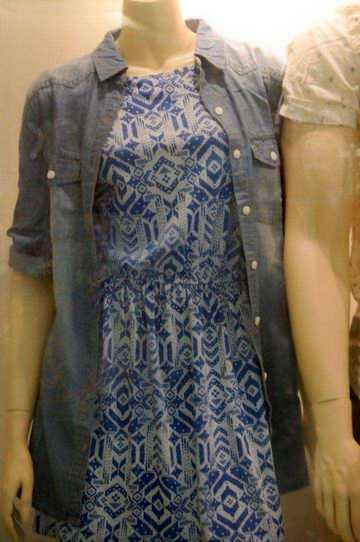} \\
    \multicolumn{1}{c|}{\includegraphics[width=0.23\columnwidth]{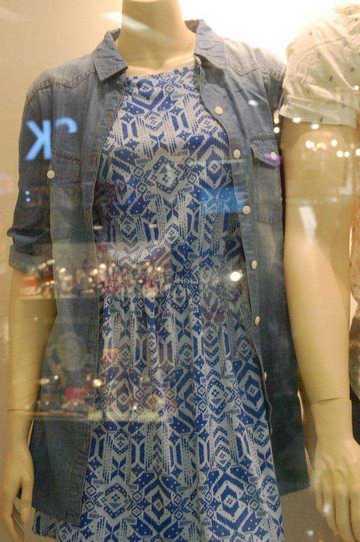}} & \raisebox{5.0\normalbaselineskip}[0pt][0pt]{\rotatebox[origin=c]{90}{Reflection}} &  \includegraphics[width=0.23\columnwidth]{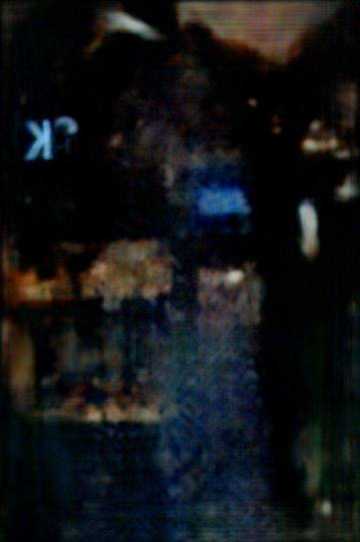} & \includegraphics[width=0.23\columnwidth]{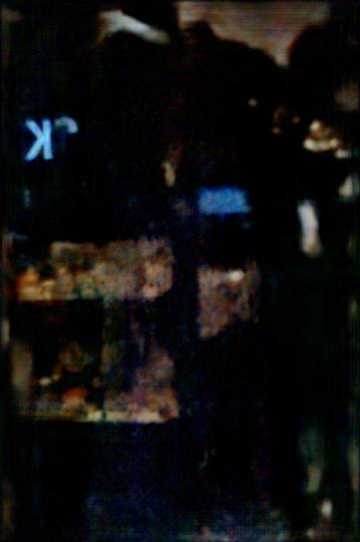} & \includegraphics[width=0.23\columnwidth]{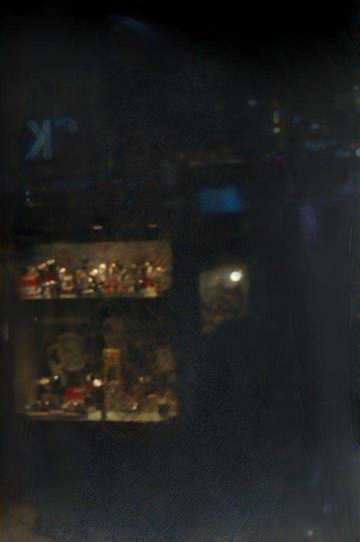} \\
    (a) Inputs & & (b) Single-scale & (c) 3 scales & (d) 5 scales \\
\end{tabular}
\caption{\revision{\textbf{Visual comparison on different pyramid levels.}}} 
\label{fig:single_scale}
\end{figure}

\begin{figure}[t]
\centering
\includegraphics[width=0.5\textwidth]{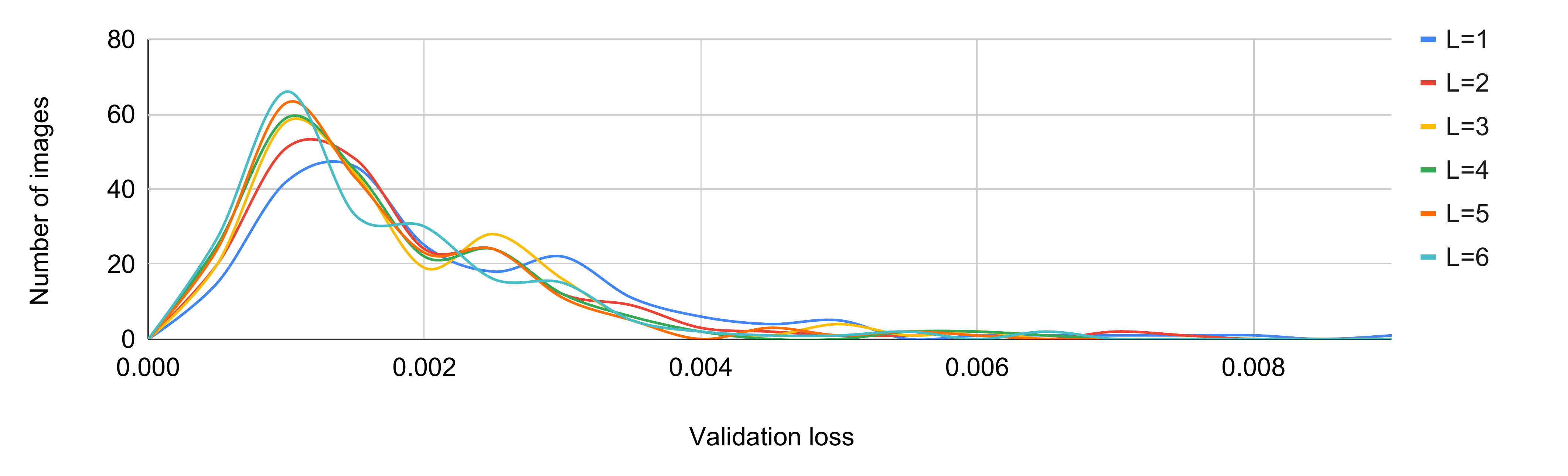}
\caption{\textbf{\revisionsecond{Distributions of validation losses with different numbers of pyramid levels.}}} 
\label{fig:validation_loss}
\vspace{4mm}
\begin{tabular}{c}
\includegraphics[width=0.45\textwidth]{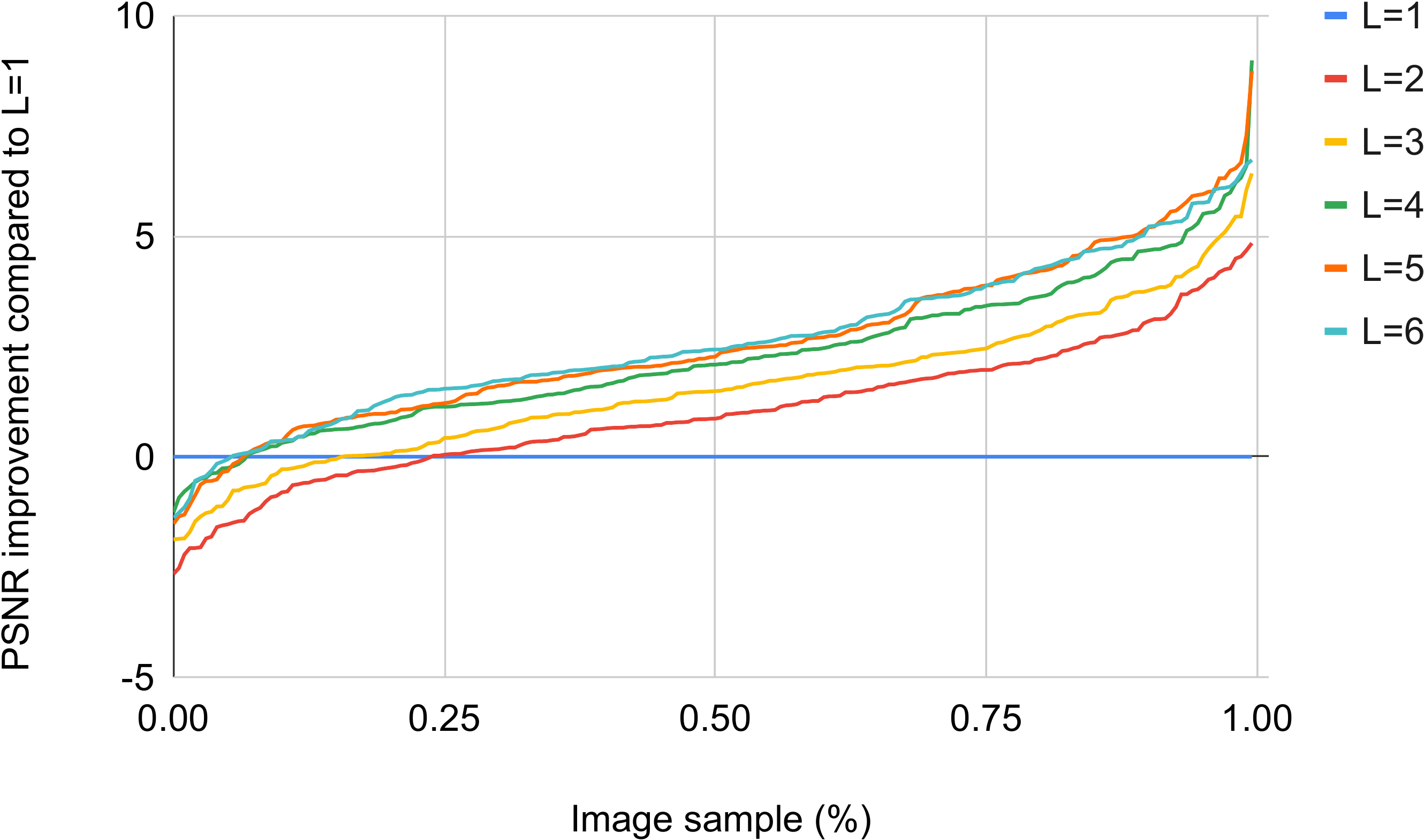} %
\end{tabular}
\vspace*{-2mm}
\caption{\textbf{\revisionsecond{PSNR improvement compared to $L=1$.}}} 
\label{fig:positive_plot_psnr}
\end{figure}

\begin{figure}[t]
\centering
\footnotesize
\renewcommand{\tabcolsep}{1pt} %
\renewcommand{\arraystretch}{1} %
\begin{tabular}{cc}
    \includegraphics[width=0.49\columnwidth]{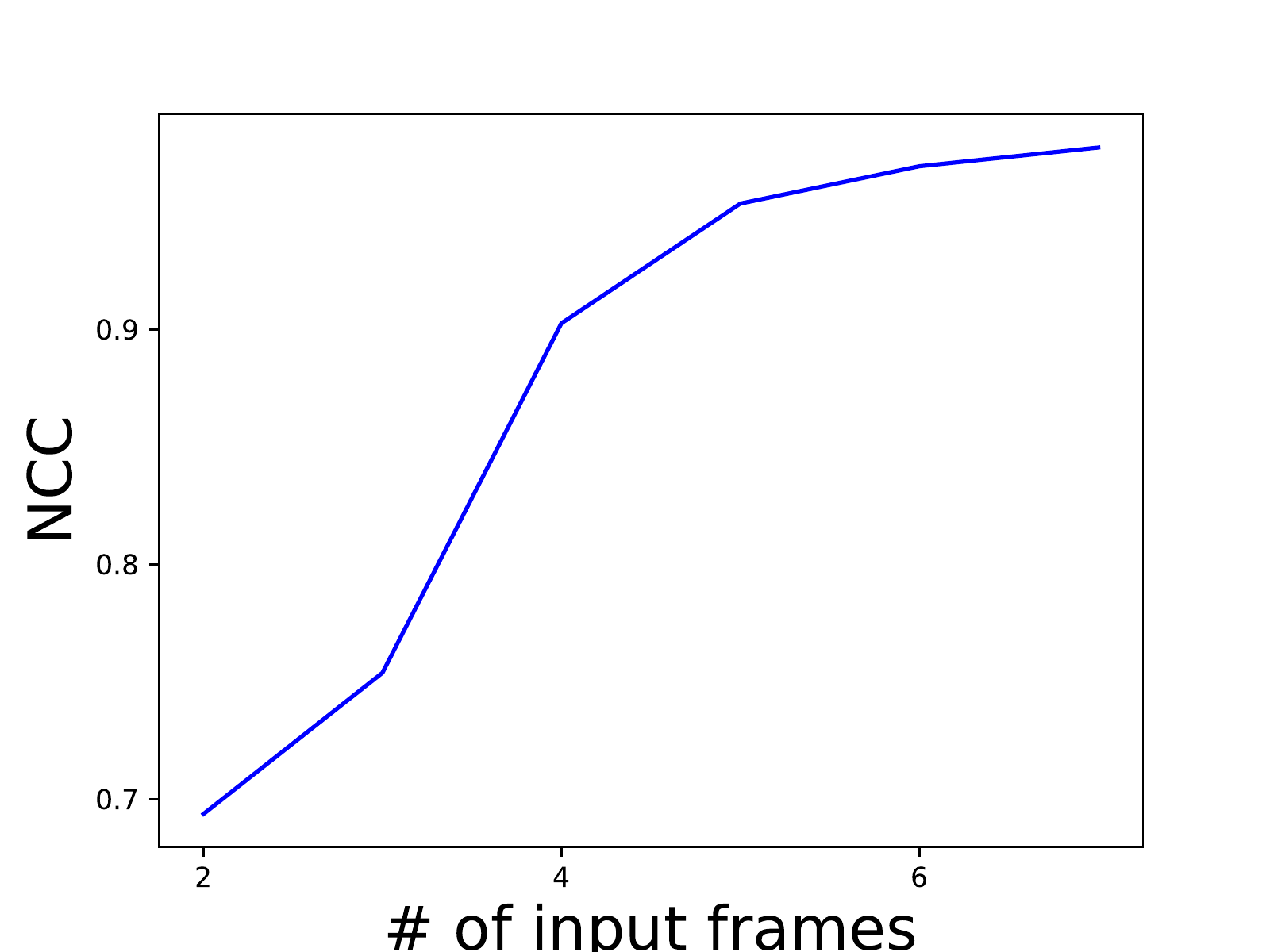} & 
    \includegraphics[width=0.49\columnwidth]{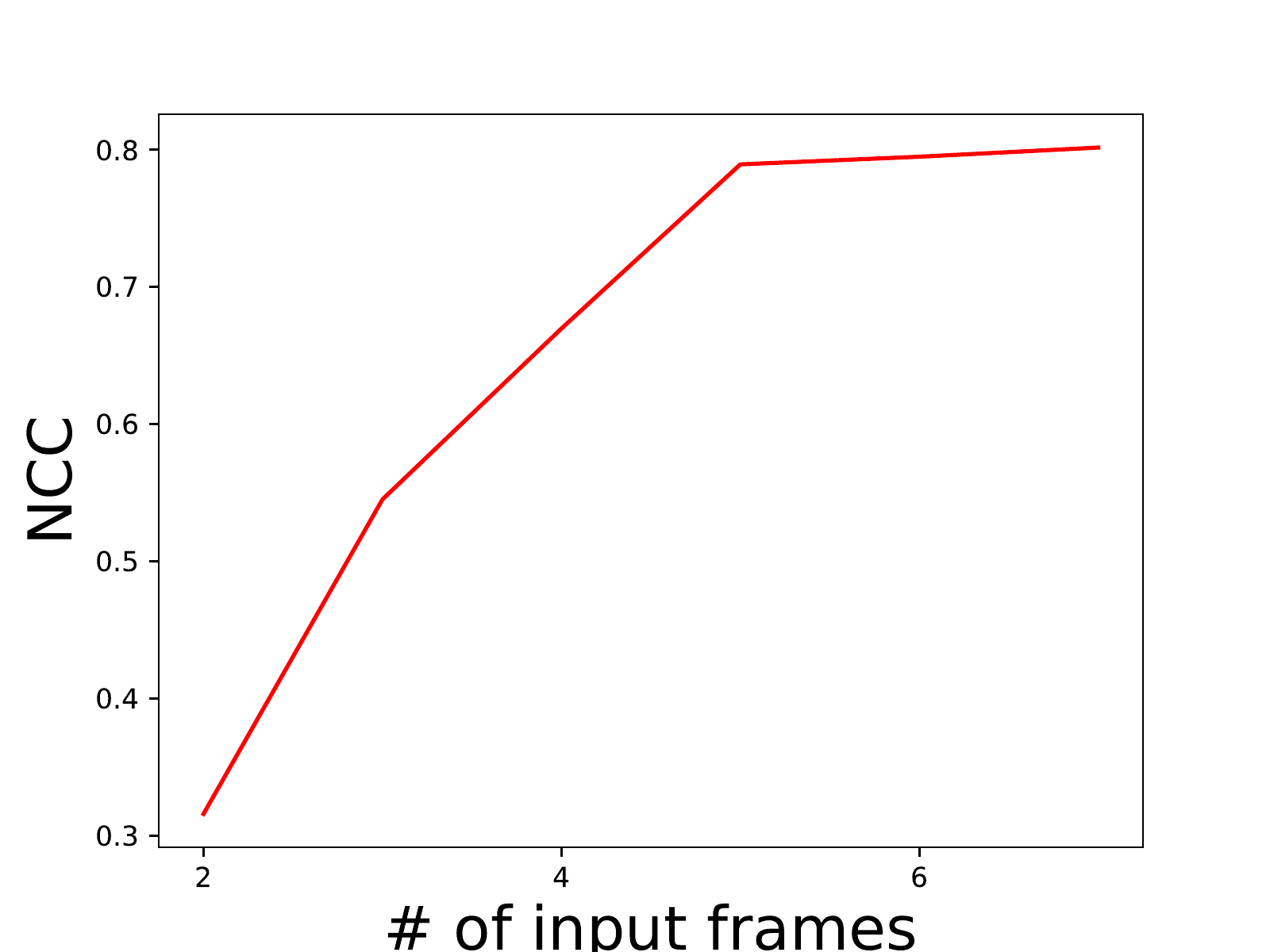} \\
    (a) NCC of the background layer &
    (b) NCC of the obstruction layer \\
    \includegraphics[width=0.49\columnwidth]{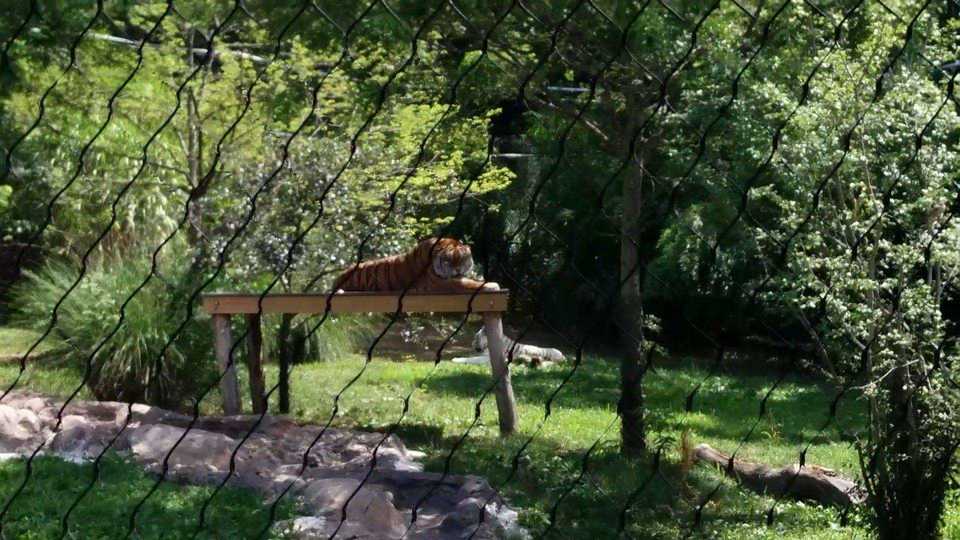} & 
    \includegraphics[width=0.49\columnwidth]{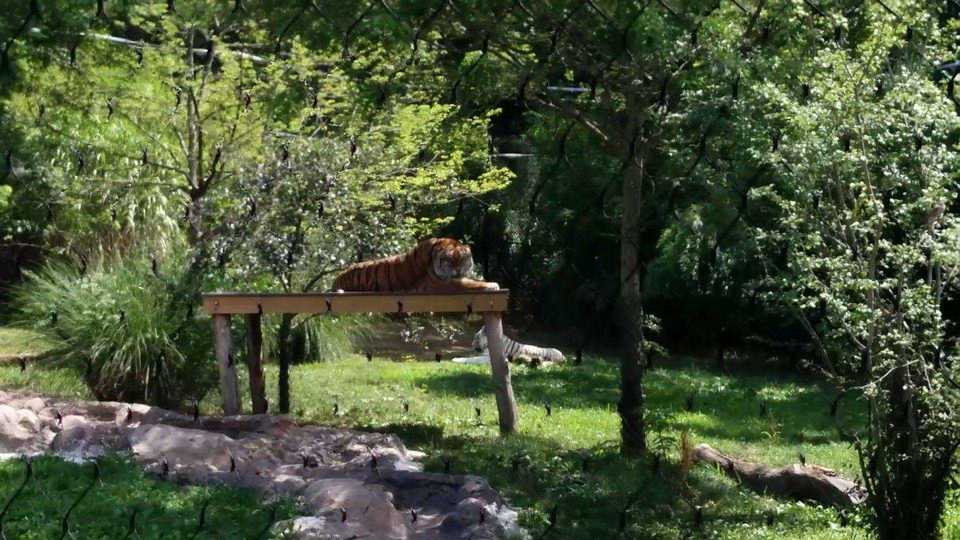} \\
    (c) 3 input frames & 
    (d) 5 input frames \\
    \includegraphics[width=0.49\columnwidth]{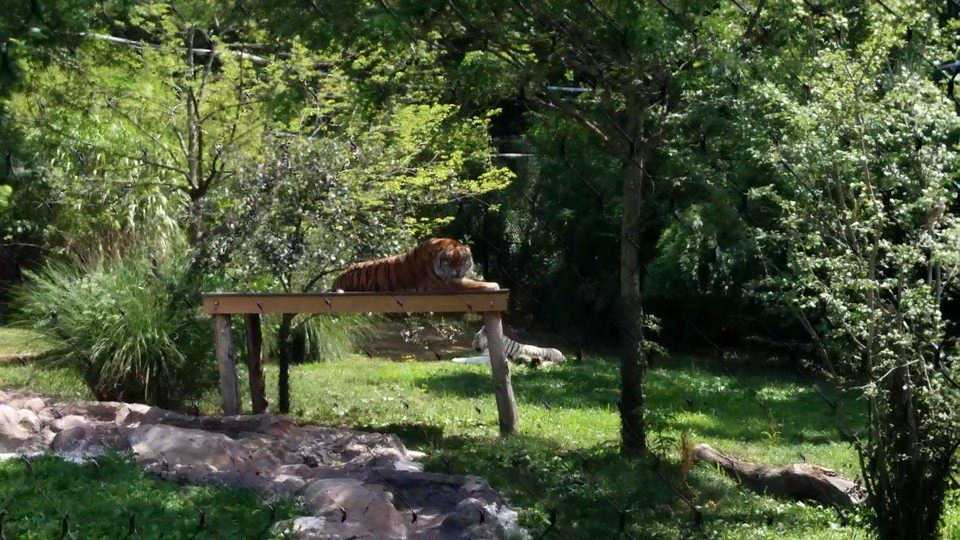} & 
    \includegraphics[width=0.49\columnwidth]{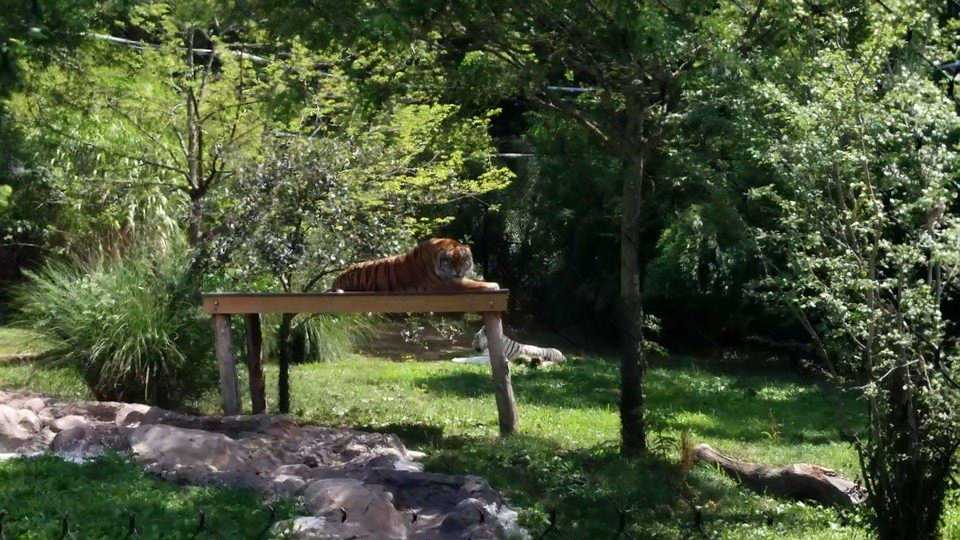} \\
    (e) 7 input frames & 
    (f) 9 input frames \\
\end{tabular}
\caption{\textbf{Effect of number of input frames.} Introducing new frames as input is always likely to improve the result accuracy of reconstructed background and obstruction images. %
}
\label{fig:num_input_frames}
\end{figure}

\vspace{\paramargin}
{\flushleft {\bf Image reconstruction network.}}
To demonstrate the effectiveness of the image reconstruction network, we replace it with a simple temporal filter to fuse the neighbor frames after warping and aligning them with the optical flows.
We show in \tabref{ablation}(b) that both the temporal mean and median filters result in large errors (in terms of the validation loss of~\eqref{eq:validation}) as the errors are accumulated across levels.

\vspace{\paramargin}
{\flushleft {\bf Online optimization.}}
\tabref{ablation}(c) shows that both the network pre-training with synthetic data and online optimization with real data are beneficial to the performance of our model.
In \figref{online}, we show that the model without pre-training cannot separate the reflection (or fence) well on the real input sequence.
Without online optimization, the background image contains residuals from the reflection layer.
After online optimization, our method is able to reconstruct both background and reflection layers well.

\vspace{\paramargin}
{\flushleft {\bf Pyramid level.}}
\revision{
\tabref{ablation_coarse_to_fine} shows the ablation study on the number of pyramid levels $L$. 
Without the coarse-to-fine strategy ($L=1$), the method gives the worst validation loss. 
By increasing the pyramid levels, the loss decreases.   
We choose $L=5$ because (1) the performance nearly saturates after five levels and (2) more levels consume more memory.
\figref{single_scale} compares the visual results of the reconstructed background and reflection with different pyramid levels.
With more pyramid levels, our method can reconstruct the background more faithfully. 
}
\revisionsecond{
In~\figref{validation_loss}, we plot the distributions of the validation losses on the 200 images in the validation set.
The plot clearly shows that the validation losses distribute more toward the left end (i.e., more images with lower validation losses) when using more pyramid levels. 
In~\figref{positive_plot_psnr}, we plot the PSNR improvement of the coarse-to-fine (multi-scale) scheme compared to the single-scale baseline.
The plot shows that with the pyramid level $L=2$, about 75\% of the images in the validation set have higher PSNR than those with only a single scale. 
With more pyramid levels, the winning ratio further increases. 
With the pyramid level $L=5$, about 50\% of the images have more than 2.5dB PSNR improvement compared to the single-scale scheme.
About 10\% of the images do not gain improvements from the multi-scale scheme with the pyramid level $L=5$.
}

\begin{table}[t]
\caption{\textbf{Effect of realistic training data generation.}}
\label{tab:realistic_training_data_generation}
\centering
    \footnotesize
    \begin{tabular}{lcc}
    \toprule
    Method & $B$ & $R$ \\
    \midrule
    Data generation used in~\cite{liu2020learning} & 0.9390 & 0.5886 \\
    Ours realistic training data generation & \pmb{0.9428} & \pmb{0.6032} \\
    \bottomrule
    \end{tabular}
\end{table}

\vspace{\paramargin}
{\flushleft {\bf Realistic training data generation.}}
We conduct an experiment to demonstrate the impact of training data generation using controlled sequences \emph{Stone} and \emph{Toy}.
\tabref{realistic_training_data_generation} shows that the NCC scores of reconstructed background and reflection layers are higher by using the proposed realistic training data generation.
By training with proposed realistic training data generation, the reconstructed background and reflection layers are much sharper and contain more details.

\vspace{\paramargin}
{\flushleft {\bf Number of input frames.}}
We analyze the effect of the number of input frames on the reconstruction quality using our synthetic sequences.
\figref{num_input_frames} shows that the results of fence removal are better by giving additional input frames. 
\figref{num_input_frames} also shows that adding additional frames as input leads to improvement in NCC of the reconstructed background and reflection layers.

\vspace{\paramargin}
{\flushleft {\bf Input features.}}
\revision{\tabref{ablation_input_features} studies the impact of the five input features. 
Among all the input features, the most important is the registered frames, and the least important is the difference maps.
The combination of all five features gives the best performance.}

\begin{table*}
    \centering
    \footnotesize
    \caption{
        \revision{\textbf{Ablation study on the input features.}}
    }
    \label{tab:ablation_input_features}
    \begin{tabular}{ccccc|c}
        \toprule
        Registered frame & Difference map & Visibility mask & Upsampled background & Upsampled reflection & Validation loss \\
        \midrule
         & \checkmark & \checkmark & \checkmark & \checkmark & 0.416 \\
        \checkmark &  & \checkmark & \checkmark & \checkmark & 0.288 \\
        \checkmark & \checkmark &  & \checkmark & \checkmark & 0.294 \\
        \checkmark & \checkmark & \checkmark &  & \checkmark & 0.314 \\
        \checkmark & \checkmark & \checkmark & \checkmark &  & 0.322 \\
        \checkmark & \checkmark & \checkmark & \checkmark & \checkmark & 0.283 \\
        \bottomrule
    \end{tabular} 
\end{table*}

\begin{figure}[t]
\centering
\footnotesize
\includegraphics[width=\columnwidth]{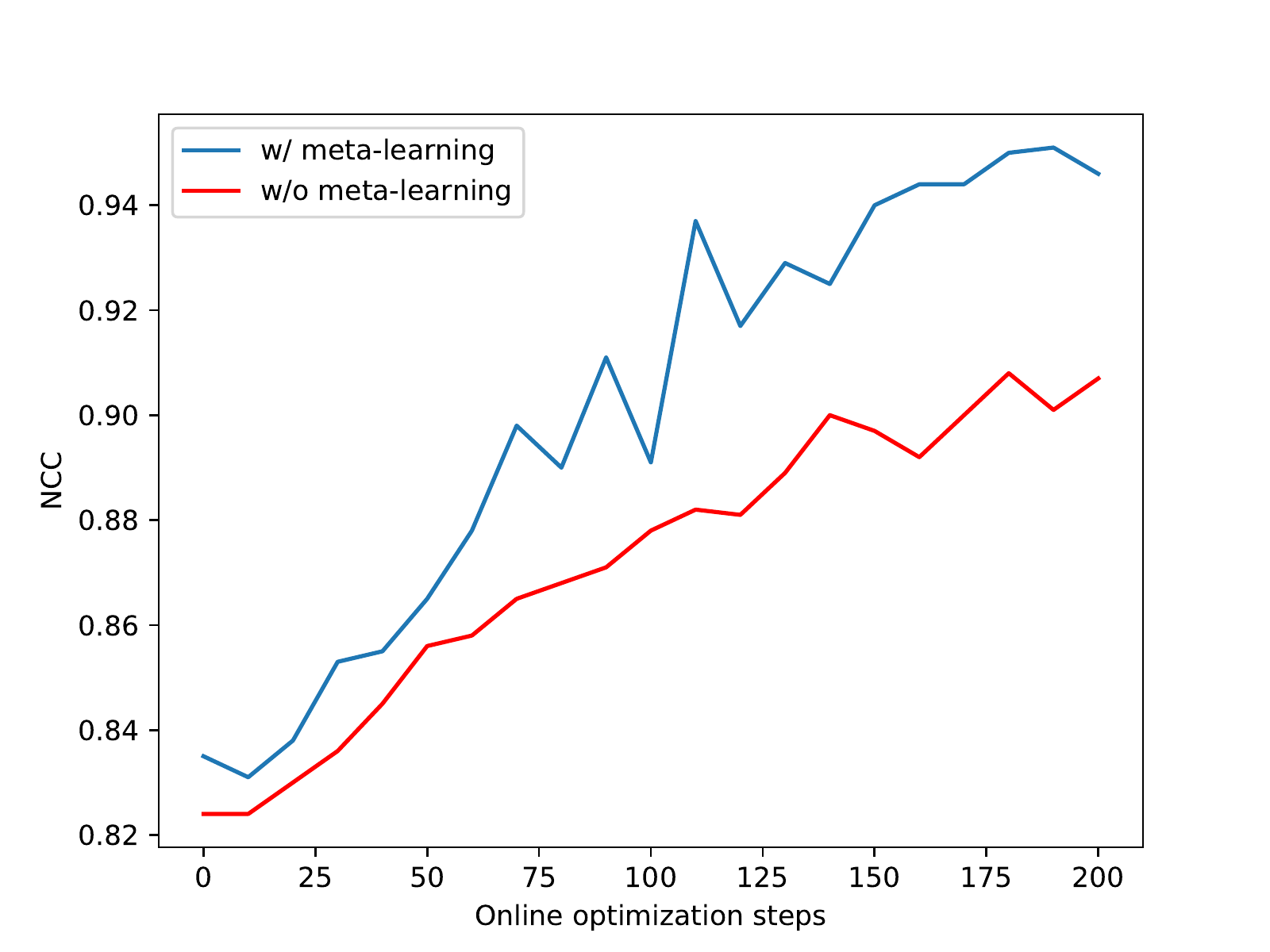}
\caption{\textbf{Effect of meta-learning.} Online optimization trained with meta-learning convergences faster than without.} 
\label{fig:maml}
\end{figure}

\vspace{\paramargin}
{\flushleft {\bf Meta-learning.}}
We use the controlled sequence \emph{Toy} to demonstrate the effectiveness of the meta-learning.
\figref{maml} shows that the model pre-trained with meta-learning (blue curve) is able to converge faster and achieve better NCC scores at the same number of online optimization steps.
With meta-learning, the number of online optimization steps can be reduced from 1,000 to 200 and achieve similar quality as in~\cite{liu2020learning}.
\figref{online} also demonstrate that the meta-learning can improve the visual quality of the reconstructed background and reflection/fence layers.
\tabref{running_time_ours} shows the comparisons of running time.

\begin{table}[t]
\caption{\textbf{Running time comparison (in seconds) of the proposed method.} With meta-learning, our model can run about $4\times$ to $5\times$ faster while achieving similar or better reconstruction quality.}
\label{tab:running_time_ours}
\centering
\footnotesize
\renewcommand{\tabcolsep}{4pt} %
\begin{tabular}{cc|ccc}
\toprule
\makecell{Online\\optimization} & \makecell{Meta-\\learning} & \makecell{QVGA\\($320\times240$)}  & \makecell{VGA\\($640\times480$)} & \makecell{720p\\($1280\times720$)} \\
\midrule
$\times$ & $\times$ & 1.107 & 2.216 & 9.857\\
$\checkmark$ & $\times$ & 66.056 & 264.227 & 929.182\\
$\checkmark$ & $\checkmark$ & 28.436 & 69.224 & 187.439\\
\bottomrule
\end{tabular}
\end{table}

\vspace{\paramargin}
{\flushleft {\bf Running time.}}
In \tabref{running_time}, we compare the execution time of two optimization-based algorithms~\cite{guo2014robust, li2013exploiting} and a recent CNN-based method~\cite{alayrac2019visual} with different input sequences resolutions on a computer with Intel Core i7-8550U CPU and NVIDIA TITAN Xp GPU.
Alayrac et al.~\cite{alayrac2019visual} use a 3D CNN architecture without explicit motion estimation, which results in a faster inference speed.
In contrast, our method computes bi-directional optical flows for every pair of input frames in a coarse-to-fine manner, which is slower but achieves much better reconstruction performance.

\begin{table}[t]
\caption{\textbf{Running time comparison (in seconds).} CPU: Intel Core i7-8550U, GPU: NVIDIA TITAN Xp. * denotes methods using GPU.}
\label{tab:running_time}
\centering
\footnotesize
\begin{tabular}{l|ccc}
\toprule
Method & \makecell{QVGA\\($320\times240$)} & \makecell{VGA\\($640\times480$)} & \makecell{720p\\($1280\times720$)} \\
\midrule
Li and Brown~\cite{li2013exploiting} & 82.591 & 388.235 & 1304.231 \\
Guo~\etal~\cite{guo2014robust} & 64.251 & 369.200 & 1129.125 \\
*Alayrac~\etal~\cite{alayrac2019visual} & 0.549 & 2.011 & 6.327 \\
*Ours & 28.436 & 69.224 & 187.439\\
\bottomrule
\end{tabular}
\end{table}

\begin{figure}[t]
\centering
\footnotesize
\begin{tabular}{ccc}
     \raisebox{5.5\normalbaselineskip}[0pt][0pt]{\rotatebox[origin=c]{90}{NCC}} &  
     \multicolumn{2}{c}{\includegraphics[width=0.9\columnwidth]{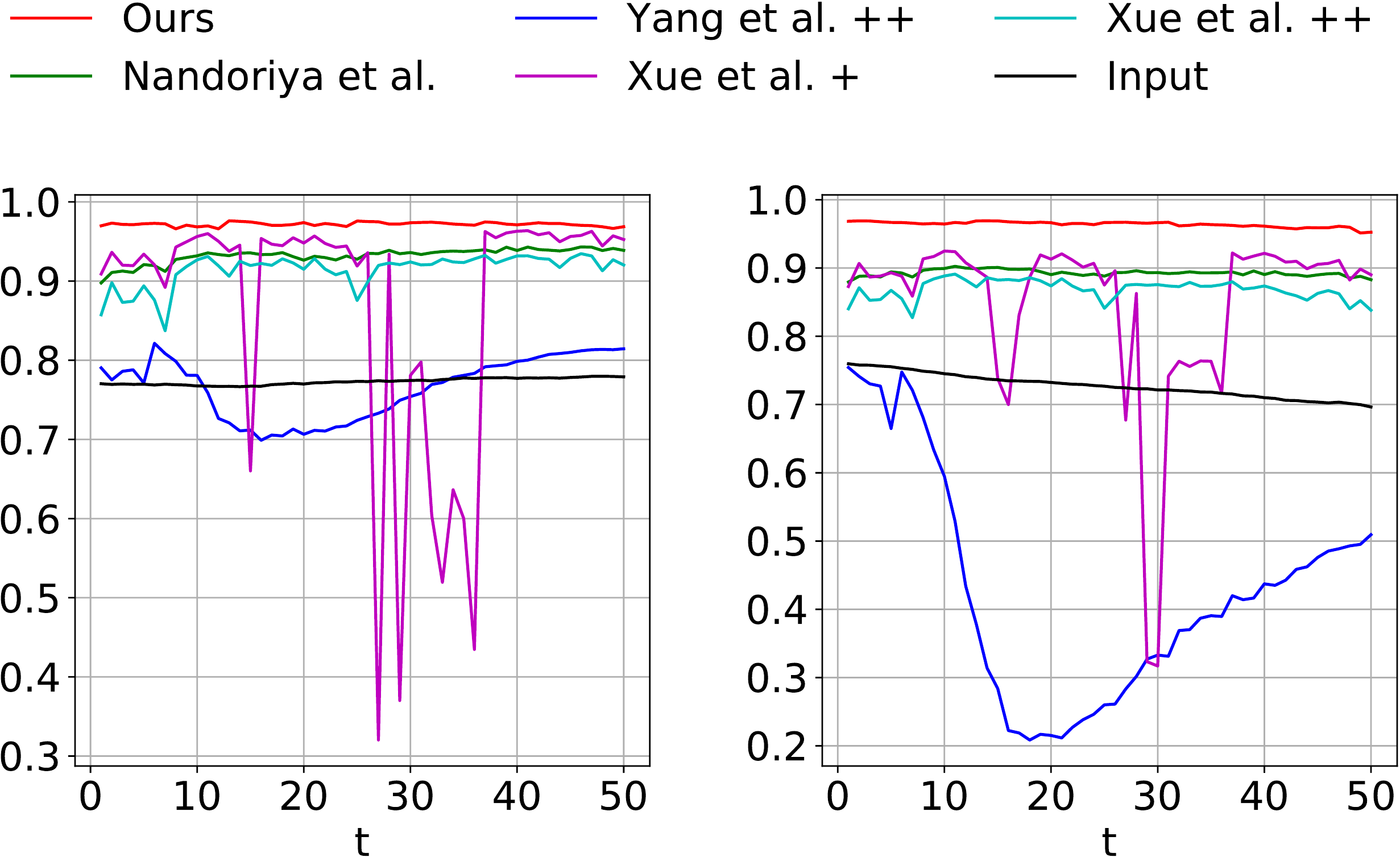}} \\
     \\
     \raisebox{5.5\normalbaselineskip}[0pt][0pt]{\rotatebox[origin=c]{90}{SSIM}} &  
     \multicolumn{2}{c}{\includegraphics[width=0.9\columnwidth]{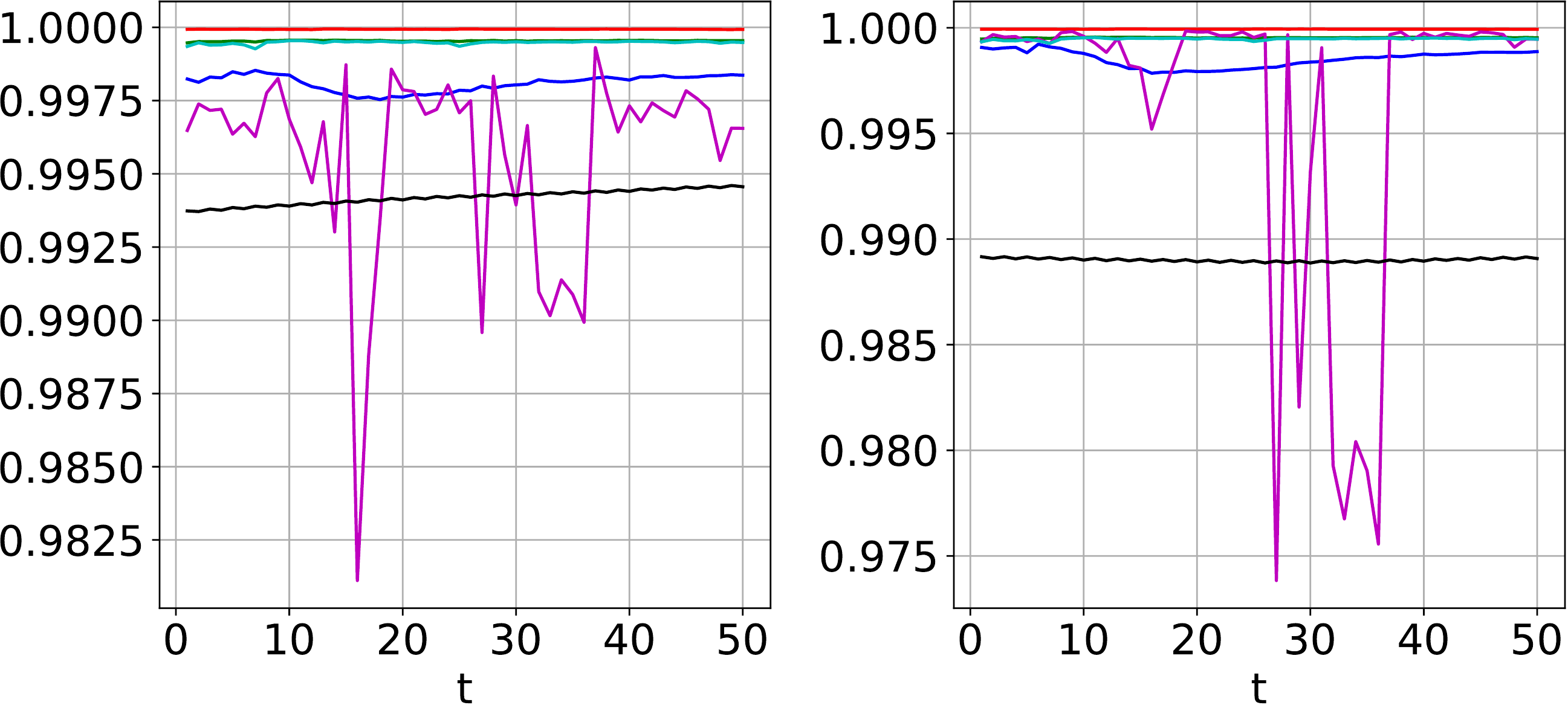}} \\
     & \enspace\enspace\enspace\enspace\enspace\enspace\enspace\enspace\enspace Background & \enspace\enspace\enspace\enspace\enspace\enspace\enspace\enspace\enspace Reflection \\
\end{tabular}
\caption{\textbf{Evaluation different reflection removal methods on a controlled synthetic sequence provided by~\cite{nandoriya2017video}.} Our method generates the best temporal coherency and layer separation. 
}
\label{fig:video_NCC_SSIM}
\end{figure}

\vspace{\paramargin}
{\flushleft {\bf Video obstruction removal.}}
The proposed method takes multiple neighboring frames as input and generates the separation results for a single reference frame at a time.
Although predicting each reference frame \emph{independently}, our method still generates temporally coherent results on the entire video.
Here, we compare our method with four video reflection removal approaches~\cite{xue2015computational, yang2016robust, nandoriya2017video} and report results in \figref{video_NCC_SSIM}.
Both the methods of Xue~\etal~\cite{xue2015computational} and Yang~\etal~\cite{yang2016robust} take multiple frames as input and generates the middle frame, similar to our model.
Xue~\etal+~\cite{xue2015computational} is an extension of~\cite{xue2015computational} which uses the moving window strategy in~\cite{yang2016robust} to improve the temporal consistency.
Both Xue~\etal++~\cite{xue2015computational} and Yang~\etal++~\cite{yang2016robust} adopt a temporal average filtering to further reduce the temporal flickering.
Nandoriya~\etal~\cite{nandoriya2017video} use a spatiotemporal optimization method to process the entire video sequence jointly.

\begin{figure}[t]
\centering
\footnotesize
\renewcommand{\tabcolsep}{1pt} %
\renewcommand{\arraystretch}{1} %
\newcommand{\imagewidth}{0.42\columwidth}
\begin{tabular}{ccc}
    \includegraphics[width=0.42\columnwidth]{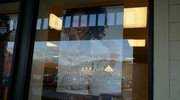} & 
    \raisebox{2.5\normalbaselineskip}[0pt][0pt]{\rotatebox[origin=c]{90}{Background}} &  
    \includegraphics[width=0.42\columnwidth]{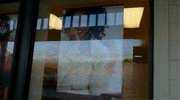} 
    \\
    \includegraphics[width=0.42\columnwidth]{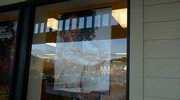} & 
    \raisebox{2.5\normalbaselineskip}[0pt][0pt]{\rotatebox[origin=c]{90}{Reflection}} &  
    \includegraphics[width=0.42\columnwidth]{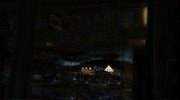} 
    \\
    Representative input frame & & Our results \\
\end{tabular}
\caption{\textbf{A failure case.} Our method fails to recover the correct flow fields for each layer, leading to ineffective reflection removal.}
\label{fig:failure}
\end{figure}

\begin{figure}[t]
\centering
\footnotesize
\renewcommand{\tabcolsep}{1pt} %
\renewcommand{\arraystretch}{1} %
\newcommand{\imagewidth}{0.42\columwidth}
\begin{tabular}{ccc}
    \includegraphics[width=0.42\columnwidth]{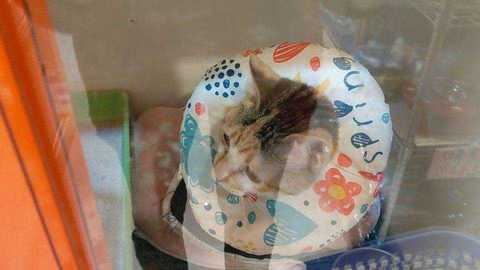} & 
    \raisebox{2.5\normalbaselineskip}[0pt][0pt]{\rotatebox[origin=c]{90}{Background}} &  
    \includegraphics[width=0.42\columnwidth]{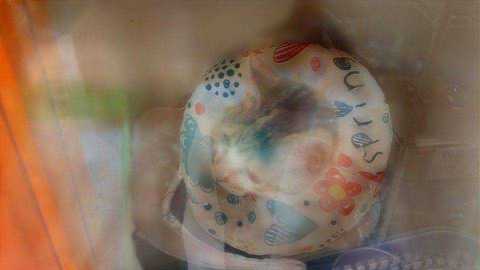} 
    \\
    \includegraphics[width=0.42\columnwidth]{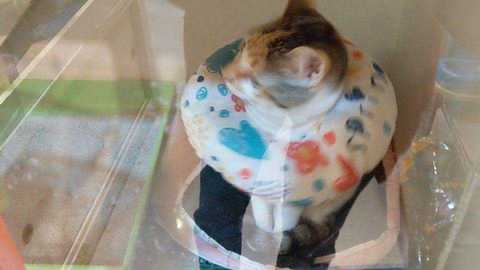} & 
    \raisebox{2.5\normalbaselineskip}[0pt][0pt]{\rotatebox[origin=c]{90}{Reflection}} &  
    \includegraphics[width=0.42\columnwidth]{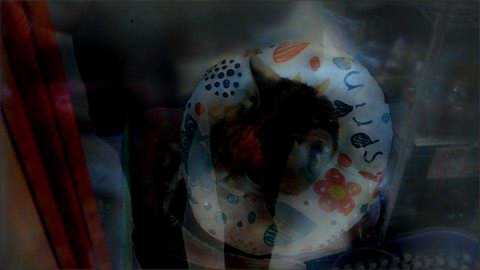} 
    \\
    Representative input frame & & Our results \\
\end{tabular}
\caption{\textbf{A failure case with a highly dynamic scene.}}
\label{fig:failure2}
\vspace{5mm}
\centering
\scriptsize
\renewcommand{\tabcolsep}{1pt} %
\renewcommand{\arraystretch}{1} %
\begin{tabular}{cccccccc}
     \multicolumn{2}{c}{\includegraphics[width=0.23\columnwidth]{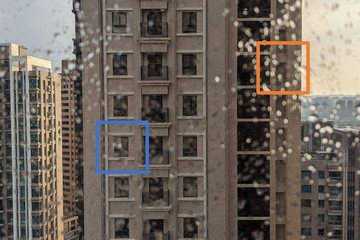}} & 
     \multicolumn{2}{c}{\includegraphics[width=0.23\columnwidth]{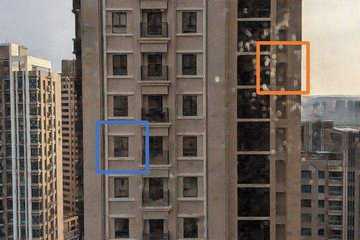}} &
     \multicolumn{2}{c}{\includegraphics[width=0.23\columnwidth]{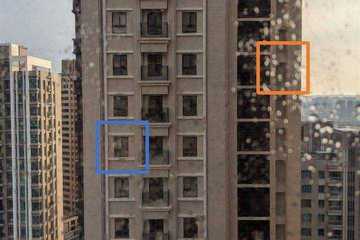}} & \multicolumn{2}{c}{\includegraphics[width=0.23\columnwidth]{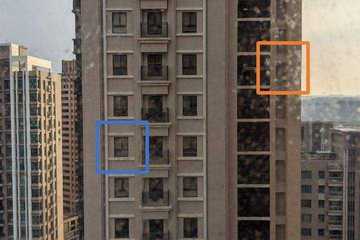}} \\
     \includegraphics[width=0.11\columnwidth]{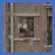} & \includegraphics[width=0.11\columnwidth]{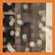} & \includegraphics[width=0.11\columnwidth]{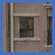} & \includegraphics[width=0.11\columnwidth]{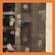} & \includegraphics[width=0.11\columnwidth]{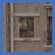} & \includegraphics[width=0.11\columnwidth]{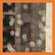} & \includegraphics[width=0.11\columnwidth]{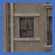} & \includegraphics[width=0.11\columnwidth]{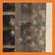} \\
    \multicolumn{2}{c}{Input} & \multicolumn{2}{c}{DeRaindrop~\cite{qian2018attentive}} & \multicolumn{2}{c}{Liu et al.~\cite{liu2020learning}} & \multicolumn{2}{c}{Ours} \\
\end{tabular}
\caption{\textbf{\revisionsecond{A failure example with severe raindrops.}}} 
\label{fig:severe_raindrop}
\end{figure}

\vspace{\paramargin}
{\flushleft {\bf Failure cases.}}
Our method has difficulty in handling complex scenes with multiple layers and highly dynamic objects. 
\figref{failure} shows that our method does not separate the reflection layer well.
This example is particularly challenging as there are two layers of reflections: the top part contains the wooden beams, and the bottom part comes from the street behind the camera.
\figref{failure2} shows an example of a sequence containing a highly dynamic object (e.g., cat). 
As flow estimation cannot compensate for the motion well, our method produces blurry background reconstruction.
\revisionsecond{
Severe occlusions could also cause problems.
In~\figref{severe_raindrop}, we show a scene with severe adherent raindrops, which cause all methods to fail to remove the raindrops.
In the blue zoom-in regions, our method successfully removes the adherent raindrops and recovers scene content better than DeRaindrop~\cite{qian2018attentive}.
However, in the orange zoom-in regions, all methods fail to remove the adherent raindrops due to severe occlusions.
}

%% file: 5_conclusion.tex
\section{Conclusions}
\label{sec:conclusion}

In this work, we propose a novel method for multi-frame reflections and obstructions removal.
Our key insight is to leverage a CNN to reconstruct background and reflection layers from flow-warped images.
Integrating optical flow estimation and coarse-to-fine refinement enable our model to robustly recover the underlying clean images from challenging real-world sequences.
Our method can be applied to different tasks such as fence or adherent raindrop removal with minimum changes in our design.
We also show that online optimization on testing sequences leads to improved visual quality.
Extensive visual comparisons and quantitative evaluation demonstrate that our approach performs well on a wide variety of scenes.

%% file: supp_content.tex
\section{Overview}
\label{sec:overview}
In this supplementary material, we present additional results to complement the main manuscript. 
First, we illustrate the network architecture of the initial flow decomposition network (Section~\ref{sec:network_details}).
Second, we show the detailed procedure for our synthetic reflection sequences generation process  (Section~\ref{sec:dataset_generation}).
Third, we analyze the effect of initial flow decomposition, background/reflection layer reconstruction, TV loss, and realistic training data generation (Section~\ref{sec:additional_analysis}).
Finally, we analyze and visualize the temporal consistency of our video reflection and fence removal results in Section 5. 
We also provide comprehensive visual results on our project website \url{https://alex04072000.github.io/SOLD/}.

\section{Initial Flow Decomposition Network}
\label{sec:network_details}
\begin{figure}
    \centering
    \begin{tabular}{c}
    \includegraphics[width=0.9\columnwidth]{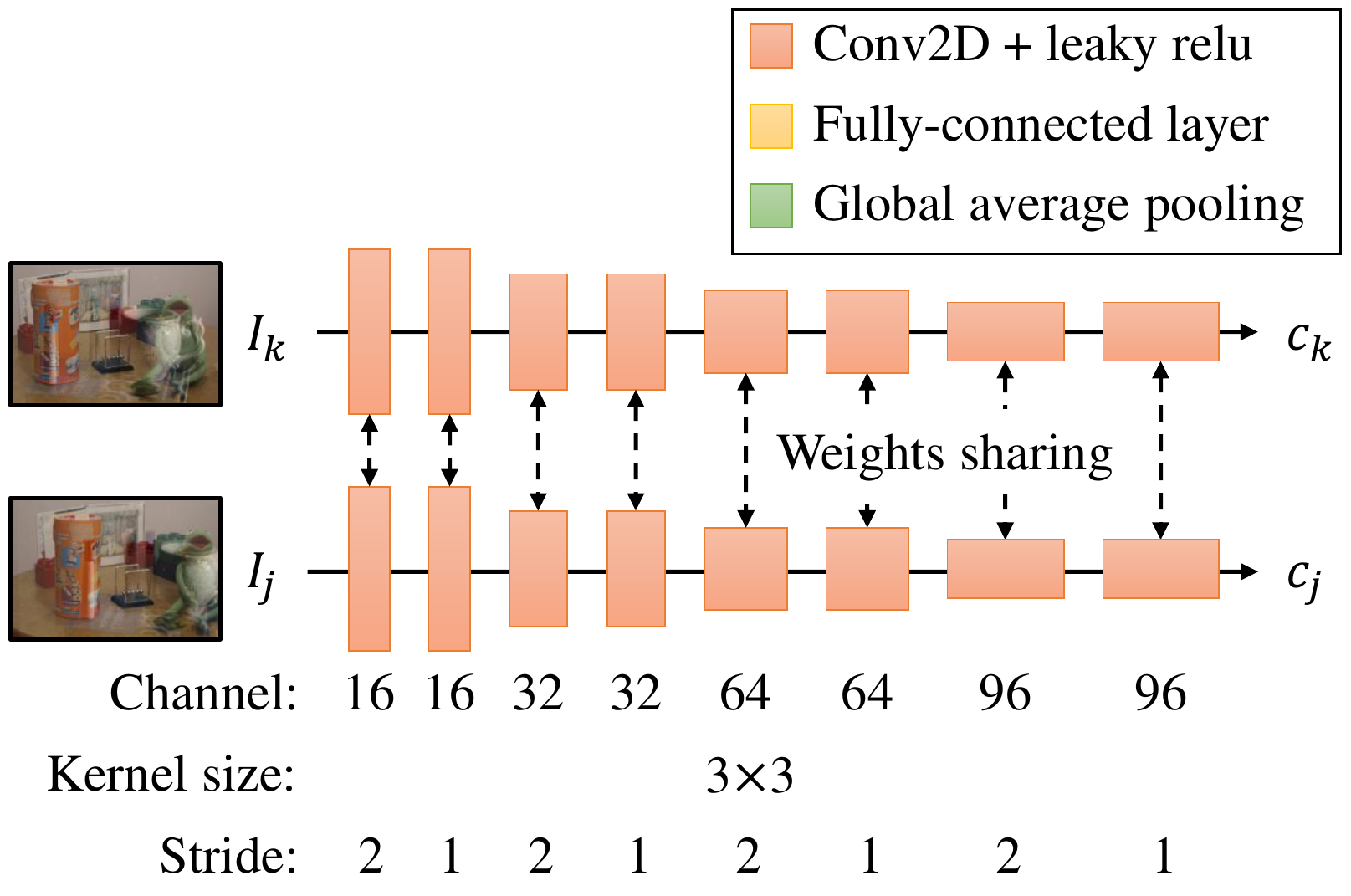} \\
    (a) Feature extractor \\
    \\
    \includegraphics[width=0.9\columnwidth]{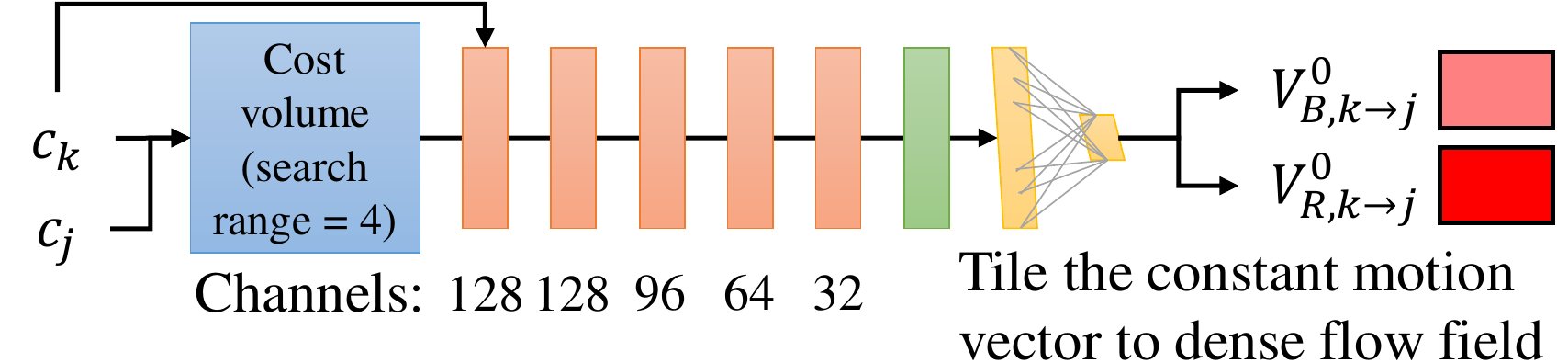} \\
    (b) Layer flow estimator
    \end{tabular}
    \caption{
        \textbf{Architecture of initial flow decomposition network.} 
        Given a keyframe $I_k$ and a reference frame $I_j$, the feature extractor first generates two features $c_k$ and $c_j$. 
        Then, we construct a cost volume with the two features and use six convolutional layers, a global average pooling layer, and a fully connected layer to generate two motion vectors.
        We then tile these two vectors into constant flow fields $V^0_{B, k\rightarrow j}$ and $V^0_{R, k\rightarrow j}$ for the background and reflection layers, respectively.
    }
    \label{fig:initFlow}
\end{figure}

We show the overall architecture of the initial flow decomposition network in Figure~\ref{fig:initFlow}.
Our initial flow decomposition network consists of two sub-modules: 1) a feature extractor, and 2) a layer flow estimator.
The feature extractor first generates deep features from the two input frames, and then the layer flow estimator applies a correlation layer to construct a cost volume from the two input features and predicts a global motion vector through a global average pooling layer.
Finally, we tile the global motion vectors into two \emph{uniform} flow fields $V^0_{B, k\rightarrow j}$ and $V^0_{R, k\rightarrow j}$, for the background and reflection layers, respectively.

\section{Dataset Generation}
\label{sec:dataset_generation}
\begin{figure*}
    \centering
    \begin{tabular}{c}
        \includegraphics[width=0.85\linewidth]{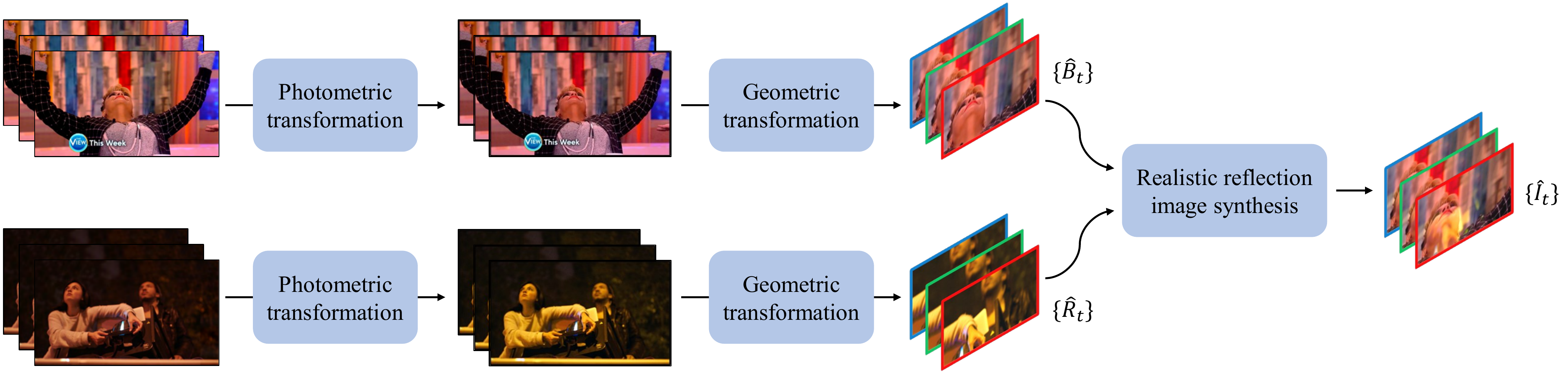}
        \\
        (a) Reflection synthesis 
        \\
        \includegraphics[width=0.85\linewidth]{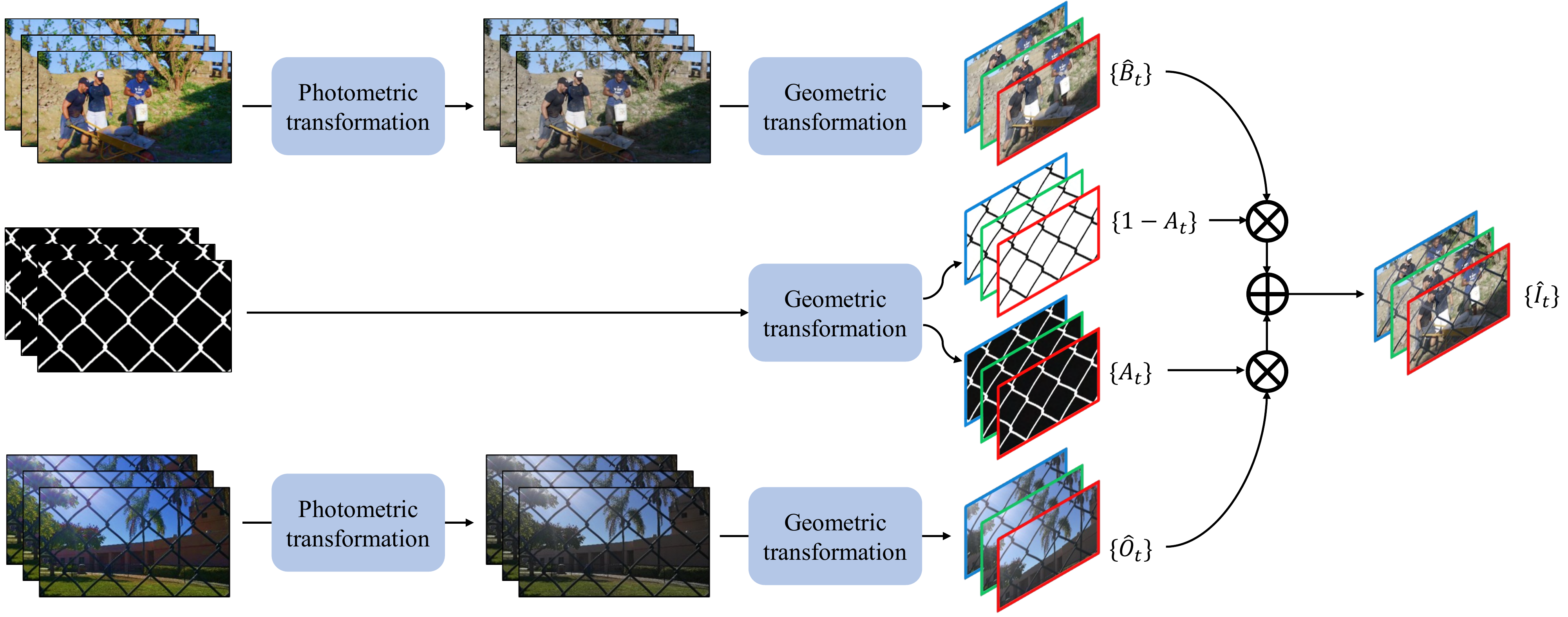} \\
        (b) Fence generation
    \end{tabular}
    \caption{
        \textbf{Synthetic sequence generation.}
        Given two randomly selected sequences, we first apply random color augmentation independently on both the background and foreground layers.
        Then, we apply random homography transformations independently on every frame. 
        Afterward, we apply random walk cropping to simulate camera movements.
        We use the realistic reflection image synthesis model in~\cite{fan2017generic, zhang2018single} to generate a sequence with reflections.
        Finally, we augment random Gaussian noise and random JPEG compression artifacts on every fused frame.
    }
    \label{fig:DA}
\end{figure*}

\begin{figure*}[h!]
\centering
\footnotesize
\renewcommand{\tabcolsep}{1pt} %
\renewcommand{\arraystretch}{1} %
\newcommand{\imagewidth}{0.32\columnwidth}
\begin{tabular}{cccccccc}
    \raisebox{2.0\normalbaselineskip}[0pt][0pt]{\rotatebox[origin=c]{90}{Frame 0}} &
    \includegraphics[width=\imagewidth]{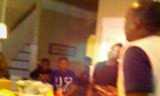} & 
    \includegraphics[width=\imagewidth]{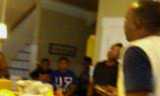} & 
    \includegraphics[width=\imagewidth]{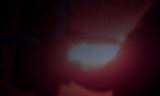} & \raisebox{2.0\normalbaselineskip}[0pt][0pt]{\rotatebox[origin=c]{90}{Frame 0}} &
    \includegraphics[width=\imagewidth]{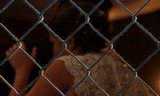} & 
    \includegraphics[width=\imagewidth]{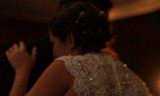} & 
    \includegraphics[width=\imagewidth]{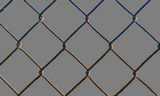} \\
    
    \raisebox{2.0\normalbaselineskip}[0pt][0pt]{\rotatebox[origin=c]{90}{Frame 1}} &
    \includegraphics[width=\imagewidth]{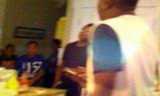} & 
    \includegraphics[width=\imagewidth]{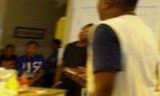} & 
    \includegraphics[width=\imagewidth]{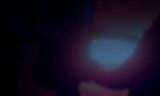} & \raisebox{2.0\normalbaselineskip}[0pt][0pt]{\rotatebox[origin=c]{90}{Frame 1}} &
    \includegraphics[width=\imagewidth]{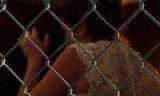} & 
    \includegraphics[width=\imagewidth]{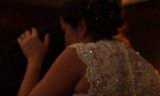} & 
    \includegraphics[width=\imagewidth]{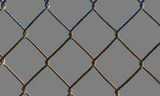} \\
    
    \raisebox{2.0\normalbaselineskip}[0pt][0pt]{\rotatebox[origin=c]{90}{Frame 2}} &
    \includegraphics[width=\imagewidth]{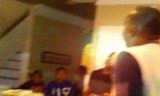} & 
    \includegraphics[width=\imagewidth]{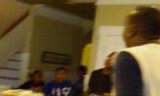} & 
    \includegraphics[width=\imagewidth]{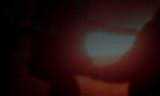} & \raisebox{2.0\normalbaselineskip}[0pt][0pt]{\rotatebox[origin=c]{90}{Frame 2}} &
    \includegraphics[width=\imagewidth]{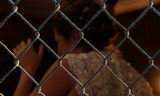} & 
    \includegraphics[width=\imagewidth]{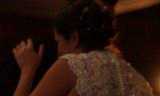} & 
    \includegraphics[width=\imagewidth]{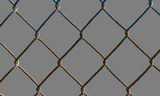} \\
    
    \raisebox{2.0\normalbaselineskip}[0pt][0pt]{\rotatebox[origin=c]{90}{Frame 3}} &
    \includegraphics[width=\imagewidth]{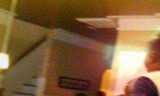} & 
    \includegraphics[width=\imagewidth]{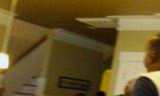} & 
    \includegraphics[width=\imagewidth]{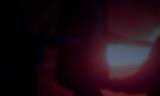} & \raisebox{2.0\normalbaselineskip}[0pt][0pt]{\rotatebox[origin=c]{90}{Frame 3}} &
    \includegraphics[width=\imagewidth]{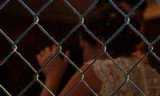} & 
    \includegraphics[width=\imagewidth]{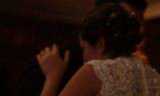} & 
    \includegraphics[width=\imagewidth]{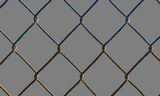} \\
    
    \raisebox{2.0\normalbaselineskip}[0pt][0pt]{\rotatebox[origin=c]{90}{Frame 4}} &
    \includegraphics[width=\imagewidth]{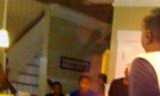} & 
    \includegraphics[width=\imagewidth]{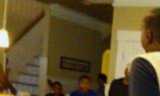} & 
    \includegraphics[width=\imagewidth]{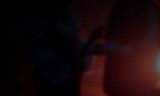} & \raisebox{2.0\normalbaselineskip}[0pt][0pt]{\rotatebox[origin=c]{90}{Frame 4}} &
    \includegraphics[width=\imagewidth]{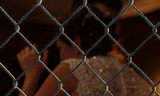} & 
    \includegraphics[width=\imagewidth]{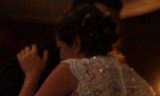} & 
    \includegraphics[width=\imagewidth]{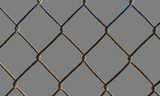} \\
    & $I$ & $\hat{B}$ & $\hat{R}$ & & $I$ & $\hat{B}$ & $\hat{O}$\\
\end{tabular}
\caption{\textbf{Training pairs generated by our synthetic reflection data generation pipeline.}}
\label{fig:DA_examples}
\end{figure*}

During the training stage, we apply on-the-fly random color augmentation, including hue, saturation, brightness, and contrast, on both background and reflection layers, as shown in Figure~\ref{fig:DA}.
We improve the data synthesis pipeline in~\cite{liu2020learning} to generate more diverse training data.
The detailed differences between the reflection image blending method of~\cite{liu2020learning} and ours are summarized in Table~\ref{tab:DA_differences}.
We present examples of the training pairs generated from our pipeline in Figure~\ref{fig:DA_examples}.

\begin{table}[]
    \centering
    \footnotesize
    \caption{\textbf{Detailed differences between the reflection image blending method of~\cite{liu2020learning} and ours.}}
    \label{tab:DA_differences}
    \begin{tabular}{l|cc}
    \toprule 
    Augmentations & \cite{liu2020learning} & Ours \\
    \midrule 
    Kernel size of Gaussian blur of reflection & $\times$ & [3, 17] \\
    \revision{Vignette with random Gaussian kernel size} & $\times$ & [300, 1000] \\
    Random color augmentation & $\times$ & $\checkmark$ \\
    Standard deviations of Gaussian noise & $\times$ & [0, 0.02] \\
    Quality of random JPEG compression & $\times$ & [50, 100] \\
    Motion range between frames & [-20, 20] & [-40, 40] \\
    Number of input frames & 5 & [2, 7] \\
    \bottomrule
    \end{tabular}
\end{table}

\begin{figure*}
\centering
\begin{tabular}{cccc}
     \multicolumn{1}{c|}{\includegraphics[width=0.6\columnwidth]{Figures/results/reflection/input/00002_I2.jpg}} &  \raisebox{3.0\normalbaselineskip}[0pt][0pt]{\rotatebox[origin=c]{90}{Background}} &
     \includegraphics[width=0.6\columnwidth]{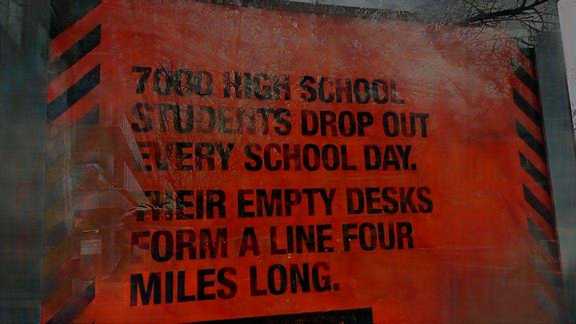} & \includegraphics[width=0.6\columnwidth]{Figures/results/reflection/Ours_online/00002B2_norm.jpg} \\
    \multicolumn{1}{c|}{\includegraphics[width=0.6\columnwidth]{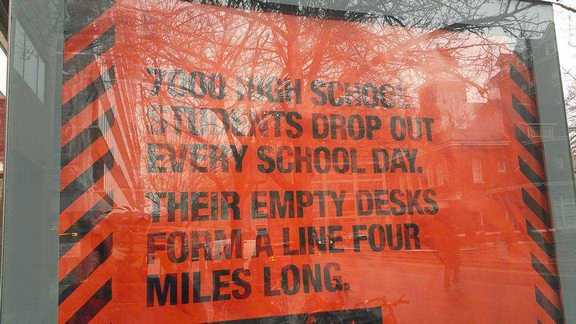}} & \raisebox{3.0\normalbaselineskip}[0pt][0pt]{\rotatebox[origin=c]{90}{Reflection}} &  \includegraphics[width=0.6\columnwidth]{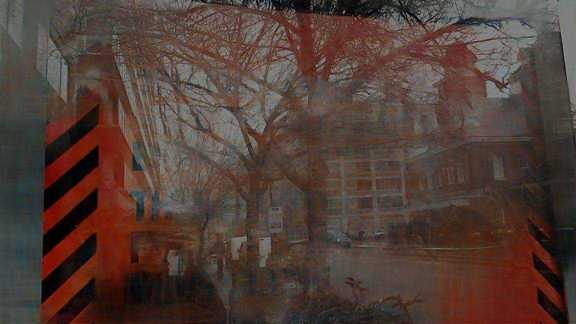} & \includegraphics[width=0.6\columnwidth]{Figures/results/reflection/Ours_online/00002F2_norm.jpg} \\
    (a) Input (two rep. frames) & & (b) Estimate dense flow field & (c) Estimate global translation vector \\
\end{tabular}
\caption{\textbf{Analysis on initialization flow decomposition.} Predicting dense flow fields at the coarsest level often lead to noisy prediction, resulting in the same reconstructed background and foreground layers. Instead, our model predicts global translation vectors as the initial flows, which provide more consistent layer separation at the coarsest level. 
} 
\label{fig:initFlow_ablation}
\end{figure*}

\begin{figure}
\centering
\footnotesize
\renewcommand{\tabcolsep}{1pt} %
\renewcommand{\arraystretch}{1} %
\begin{tabular}{ccccc}
     \multicolumn{1}{c|}{\includegraphics[width=0.23\columnwidth]{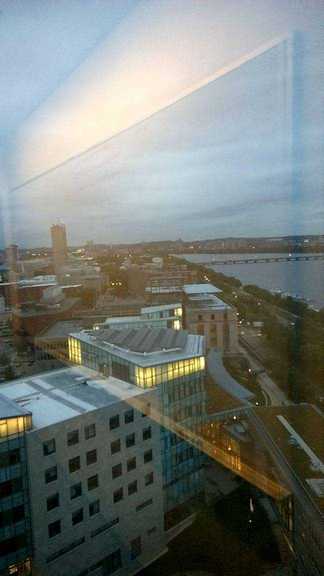}} &  \raisebox{5.0\normalbaselineskip}[0pt][0pt]{\rotatebox[origin=c]{90}{Background}} &
     \includegraphics[width=0.23\columnwidth]{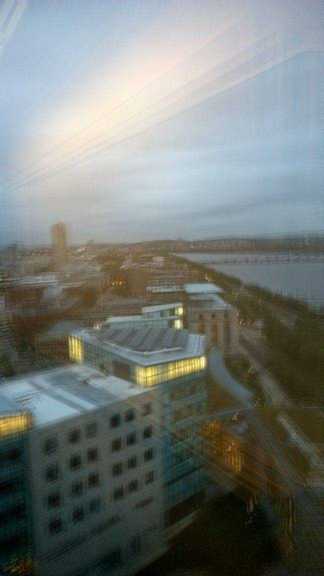} & \includegraphics[width=0.23\columnwidth]{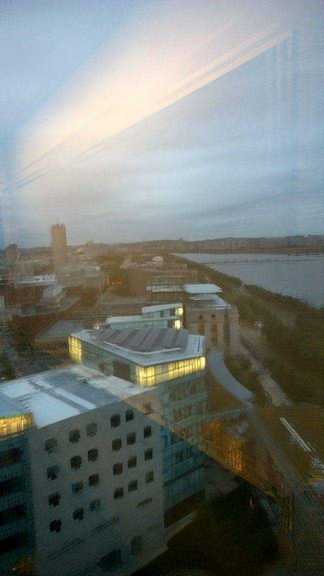} & \includegraphics[width=0.23\columnwidth]{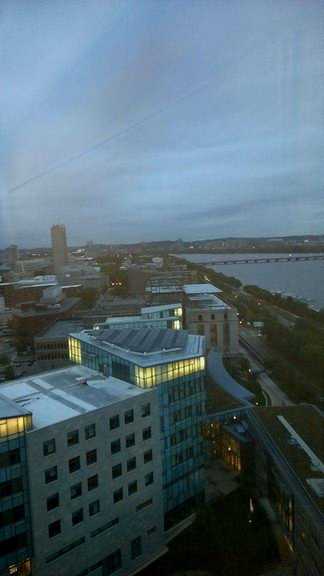} \\
    \multicolumn{1}{c|}{\includegraphics[width=0.23\columnwidth]{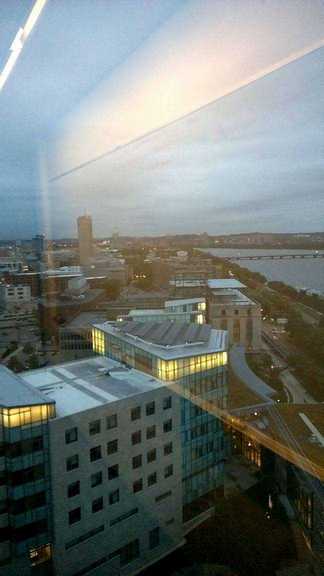}} & \raisebox{5.0\normalbaselineskip}[0pt][0pt]{\rotatebox[origin=c]{90}{Reflection}} &  \includegraphics[width=0.23\columnwidth]{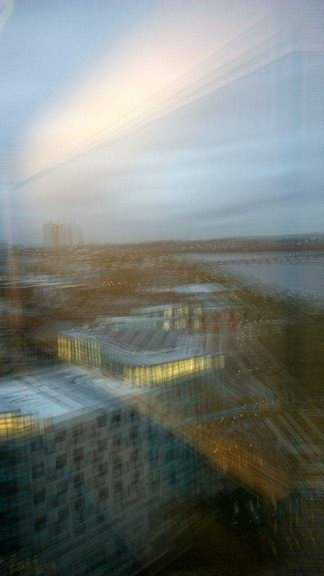} & \includegraphics[width=0.23\columnwidth]{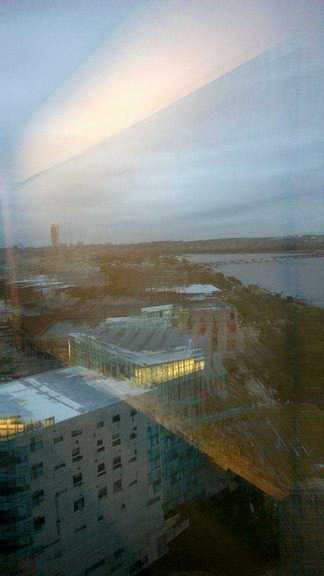} & \includegraphics[width=0.23\columnwidth]{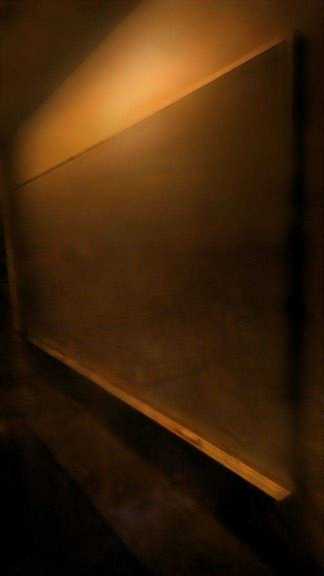} \\
    (a) Inputs & & (b) Mean & (c) Median & (d) Ours \\
\end{tabular}
\caption{\textbf{Effect of image reconstruction network.} Applying a simple mean or median temporal filter to the aligned frames cannot separate the background and reflection layers effectively. In contrast, our image reconstruction network learns to compensate warping errors and provide better separation results.} 
\label{fig:image_net}
\end{figure}

\begin{figure}
\centering
\footnotesize
\renewcommand{\tabcolsep}{1pt} %
\renewcommand{\arraystretch}{1} %
\begin{tabular}{cccc}
     \multicolumn{1}{c|}{\includegraphics[width=0.31\columnwidth]{Figures/results/reflection/input/00009_I2.jpg}} &  \raisebox{2.1\normalbaselineskip}[0pt][0pt]{\rotatebox[origin=c]{90}{Background}} &
     \includegraphics[width=0.31\columnwidth]{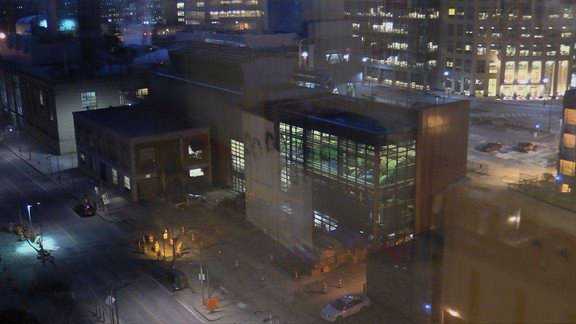} & \includegraphics[width=0.31\columnwidth]{Figures/results/reflection/Ours_online/00009B2_norm.jpg} \\
    \multicolumn{1}{c|}{\includegraphics[width=0.31\columnwidth]{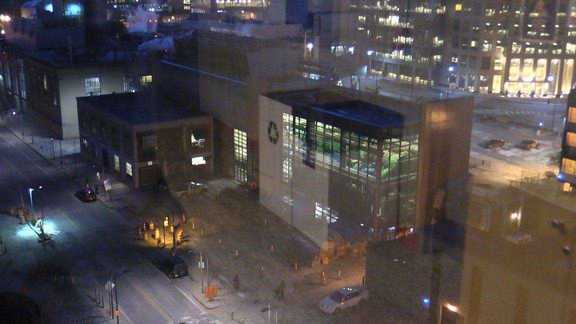}} & \raisebox{2.0\normalbaselineskip}[0pt][0pt]{\rotatebox[origin=c]{90}{Reflection}} &  \includegraphics[width=0.31\columnwidth]{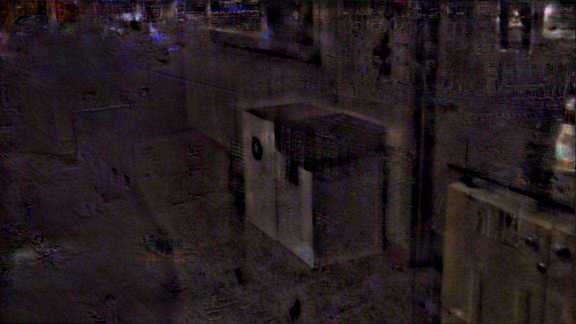} & \includegraphics[width=0.31\columnwidth]{Figures/results/reflection/Ours_online/00009F2_norm.jpg} \\
    (a) Input  & & (b) w/o TV loss & (c) w/ TV loss \\
\end{tabular}
\caption{\textbf{Effect of TV loss in online optimization.}
TV loss regularizes the model to predict sparse image gradients, leading to better separation of reflection and background.
} 
\label{fig:tvloss_ablation}
\end{figure}

\section{Additional Analysis}
\label{sec:additional_analysis}
In this section, we provide additional ablation studies and analysis on our initial flow decomposition, image reconstruction network, TV loss, and realistic training data generation.

\subsection{Initial Flow Decomposition}
Figure~\ref{fig:initFlow_ablation} shows that estimating dense flow fields at the coarsest level may result in noisy predictions and lead to inconsistent layer separation.
In contrast, our uniform flow prediction serves as a good initial prediction to facilitate the following background reconstruction and flow refinement steps.

\subsection{Background/Reflection Layer Reconstruction}
In Figure~\ref{fig:image_net}, we show that the model using temporal mean or median filter for image reconstruction does not perform well and often generates ghosting artifacts.
In contrast, our image reconstruction network learns to reduce warping and alignment errors and generates clean foreground and background images.
\subsection{TV Loss}
Figure~\ref{fig:tvloss_ablation} shows that our online optimization without TV loss results in noisy predictions.
In contrast, TV loss helps the network generating smooth predictions by regularizing sparse image gradient priors.

\subsection{Realistic Training Data Generation}
\begin{figure*}
\centering
\footnotesize
\renewcommand{\tabcolsep}{1pt} %
\renewcommand{\arraystretch}{1} %
\begin{tabular}{cccccccc}
     \multicolumn{1}{c|}{\includegraphics[width=0.3\columnwidth]{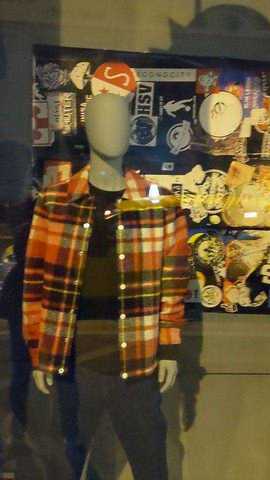}} &  \raisebox{7.5\normalbaselineskip}[0pt][0pt]{\rotatebox[origin=c]{90}{Background}} &
     \includegraphics[width=0.3\columnwidth]{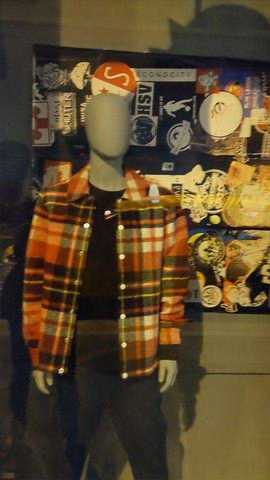} & \includegraphics[width=0.3\columnwidth]{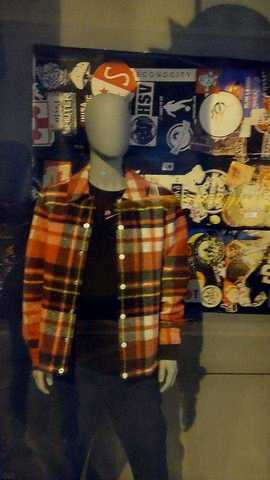} &
     \multicolumn{1}{c|}{\includegraphics[width=0.3\columnwidth]{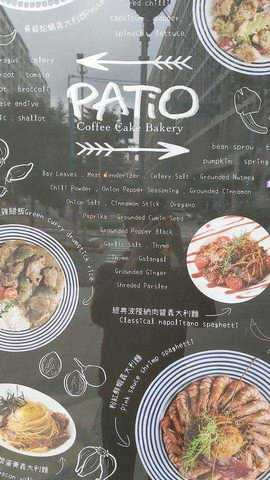}} &  \raisebox{7.5\normalbaselineskip}[0pt][0pt]{\rotatebox[origin=c]{90}{Background}} &
     \includegraphics[width=0.3\columnwidth]{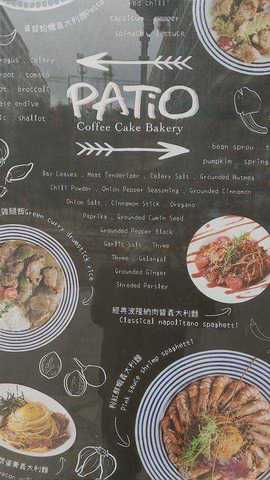} & \includegraphics[width=0.3\columnwidth]{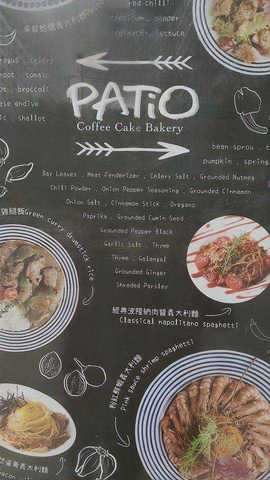} 
     \\
     & \raisebox{7.5\normalbaselineskip}[0pt][0pt]{\rotatebox[origin=c]{90}{Reflection}} &  \includegraphics[width=0.3\columnwidth]{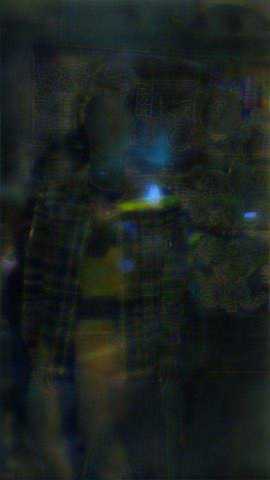} & \includegraphics[width=0.3\columnwidth]{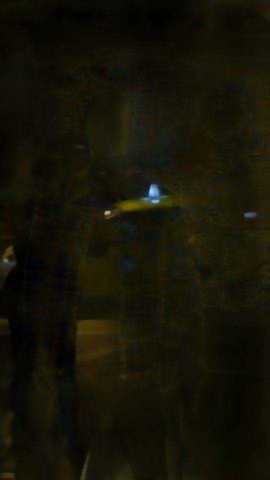} 
     &
     & \raisebox{7.5\normalbaselineskip}[0pt][0pt]{\rotatebox[origin=c]{90}{Reflection}} &  \includegraphics[width=0.3\columnwidth]{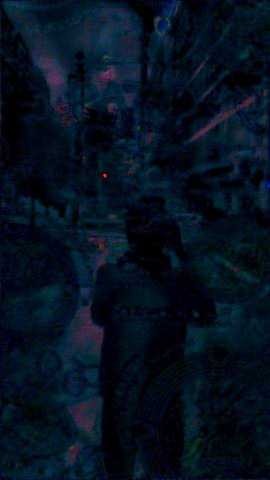} & \includegraphics[width=0.3\columnwidth]{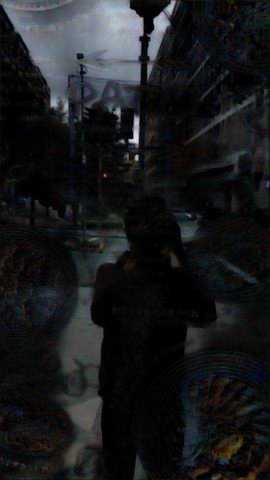} 
     \\
     Input & & Liu et al.~\cite{liu2020learning} & Ours &
     Input & & Liu et al.~\cite{liu2020learning} & Ours
     \\
\end{tabular}
\caption{\textbf{Effect of realistic training data generation.}
} 
\label{fig:realistic_training_data_generation}
\vspace{-3mm}
\end{figure*}

Figure~\ref{fig:realistic_training_data_generation} shows that our realistic training data generation leads to better separation of background and reflection layers both qualitatively and quantitatively.

\begin{figure*}
\centering
\footnotesize
\renewcommand{\tabcolsep}{1pt} %
\renewcommand{\arraystretch}{1} %
\begin{tabular}{ccccccc}
    
    \multicolumn{1}{c|}{\includegraphics[width=0.32\columnwidth]{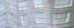}} & &  \multicolumn{1}{c}{\includegraphics[width=0.32\columnwidth]{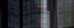}} &  \multicolumn{1}{c}{\includegraphics[width=0.32\columnwidth]{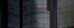}} &  \multicolumn{1}{c}{\includegraphics[width=0.32\columnwidth]{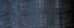}} &  \multicolumn{1}{c}{\includegraphics[width=0.32\columnwidth]{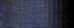}}  &  \multicolumn{1}{c}{\includegraphics[width=0.32\columnwidth]{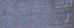}} \\
    \multicolumn{1}{c|}{\includegraphics[width=0.32\columnwidth]{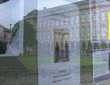}} & \raisebox{3.0\normalbaselineskip}[0pt][0pt]{\rotatebox[origin=c]{90}{Background}} &  \multicolumn{1}{c}{\includegraphics[width=0.32\columnwidth]{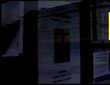}} &  \multicolumn{1}{c}{\includegraphics[width=0.32\columnwidth]{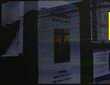}} &  \multicolumn{1}{c}{\includegraphics[width=0.32\columnwidth]{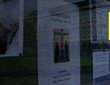}} &  \multicolumn{1}{c}{\includegraphics[width=0.32\columnwidth]{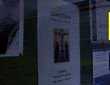}}  &  \multicolumn{1}{c}{\includegraphics[width=0.32\columnwidth]{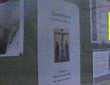}} \\
    \multicolumn{1}{c|}{\includegraphics[width=0.32\columnwidth]{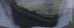}} & &  \multicolumn{1}{c}{\includegraphics[width=0.32\columnwidth]{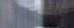}} &  \multicolumn{1}{c}{\includegraphics[width=0.32\columnwidth]{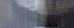}} &  \multicolumn{1}{c}{\includegraphics[width=0.32\columnwidth]{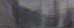}} &  \multicolumn{1}{c}{\includegraphics[width=0.32\columnwidth]{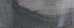}}  &  \multicolumn{1}{c}{\includegraphics[width=0.32\columnwidth]{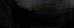}} \\
    \multicolumn{1}{c|}{\includegraphics[width=0.32\columnwidth]{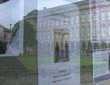}} & \raisebox{3.0\normalbaselineskip}[0pt][0pt]{\rotatebox[origin=c]{90}{Reflection}} &  \multicolumn{1}{c}{\includegraphics[width=0.32\columnwidth]{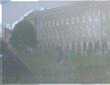}} &  \multicolumn{1}{c}{\includegraphics[width=0.32\columnwidth]{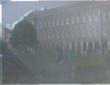}} &  \multicolumn{1}{c}{\includegraphics[width=0.32\columnwidth]{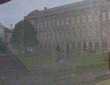}} &  \multicolumn{1}{c}{\includegraphics[width=0.32\columnwidth]{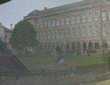}}  &  \multicolumn{1}{c}{\includegraphics[width=0.32\columnwidth]{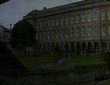}} \\

    \multicolumn{1}{c|}{\includegraphics[width=0.32\columnwidth]{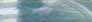}} & &  \multicolumn{1}{c}{\includegraphics[width=0.32\columnwidth]{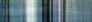}} &  \multicolumn{1}{c}{\includegraphics[width=0.32\columnwidth]{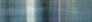}} &  \multicolumn{1}{c}{\includegraphics[width=0.32\columnwidth]{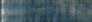}} &  \multicolumn{1}{c}{\includegraphics[width=0.32\columnwidth]{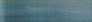}}  &  \multicolumn{1}{c}{\includegraphics[width=0.32\columnwidth]{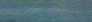}} \\
    \multicolumn{1}{c|}{\includegraphics[width=0.32\columnwidth]{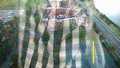}} & \raisebox{2.0\normalbaselineskip}[0pt][0pt]{\rotatebox[origin=c]{90}{Background}} &  \multicolumn{1}{c}{\includegraphics[width=0.32\columnwidth]{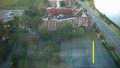}} &  \multicolumn{1}{c}{\includegraphics[width=0.32\columnwidth]{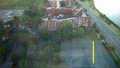}} &  \multicolumn{1}{c}{\includegraphics[width=0.32\columnwidth]{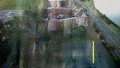}} &  \multicolumn{1}{c}{\includegraphics[width=0.32\columnwidth]{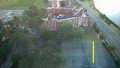}}  &  \multicolumn{1}{c}{\includegraphics[width=0.32\columnwidth]{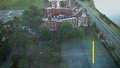}} \\
    \multicolumn{1}{c|}{\includegraphics[width=0.32\columnwidth]{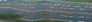}} & &  \multicolumn{1}{c}{\includegraphics[width=0.32\columnwidth]{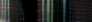}} &  \multicolumn{1}{c}{\includegraphics[width=0.32\columnwidth]{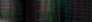}} &  \multicolumn{1}{c}{\includegraphics[width=0.32\columnwidth]{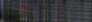}} &  \multicolumn{1}{c}{\includegraphics[width=0.32\columnwidth]{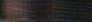}}  &  \multicolumn{1}{c}{\includegraphics[width=0.32\columnwidth]{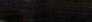}} \\
    \multicolumn{1}{c|}{\includegraphics[width=0.32\columnwidth]{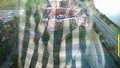}} & \raisebox{2.0\normalbaselineskip}[0pt][0pt]{\rotatebox[origin=c]{90}{Reflection}} &  \multicolumn{1}{c}{\includegraphics[width=0.32\columnwidth]{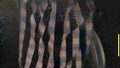}} &  \multicolumn{1}{c}{\includegraphics[width=0.32\columnwidth]{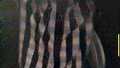}} &  \multicolumn{1}{c}{\includegraphics[width=0.32\columnwidth]{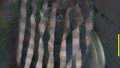}} &  \multicolumn{1}{c}{\includegraphics[width=0.32\columnwidth]{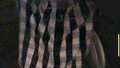}}  &  \multicolumn{1}{c}{\includegraphics[width=0.32\columnwidth]{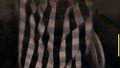}} \\

    Input (rep. frame) & & Xue~\etal +~\cite{xue2015computational} & Xue~\etal ++~\cite{xue2015computational} & Yang~\etal ++~\cite{yang2016robust} & Nandoriya~\etal~\cite{nandoriya2017video} & Ours\\
\end{tabular}
\caption{\textbf{Our method generate better layer separation with temporal coherency (yellow slice).} '+': applies the original method using moving window strategy as mentioned in~\cite{yang2016robust}. '++': uses a moving temporal average filtering to reduce flickering based on '+'.}
\label{fig:video}
\end{figure*}

\begin{figure*}
\centering
\footnotesize
\renewcommand{\tabcolsep}{1pt} %
\renewcommand{\arraystretch}{1} %
\newcommand{\imagewidth}{0.235\textwidth} %
\begin{tabular}{ccccc}
     \raisebox{5.0\normalbaselineskip}[0pt][0pt]{\rotatebox[origin=c]{90}{Input}} &  
     \includegraphics[width=\imagewidth]{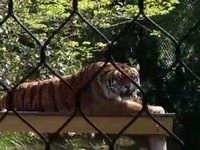} &
     \includegraphics[width=\imagewidth]{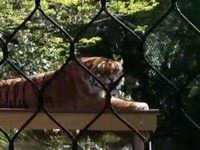} &
     \includegraphics[width=\imagewidth]{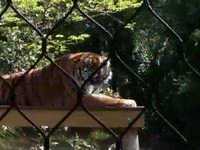} &
     \includegraphics[width=\imagewidth]{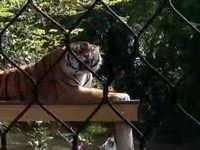} \\
     \raisebox{5.0\normalbaselineskip}[0pt][0pt]{\rotatebox[origin=c]{90}{Background}} &
     \includegraphics[width=\imagewidth]{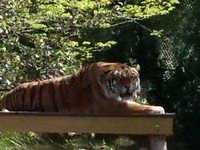} &
     \includegraphics[width=\imagewidth]{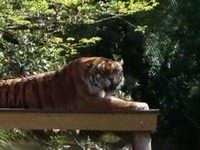} &
     \includegraphics[width=\imagewidth]{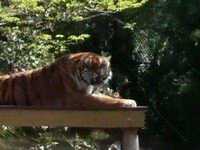} &
     \includegraphics[width=\imagewidth]{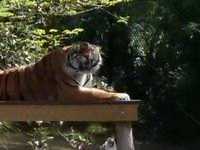} \\
     \raisebox{5.0\normalbaselineskip}[0pt][0pt]{\rotatebox[origin=c]{90}{Input}} &       
     \includegraphics[width=\imagewidth]{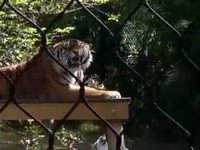} &
     \includegraphics[width=\imagewidth]{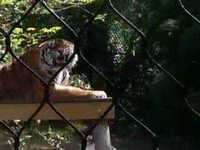} &
     \includegraphics[width=\imagewidth]{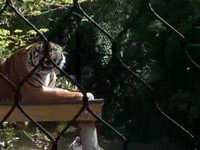} &
     \includegraphics[width=\imagewidth]{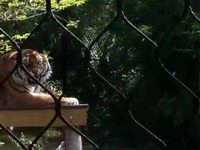} \\
     \raisebox{5.0\normalbaselineskip}[0pt][0pt]{\rotatebox[origin=c]{90}{Background}} &
     \includegraphics[width=\imagewidth]{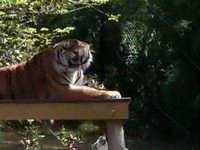} &
     \includegraphics[width=\imagewidth]{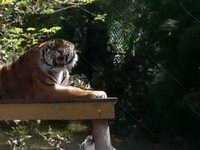} &
     \includegraphics[width=\imagewidth]{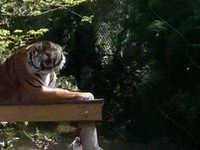} &
     \includegraphics[width=\imagewidth]{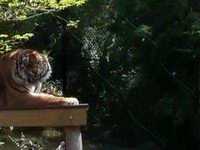} \\
\end{tabular}
\caption{\textbf{Video results for fence removal.} Our method can still generate temporally consistent results when there are moving objects in the scene, e.g. the tiger in this example.}
\label{fig:moving_tiger}
\end{figure*}

\begin{figure*}
\centering
\footnotesize
\renewcommand{\tabcolsep}{1pt} %
\renewcommand{\arraystretch}{1} %
\begin{tabular}{cccc|cccc|cccc}
     \multicolumn{1}{c}{\includegraphics[width=0.2\columnwidth]{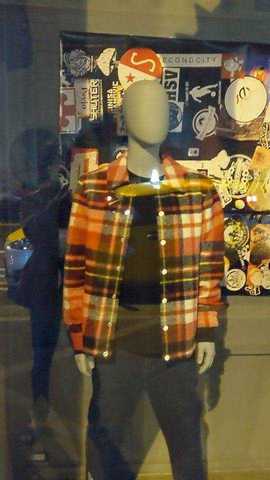}} &  \raisebox{5.0\normalbaselineskip}[0pt][0pt]{\rotatebox[origin=c]{90}{Background}} &
     \includegraphics[width=0.2\columnwidth]{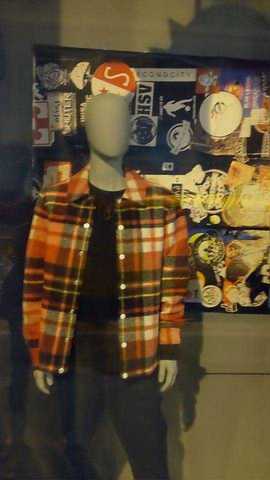} & \includegraphics[width=0.2\columnwidth]{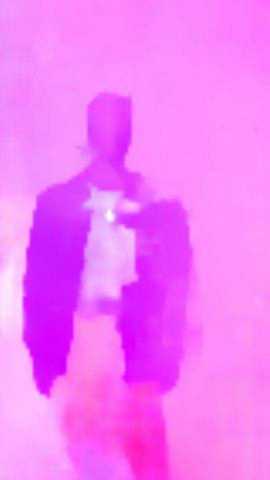} & \multicolumn{1}{c}{\includegraphics[width=0.2\columnwidth]{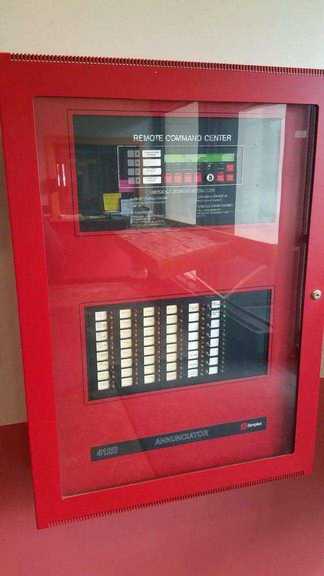}} &  \raisebox{5.0\normalbaselineskip}[0pt][0pt]{\rotatebox[origin=c]{90}{Background}} &
     \includegraphics[width=0.2\columnwidth]{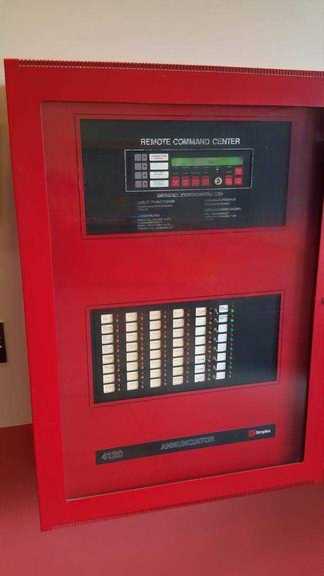} & \includegraphics[width=0.2\columnwidth]{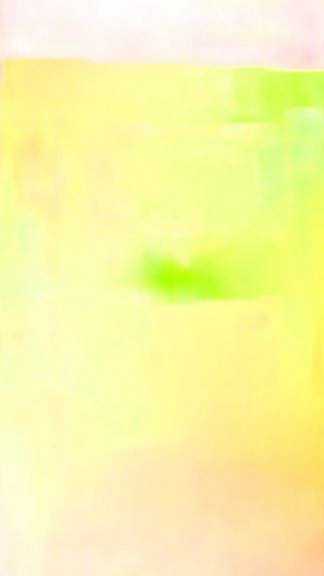} & \multicolumn{1}{c}{\includegraphics[width=0.2\columnwidth]{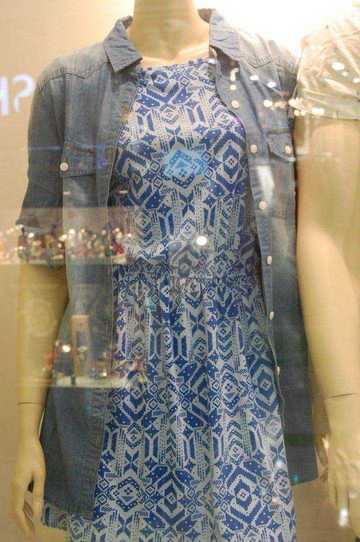}} &  \raisebox{5.0\normalbaselineskip}[0pt][0pt]{\rotatebox[origin=c]{90}{Background}} &
     \includegraphics[width=0.2\columnwidth]{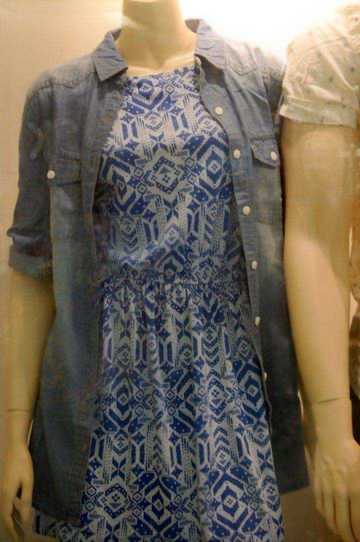} & \includegraphics[width=0.2\columnwidth]{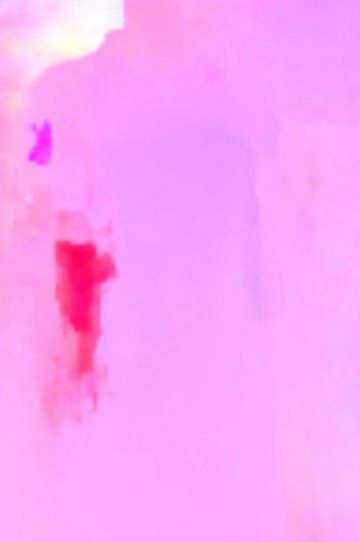} \\
    \multicolumn{1}{c}{\includegraphics[width=0.2\columnwidth]{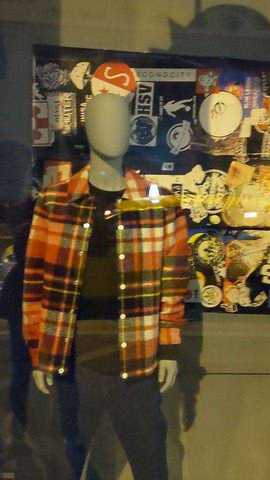}} & \raisebox{5.0\normalbaselineskip}[0pt][0pt]{\rotatebox[origin=c]{90}{Reflection}} &  \includegraphics[width=0.2\columnwidth]{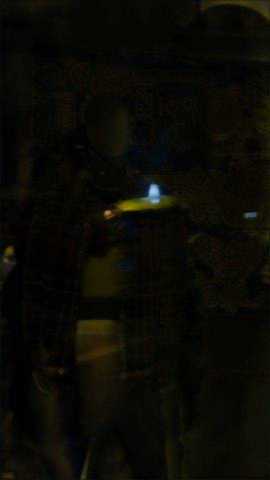} & \includegraphics[width=0.2\columnwidth]{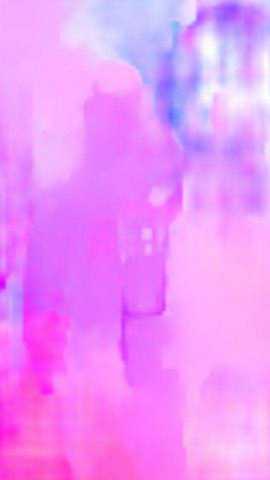} & \multicolumn{1}{c}{\includegraphics[width=0.2\columnwidth]{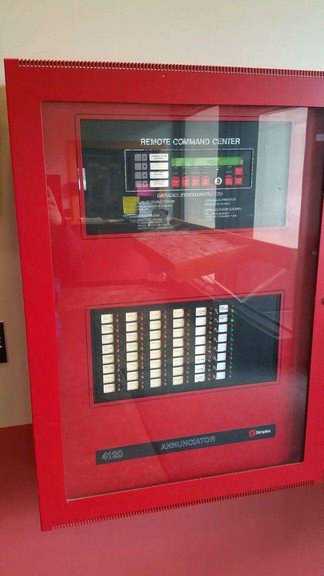}} & \raisebox{5.0\normalbaselineskip}[0pt][0pt]{\rotatebox[origin=c]{90}{Reflection}} &  \includegraphics[width=0.2\columnwidth]{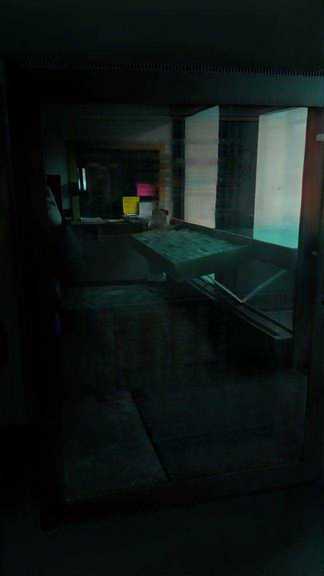} & \includegraphics[width=0.2\columnwidth]{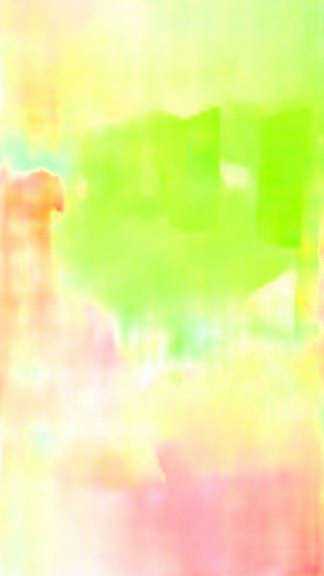} & \multicolumn{1}{c}{\includegraphics[width=0.2\columnwidth]{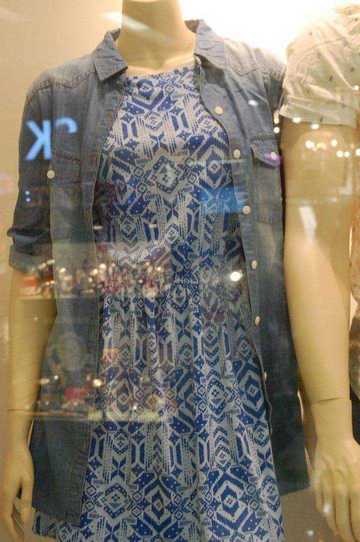}} & \raisebox{5.0\normalbaselineskip}[0pt][0pt]{\rotatebox[origin=c]{90}{Reflection}} &  \includegraphics[width=0.2\columnwidth]{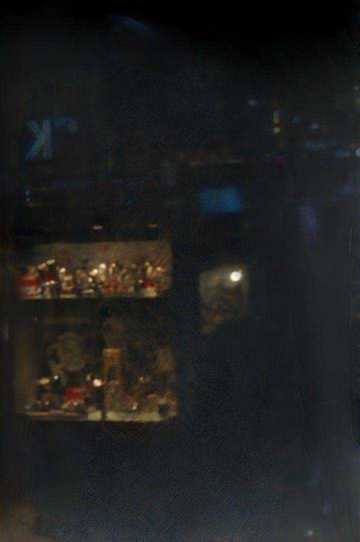} & \includegraphics[width=0.2\columnwidth]{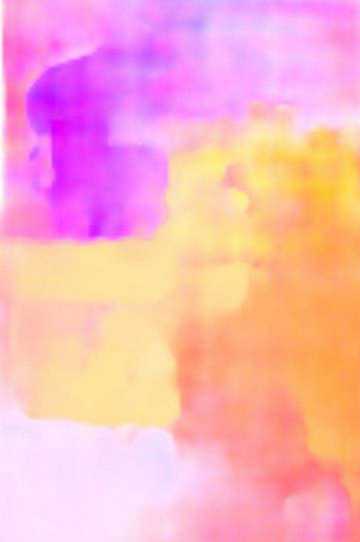} \\

    Inputs & & Image & Optical flow & Inputs & & Image & Optical flow & Inputs & & Image & Optical flow \\
\end{tabular}
\caption{\revision{\textbf{Predicted optical flows.} 
On the left, we show two representative input frames of the sequence. 
The middle shows the recovered background and reflection. 
On the right, we show the predicted flows for the background and reflection layers.}
} 
\label{fig:flow_example}
\end{figure*}

\revision{
\subsection{Predicted optical flow results}
We show the predicted optical flows for real-world sequences in~\figref{flow_example}.
}

\section{Temporal Coherency}
The proposed method takes five neighboring frames as input and generates the separation results for the reference frame.
Although predicting each reference frame \emph{independently}, our method still generates temporally coherent results on the entire video.
Here, we compare our method with four video reflection removal approaches~\cite{xue2015computational, yang2016robust, nandoriya2017video}.
Both methods by Xue~\etal~\cite{xue2015computational} and Yang~\etal~\cite{yang2016robust} use multiple frames as input and generates the middle frame, similar to our model.
Xue~\etal+~\cite{xue2015computational} is an extension of~\cite{xue2015computational} which uses the moving window strategy in~\cite{yang2016robust} to improve the temporal consistency.
Both Xue~\etal++~\cite{xue2015computational} and Yang~\etal++~\cite{yang2016robust} adopt a temporal average filtering to reduce the temporal flickering.
Nandoriya~\etal~\cite{nandoriya2017video} use a spatio-temporal optimization to process the entire video sequence jointly.

We evaluate temporal consistency of each method on a controlled synthetic video sequence provided by~\cite{nandoriya2017video}, which blends two videos through an alpha blending.
The two layers have different global movements.
In addition, there is a third layer on the background which contains a flying bird to simulate local moving objects. 
In Figure~\ref{fig:video}, we show separation results on real input sequences, where the proposed method not only separates the background and reflection layers well but also preserves temporal coherency.
Figure~\ref{fig:moving_tiger} shows another example that our method can deal with moving scenes objects.

\ignore{
\begin{figure*}
\centering
\footnotesize
\renewcommand{\tabcolsep}{1pt} %
\renewcommand{\arraystretch}{1} %
\newcommand{\imagewidth}{0.235\columnwidth} %
\begin{tabular}{ccccccccc}
     \raisebox{2.2\normalbaselineskip}[0pt][0pt]{\rotatebox[origin=c]{90}{Input}} &  
     \includegraphics[width=\imagewidth]{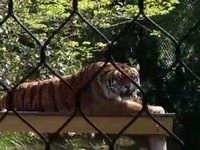} &
     \includegraphics[width=\imagewidth]{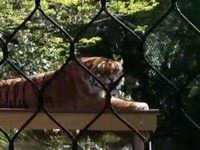} &
     \includegraphics[width=\imagewidth]{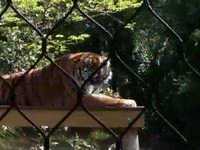} &
     \includegraphics[width=\imagewidth]{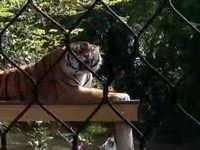} &
     \includegraphics[width=\imagewidth]{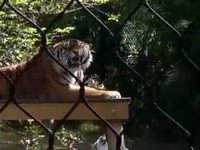} &
     \includegraphics[width=\imagewidth]{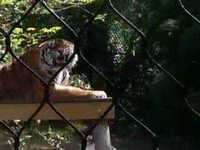} &
     \includegraphics[width=\imagewidth]{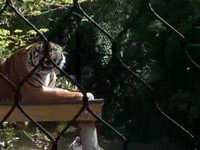} &
     \includegraphics[width=\imagewidth]{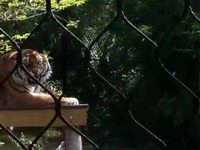} \\
     \raisebox{2.1\normalbaselineskip}[0pt][0pt]{\rotatebox[origin=c]{90}{Background}} &
     \includegraphics[width=\imagewidth]{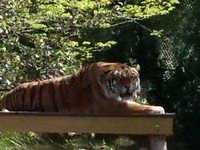} &
     \includegraphics[width=\imagewidth]{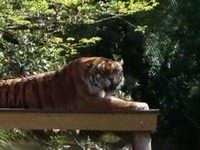} &
     \includegraphics[width=\imagewidth]{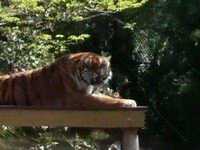} &
     \includegraphics[width=\imagewidth]{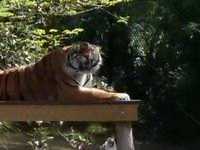} &
     \includegraphics[width=\imagewidth]{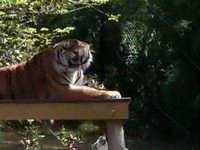} &
     \includegraphics[width=\imagewidth]{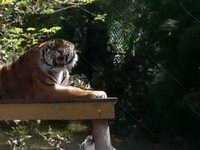} &
     \includegraphics[width=\imagewidth]{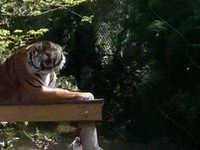} &
     \includegraphics[width=\imagewidth]{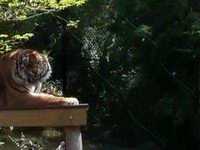} \\
\end{tabular}
\caption{\textbf{Video results for fence removal.} Our method can still generate temporally consistent results when there are moving objects in the scene, e.g. the tiger in this example.}
\label{fig:moving_tiger}
\end{figure*}
}